\documentclass{article}

\PassOptionsToPackage{numbers, compress}{natbib}

\usepackage[preprint]{neurips_2023}

\usepackage[utf8]{inputenc} 
\usepackage[T1]{fontenc}    
\usepackage{hyperref}       
\usepackage{url}            
\usepackage{booktabs}       
\usepackage{amsfonts}       
\usepackage{nicefrac}       
\usepackage{microtype}      
\usepackage{xcolor}         
\usepackage{booktabs}
\usepackage{graphicx}
\usepackage{subcaption}
\usepackage{todonotes}
\usepackage{wrapfig}
\usepackage[ruled, vlined, linesnumbered]{algorithm2e}%
\usepackage{algpseudocode}
\usepackage{xspace}
\usepackage{outlines}
\usepackage{enumitem}
\usepackage{rotating}
\usepackage{tabulary}
\usepackage[toc,page,header]{appendix}
\usepackage{minitoc}

\usepackage{amsmath}
\usepackage{amssymb}
\usepackage{mathtools}
\usepackage{amsthm}
\theoremstyle{plain}

\theoremstyle{definition}

\theoremstyle{remark}

\let\cite\citep

\newcommand{\model}{\textsc{Milo}}

\usepackage{amsmath,amsfonts,bm}

\def\figref#1{figure~\ref{#1}}

\def\eqref#1{equation~\ref{#1}}

\def\1{\bm{1}}

\def\vtheta{{\bm{\theta}}}

\def\vg{{\bm{g}}}

\def\vp{{\bm{p}}}

\DeclareMathAlphabet{\mathsfit}{\encodingdefault}{\sfdefault}{m}{sl}
\SetMathAlphabet{\mathsfit}{bold}{\encodingdefault}{\sfdefault}{bx}{n}

\def\gA{{\mathcal{A}}}
\def\gB{{\mathcal{B}}}

\def\gD{{\mathcal{D}}}

\def\gK{{\mathcal{K}}}

\def\gO{{\mathcal{O}}}

\def\gS{{\mathcal{S}}}

\def\gU{{\mathcal{U}}}

\newcommand{\R}{\mathbb{R}}

\DeclareMathOperator*{\argmax}{arg\,max}

\renewcommand{\figref}[1]{Figure~\ref{#1}}

\renewcommand{\eqref}[1]{Equation~(\ref{#1})}

\title{\model\ : Model-Agnostic Subset Selection Framework for Efficient Model Training and Tuning}

\author{Krishnateja Killamsetty\textsuperscript{1, 2, \thanks{A portion of this work was completed while Krishnateja was an intern at IBM Research.}} , Alexandre V. Evfimievski\textsuperscript{2}, Tejaswini Pedapati \textsuperscript{2}\\
        \textbf{Kiran Kate\textsuperscript{2} \quad Lucian Popa\textsuperscript{2}, \quad Rishabh Iyer\textsuperscript{1}}\\
    \textsuperscript{1} The University of Texas at Dallas  \\
    \textsuperscript{2} IBM Research\\
    \texttt{\{krishnateja.killamsetty, rishabh.iyer\}@utdallas.edu}\\
    \texttt{\{evfimi, tejaswinip, kakate, lpopa\}@us.ibm.com}\\
}

\begin{document}
\maketitle
\doparttoc 
\faketableofcontents 

\begin{abstract}
Training deep networks and tuning hyperparameters on large datasets is computationally intensive. One of the primary research directions for efficient training is to reduce training costs by selecting well-generalizable subsets of training data. Compared to simple adaptive random subset selection baselines, existing intelligent subset selection approaches are not competitive due to the time-consuming subset selection step, which involves computing model-dependent gradients and feature embeddings and applies greedy maximization of submodular objectives. Our key insight is that removing the reliance on downstream model parameters enables subset selection as a pre-processing step and enables one to train multiple models at no additional cost. In this work, we propose \model{}, a model-agnostic subset selection framework that decouples the subset selection from model training while enabling superior model convergence and performance by using an easy-to-hard curriculum. Our empirical results indicate that \model{} can train models $3\times - 10 \times$ faster and tune hyperparameters $20\times - 75 \times$ faster than full-dataset training or tuning without compromising performance.

\end{abstract}

\section{Introduction}
Deep learning has achieved remarkable success in a multitude of machine learning tasks, including natural language processing, computer vision, and speech recognition in recent years. This success is partially due to the availability of massive training datasets and the capacity to train large-scale neural networks. However, training deep models on extensive datasets is computationally demanding, incurring significant financial costs and generating substantial CO2 emissions \cite{strubell-etal-2019-energy, Schwartz2020GreenA}. Bhavya et al. \cite{bhavya2022compute} overview several research trajectories aimed at enhancing model convergence and reducing training time and costs, including data subset selection, curriculum learning, model architecture improvements, and optimization algorithm enhancements. Our work specifically focuses on selecting useful, generalizable data subsets for efficient deep neural network training.

Recent research \cite{toneva2018an, birodkar2019semantic, pmlr-v80-katharopoulos18a} suggests that many current training datasets are redundant, indicating that a non-redundant, informative subset could achieve comparable performance to training on the full dataset. To identify such informative samples, metrics such as prediction uncertainty \cite{Coleman2020Selection}, prediction flips \cite{toneva2018an, dynamicinstancehardness}, loss \cite{loshchilov2015online}, or gradient/gradient-norm \cite{pmlr-v80-katharopoulos18a, mirzasoleiman2020coresets, killamsetty2021glister, killamsetty2021grad, killamsetty2021retrieve} have been applied. These metrics, calculated using either the downstream machine learning model or a lightweight surrogate \cite{Coleman2020Selection}, require a fully converged model, a requirement that runs counter to the goal of efficient training. To address this, existing subset selection algorithms \cite{mirzasoleiman2020coresets, killamsetty2021glister, killamsetty2021grad, killamsetty2021retrieve, killamsetty2022automata, pmlr-v162-pooladzandi22a} for efficient learning utilize the downstream model during training for heuristic computation and periodically update the subset as the model-dependent metrics for each sample evolve.

\begin{wrapfigure}{l}{0.5\textwidth}
\centering
\begin{subfigure}[b]{0.24\textwidth}
    \includegraphics[height=2.5cm, width=2.7cm]{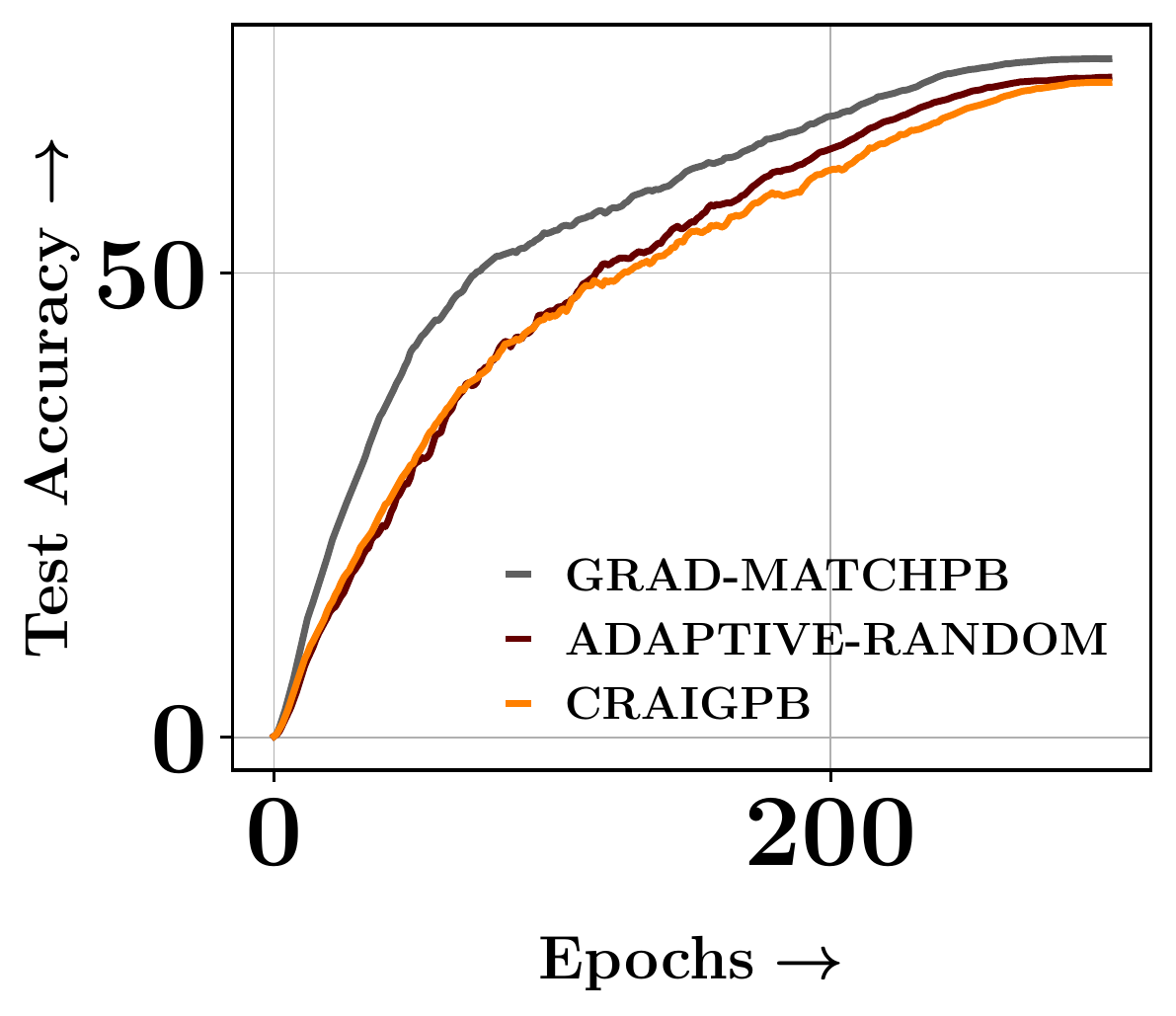}
    \caption{}
    \label{subfig:epochconvergence}
\end{subfigure}
  \hfill 
\begin{subfigure}[b]{0.24\textwidth}
    \includegraphics[height=2.5cm, width=2.7cm]{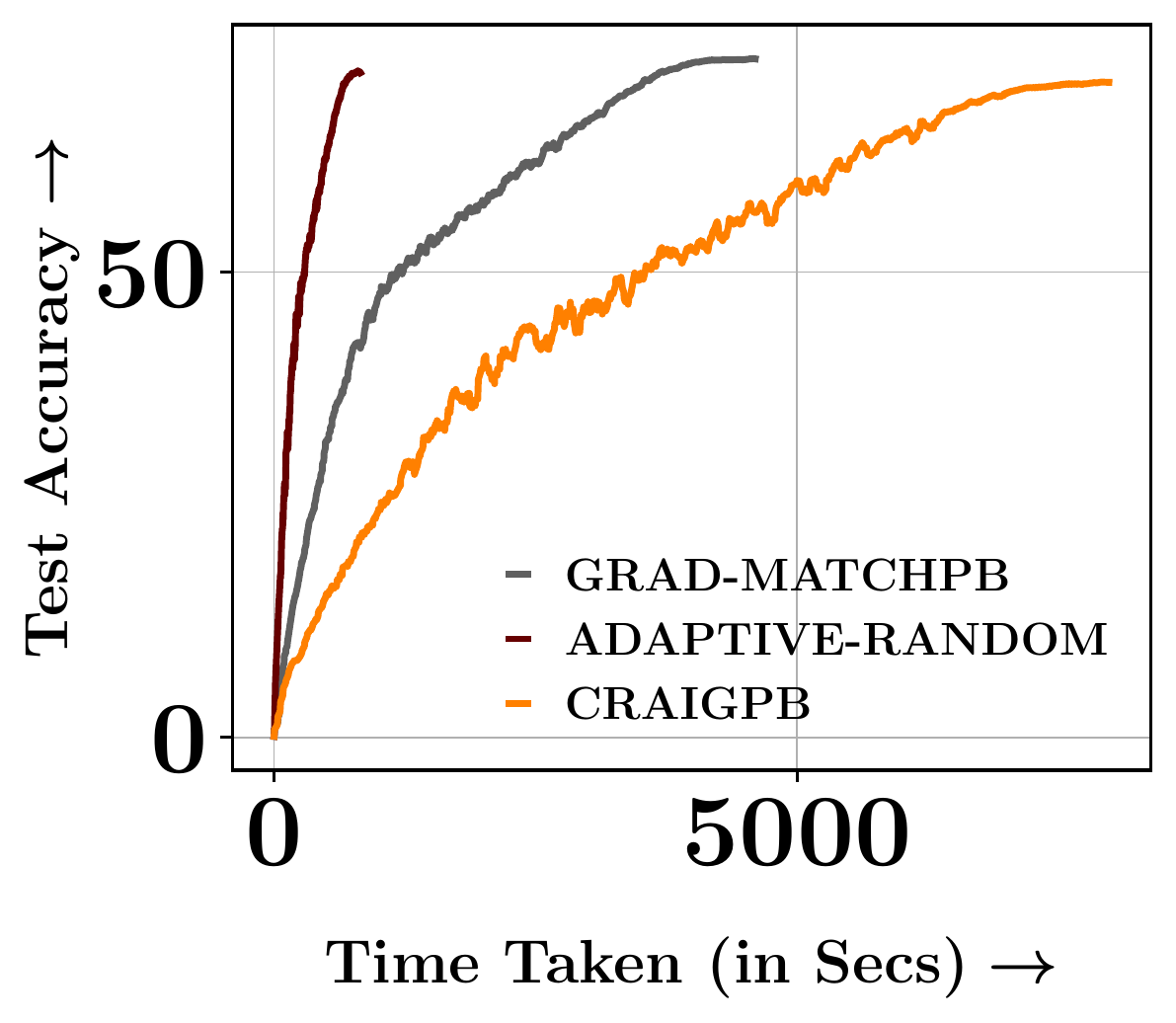}
    \caption{}
    \label{subfig:timeconvergence}
\end{subfigure}
\vspace{-2mm}
\caption{\footnotesize{Sub-figures (a) and (b) display the convergence of the ResNet18 model trained on a 10\% subset selected using Adaptive-Random, \textsc{CraigPB}, and \textsc{GradMatchPB} on the CIFAR100 dataset, with respect to epochs and time, respectively. Each strategy selects a new subset every epoch.}}
\vspace{-3mm}
\label{fig:randvsgrad}
\end{wrapfigure}
\noindent\textbf{Drawbacks of Model-Dependent Subset Selection: } Despite the theoretical advantages of existing subset selection strategies\cite{killamsetty2021glister, killamsetty2021grad, mirzasoleiman2020coresets}, they often fall short in computational efficiency compared to adaptive random subset selection, which selects random subsets periodically. This is primarily because traditional approaches rely on downstream models and typically require the calculation of sample metrics, such as gradients, before each subset selection step. Furthermore, these computationally demanding subset selection steps occur during model training. For instance, Figure \ref{fig:randvsgrad} compares the convergence rate of the ResNet18 model on the CIFAR100 dataset, in terms of both time and epochs. Here, the model uses 10\% subsets chosen every epoch by GradMatchPB \citep{killamsetty2021grad}, a state-of-the-art data subset selection strategy for efficient training, CraigPB \citep{mirzasoleiman2020coresets}, and Adaptive-Random (where a new 10\% subset is randomly selected periodically). The selection of a new subset every epoch is done to demonstrate the maximum performance attainable by GradMatchPB and CraigPB. The results indicate that GradMatchPB achieves faster epoch convergence than both Adaptive-Random and CraigPB when selecting a new subset each epoch. However, due to the necessity for a computationally intensive subset selection step every epoch, both GradMatchPB and CraigPB demonstrate considerable inefficiency in terms of training time. In their respective studies, Killamsetty et al. \citep{killamsetty2021grad} and Mirzasoleiman et al. \citep{mirzasoleiman2020coresets} suggested selecting a new subset every $R$ epochs to enhance training efficiency. However, this comes at the cost of the model's convergence rate. Lastly, model-dependent subset selection requires a computationally intensive subset selection steps each time a new model is trained.

\noindent\textbf{Model-Agnostic Selection: }Our primary insight is that a model-agnostic subset selection framework can circumvent the computationally intensive subset selection steps during model training by conducting subset selection in the preprocessing phase. By pre-selecting subsets and storing them as metadata with each dataset, we can train numerous models without incurring additional costs, effectively spreading out the expense of subset selection. In this work, we endeavor to answer the following question: \emph{\textbf{Is it possible to develop a model-agnostic subset selection method that selects new subsets in a minimal amount of time, yet achieves superior model convergence without significant compromise in test accuracy or generalization performance?}}


\subsection{Contributions} 
\begin{figure}
\centering
\includegraphics[height=2.75cm]{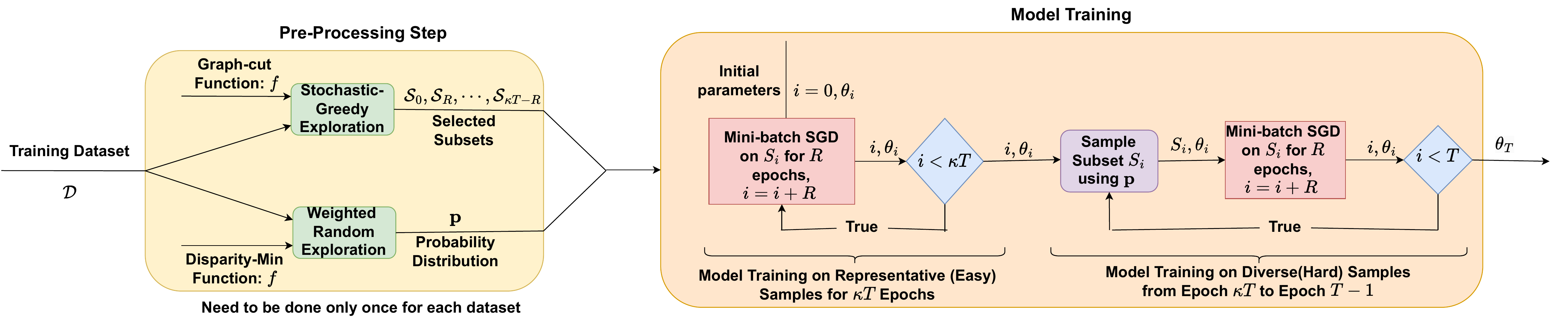}
\caption{Block diagram of \model{} for model training.} 
\vspace{-4mm}
\label{fig:pipeline}
\end{figure}

\noindent \textbf{\model{} Framework: } In this study, we introduce \model{}, a model-agnostic subset selection framework designed for efficient model training and tuning. \model\ employs submodular measures~\citep{fujishige2005submodular, kaushal2021prism}, which capture higher-order interactions between data samples for subset selection. We utilize pre-trained large language models~\citep{languagemodels} and pre-trained vision transformers~\citep{visiontransformers} as feature encoders due to their zero-shot feature encoding capabilities. These encoders compute the sample metrics in a nominal amount of time, and this computation is independent of the downstream model, rendering \model\ model-agnostic. Figure~\ref{fig:pipeline} provides a visual representation of \model\ for model training, comprising two steps: a) A pre-processing step that involves the selection of multiple subsets from the training dataset using "Stochastic-Greedy Exploration (SGE)" (refer to Section~\ref{sec:stochastic_greedy}), and the construction of a probability distribution over the entire dataset for subset sampling through "Weighted Random Exploration (WRE)" (refer to Section~\ref{sec:weighted_random}); b) A model training scheme that involves training the model on a curriculum of easy-to-hard subsets (refer to Section~\ref{sec:curriculum}) using the subsets selected and the probability distribution constructed in the pre-processing step. We also present a class-wise partitioning trick to significantly minimize the memory footprint of \model\ (refer to Section~\ref{sec:implementation}).


 \begin{figure}[!ht]
\centering
\begin{subfigure}[b]{0.63 \textwidth}
\centering
\includegraphics[width=4cm]{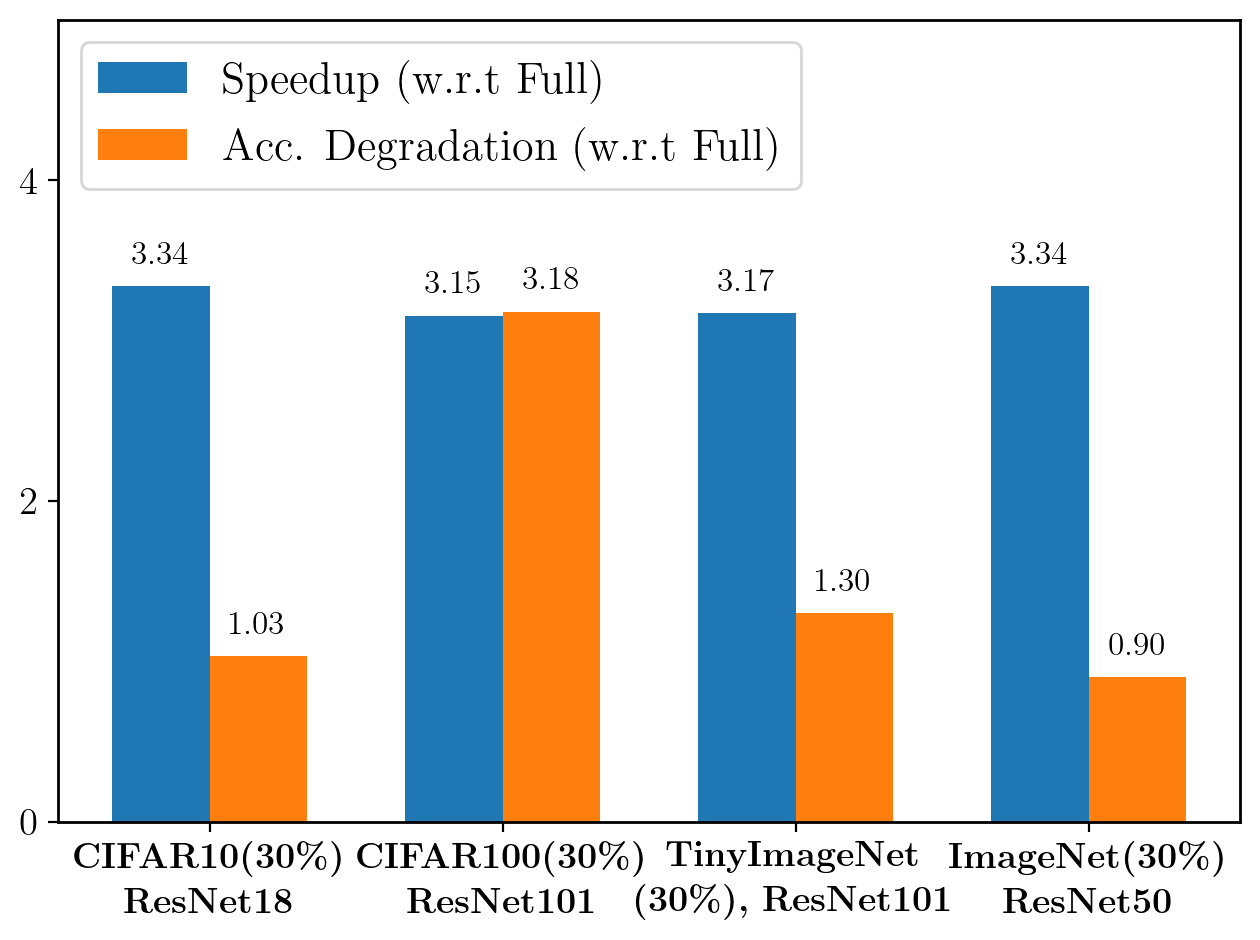}
\includegraphics[width=4cm]{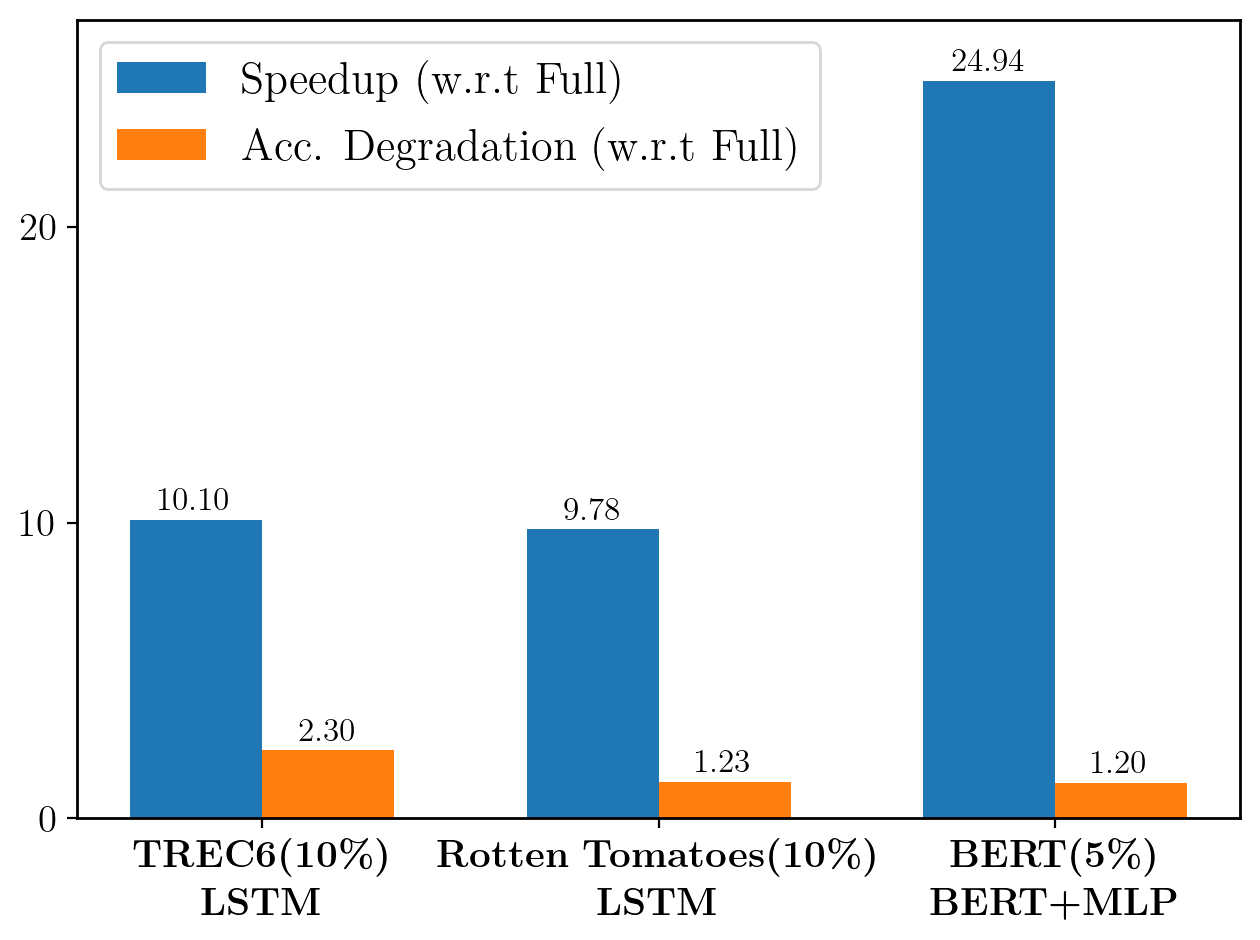}
\caption*{$\underbracket[1pt][1.0mm]{\hspace{8.5cm}}_{\substack{\vspace{-4.0mm}\\
\colorbox{white}{\scriptsize (a) Efficient Training}}}$}
\phantomcaption
\label{fig:main-eff}
\end{subfigure}\quad
\begin{subfigure}[b]{0.3 \textwidth}
\centering
\includegraphics[width=4cm]{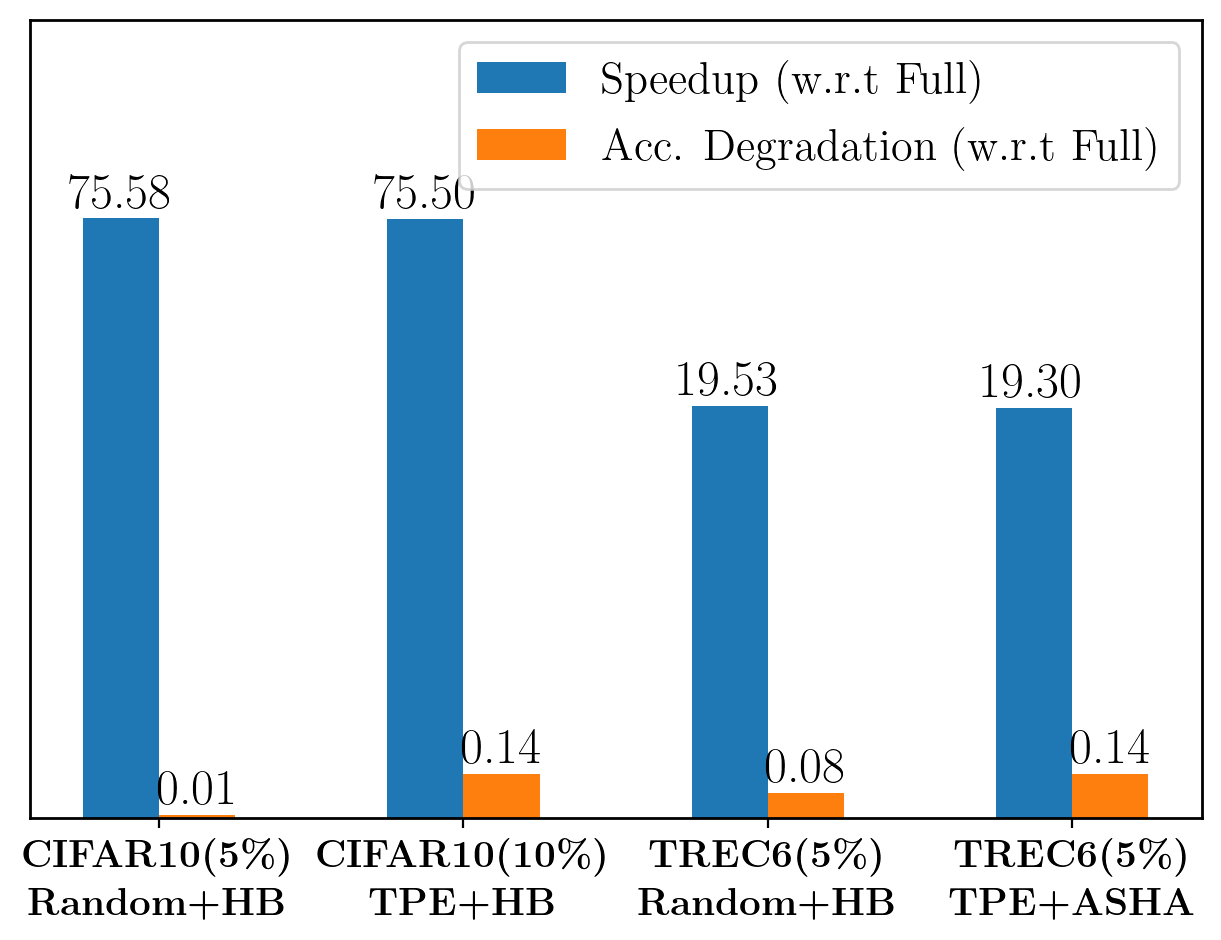} 
\caption*{$\underbracket[1pt][1.0mm]{\hspace{4.25cm}}_{\substack{\vspace{-4.0mm}\\
\colorbox{white}{\scriptsize (b) Efficient Tuning}}}$}
\phantomcaption
\label{fig:main-rob}
\end{subfigure}\quad
\caption{\footnotesize{This figure illustrates the \model{}'s tradeoff between speedup and accuracy degradation compared to full data training and tuning. For model training, speedups of $3\times$ to $10\times$ were achieved with less than a $1.5\%$ accuracy drop, and for hyper-parameter tuning, \model{} achieves speedups of $20\times$ to $75\times$ with less than a $0.15\%$ accuracy drop.}}
\vspace{-5mm}
\label{fig:mainresults}
\end{figure}


\noindent \textbf{Effectiveness of \model\ : }Through extensive experiments on multiple real-world datasets, we empirically demonstrate the effectiveness of the \model\ framework for efficient training and hyperparameter tuning. In \figref{fig:mainresults}, we provide a summary of the speedup achieved by \model\ compared to full data training and tuning, along with the corresponding relative performance. Our results show that \model\ can train models 3x to 10x faster and tune hyperparameters 20x to 75x faster, with minimal loss in performance. Furthermore, we consistently outperform the subset selection baselines that were considered. Additionally, \model\ exhibits significantly faster initial model convergence and maintains faster convergence throughout training compared to other baselines. These qualities make it highly effective for applications such as hyperparameter tuning and neural architecture search, where the ability to distinguish good models from bad models early in the tuning process is crucial for enhancing efficiency \cite{hyperband, asha}.

\paragraph{Related Work: } In recent years, data-subset selection strategies have achieved success in various machine learning applications such as speech recognition~\citep{wei2014unsupervised,wei2014submodular}, machine translation~\citep{kirchhoff2014submodularity}, active learning~\citep{sener2018active, ash2020deep, kothawade2021submodular}, hyper-parameter tuning~\citep{killamsetty2022automata}, continual learning~\citep{Tiwari2021GCRGC}, domain adaptation~\citep{karanam2022orient}, and computer vision~\citep{kaushal2019learning}. A recent empirical study~\citep{birodkar2019semantic} demonstrated that datasets frequently contain semantic redundancies. These redundancies can be removed during training without affecting performance. In response, data pruning methods~\citep{toneva2018an, EL2N, sorscher2022beyond} have developed principled selection criteria for pruning redundant samples before model training, with minimal loss in performance. However, both existing data pruning and compute-efficient learning approaches are model-dependent, resulting in them being computationally expensive. In contrast, our method is model-agnostic and capable of identifying high-quality subsets by intelligent data exploration. Furthermore, our approach can achieve superior model convergence by utilizing an easy-to-hard curriculum, transitioning from representative to diverse subsets during training.

\section{Preliminaries}
\noindent \textbf{Notation: }We briefly describe the notation for various variables that will be used throughout the remainder of this work. Denote the training dataset as $\gD = \{(x_j, y_j)\}_{j=1}^m$ with $m$ data points. Let $\gS$ be the subset of the training dataset on which the downstream model is trained. Let the feature encoder be denoted as $g: X \rightarrow Z$, which transforms the input from the feature space $X$ to an embedding space $Z$. Let the downstream model parameters be characterized by $\vtheta$. We subscript the changing variables, such as model parameters $\vtheta$ and subset $\gS$, with the timestep $t$ to denote their specific values at that timestep. Finally, let the subset size be denoted as $k$.

\noindent \textbf{Subset Selection Formulation: }The standard subset selection problem can be formulated as the maximization of a set function $f$ subject to a budget constraint $k$:
\begin{align}
\small
    \gS^* = \underset{ \gS : \gS \subseteq \gD, |\gS| = k}{\argmax} f(\gS)
    \label{eq:subset_formulation}
\end{align}

\noindent \textbf{Submodular Functions: } Given a set function $f:2^{\gD} \rightarrow \R$ that maps sets to real values. For a set $\gA \subseteq \gD$, $f(\gA)$ gives a real-valued score for $\gA$. Formally, a function $f$ is defined as submodular \citep{fujishige2005submodular} if, for any $x \in \gU$, the inequality $f(\gA \cup x) - f(\gA )\geq f(\gB \cup x) - f(\gB)$ holds for all $\gA \subseteq \gB \subseteq \gD$ and $x \notin \gB$. Submodular gain, $f(x|\gD)$, is defined as the difference between the set function value of the set $\gD \cup x$ and the set function value of set $\gD$, i.e., $f(x|\gD) = f(\gD \cup x) - f(\gD)$. A function $f$ is considered monotone if $f(\gA) \leq f(\gB)$ for any $\gA \subseteq \gB$. Although most general discrete maximization problems are NP-hard, they can be approximately solved if the set function $f$ being maximized is submodular. Furthermore, if the set function $f$ is monotone submodular, then the optimization problem in Equation~\ref{eq:subset_formulation} can be approximated to a factor of $1-\frac{1}{e}$~\citep{nemhauser1978analysis} when maximized under a cardinality constraint using a simple greedy algorithm.
 
\section{Development of \model{}} \label{sec:methodology}
In this section, we discuss the theoretical and empirical considerations that inspired the development of \model{}. Our primary objective for this new subset selection approach is to enable the selection of a new subset in a negligible amount of time while ensuring superior model convergence and performance within a specified period.

\paragraph{Choice of Feature Encoders:}The standard subset selection problem as outlined in \eqref{eq:subset_formulation} involves maximizing the set function $f$ subject to a budget constraint. To capture interactions between data samples, most set functions $f$ necessitate the computation of a similarity kernel $\gK$~\citep{kaushal2021prism}. This computation requires informative encodings of samples. Our initial design choice was to utilize pre-existing pre-trained language models or vision transformers as feature encoders $g$. These models enhance contextualization, expressiveness, and generalizability, promoting extrapolation in zero-shot settings~\citep{languagemodels, visiontransformers}. However, this prompts questions regarding their ability to effectively generalize to specialized domain datasets. We address this issue empirically in Appendix~\ref{app:specialized}, which shows that the pre-trained transformer models used in our study can generalize effectively to such domains. In cases where a pre-trained model underperforms on a specific dataset—determined by linear probing accuracies—one can fine-tune the pre-trained transformer model or train a smaller proxy model~\cite{Coleman2020Selection}. Although this adds to pre-processing costs, it boosts performance, and pre-processing only needs to be done once per dataset. The efficacy of using a proxy model is validated in Appendix~\ref{app:proxy_model}. By using a pre-trained transformer model or a proxy model, the need for downstream machine-learning models to compute sample representations is eliminated. Further, the effectiveness of various language models or vision transformers as feature encoders for subset selection is assessed in Appendix~\ref{app:feature_encoders}.

\begin{wrapfigure}{L}{0.45\textwidth}
 \vspace{-3mm}
  \centering
  \begin{subfigure}[b]{0.22\textwidth}
    \includegraphics[width=\linewidth, height=3cm]{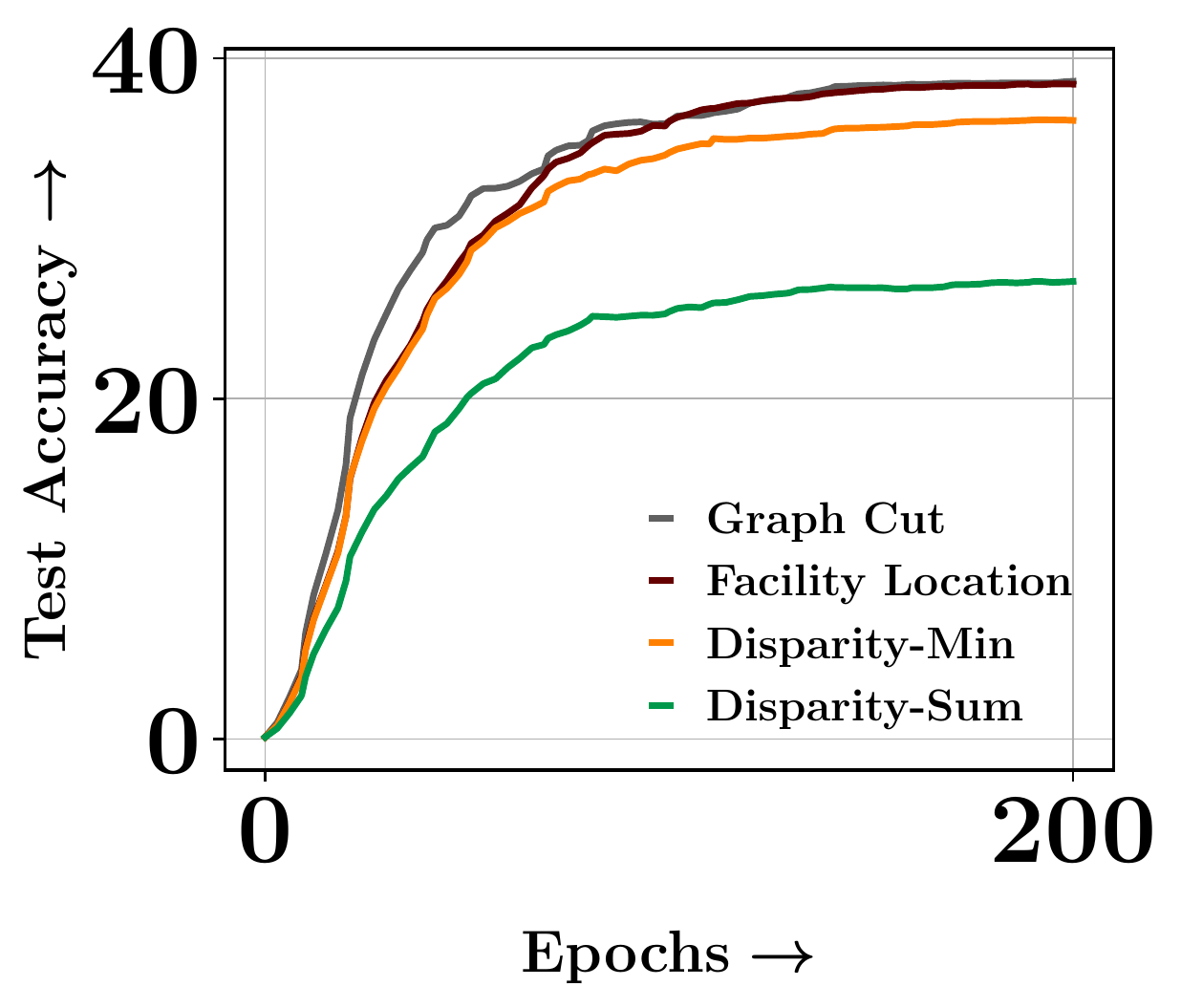}
    \caption{\textsc{Cifar100} (10\%)}
    \label{subfig:0.1_submod_comparison}
  \end{subfigure}
  \hfill 
  \begin{subfigure}[b]{0.22\textwidth}
    \includegraphics[width=\linewidth, height=3cm]{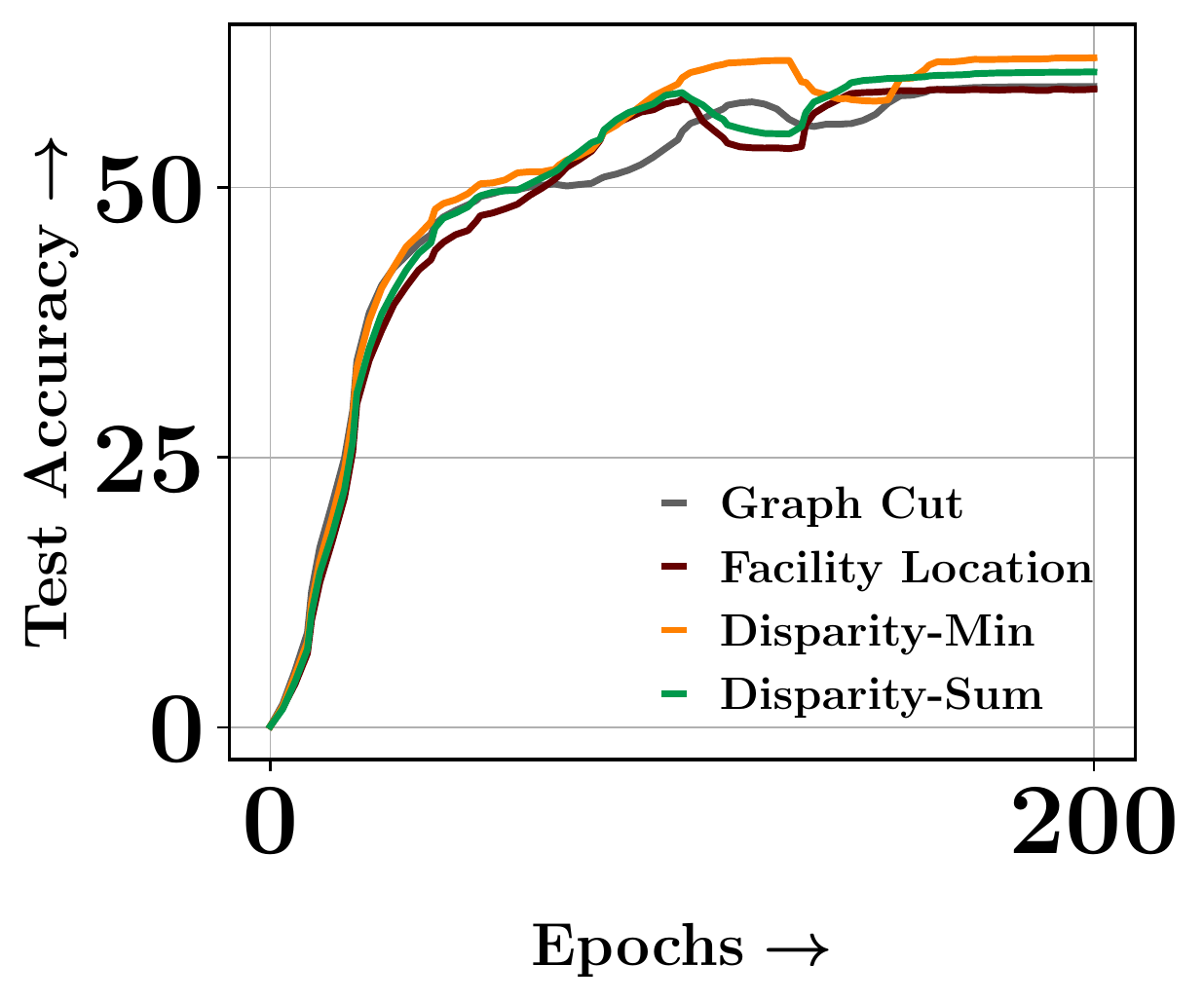}
    \caption{\textsc{Cifar100} (30\%)}
    \label{subfig:0.3_submod_comparison}
  \end{subfigure}
  \caption{\footnotesize{
ResNet18 model performance on 10\% and 30\% subsets of the \textsc{Cifar100} dataset, selected using diverse set functions for maximization.}}
  \vspace{-3mm}
  \label{fig:submod_comparison}
\end{wrapfigure}

\paragraph{Optimal Subset Composition:} The immediate question that arises is the choice of the appropriate set function for subset selection. Essentially, we need to pinpoint the key characteristics a subset should possess for optimal model performance. To identify these characteristics, we assess the empirical performance of various set functions that capture the following subset qualities: \emph{Representation:} This attribute measures the subset's effectiveness in representing the entire dataset. For instance, it ensures that the subset includes more samples from dense regions than sparse ones, typically resulting in "easier" data samples. We used facility location and graph-cut set functions to model representation; \emph{Diversity:} This characteristic evaluates the degree of distinctiveness among the data samples within the subset. For instance, it ensures that the subset covers highly diverse samples, which are generally "harder" data samples. To model diversity, we considered the disparity-sum and disparity-min set functions. We empirically support the claim that representation-based functions select easier samples, and diversity-based functions select harder samples by presenting the mean EL2N scores \citep{EL2N} for the subsets chosen by their respective functions in Appendix \ref{app:el2n}. All set functions considered, except disparity-sum and disparity-min, are submodular. Despite the non-submodular nature of disparity-min and disparity-sum functions, it has been established that they can be effectively maximized using the conventional greedy approach, leading to 1/4 and 1/2 approximations respectively \citep{dasgupta-etal-2013-summarization}. Therefore, we have deemed it appropriate to include these functions in our evaluation. We provide instantiations of the considered set functions in Appendix \ref{app:submodfuncs}. Figure \ref{fig:submod_comparison} demonstrates the performance of a ResNet18 model trained on 10\% and 30\%  fixed subsets of the \textsc{Cifar100} dataset, respectively, by maximizing different set functions. The results show that when using larger subset sizes of 30\% or more, subsets selected with diversity functions disparity-min and disparity-sum lead to superior model performance. Conversely, for smaller subset sizes of 10\% or less, subsets chosen with representation functions graph-cut and facility location yield better model performance. This observation is consistent with the recent findings of \citet{sorscher2022beyond}.

\paragraph{Issue with using fixed data subsets:}If the goal is to achieve the best performance within a certain timeframe, the model must also engage in data exploration instead of relying solely on a fixed data subset. One significant disadvantage of training models using fixed data subsets is the requirement for large subsets, approximately 70\% or more, to achieve accuracy comparable to full data training. This leads to extended training times. For instance, when training the ResNet101 model on a fixed 10\% random subset of the \textsc{Cifar10} dataset for 200 epochs, it only reached a test accuracy of 66.9\%. Conversely, the ResNet101 model achieved a test accuracy of 87.54\% when trained on an adaptive 10\% subset of \textsc{Cifar10} data for the same number of epochs, with a new subset being randomly selected after each epoch. While random data exploration is a simple and empirically effective method for data exploration, it may not be the most efficient strategy due to potential redundancy in the randomly selected subsets. Therefore, it is vital to develop a strategy that balances \emph{Subset Exploration and Subset Exploitation}.

\subsection{Informative Data Exploration}
To balance exploration and exploitation, models should train on small, informative subsets that encourage exploration. A performant subset selection approach would combine training samples from all parts of the dataset without undue bias. An ideal formulation for informative data subset exploration is as follows:
\begin{align}
\small
    P(\gS) \,\propto\, \exp{\left(\beta\!\cdot\!f(\gS)\right)}\;\;\mathrm{subject}\;\mathrm{to}\;\;|\gS| = k
    \label{eq:data_exploration}
\end{align}
In \eqref{eq:data_exploration}, $P(\gS)$ represents the probability of sampling subset~$\gS$, $f$ is our set quality measure, and ${\beta > 0}$ is the inverse temperature~\citep{gotovos2015sampling}. This formula ensures higher quality subsets are explored more frequently, without showing bias in the points sampled. Ideally, a new subset is selected for every epoch from this probability distribution. However, constructing a simple sampler according to \eqref{eq:data_exploration} requires a combinatorial number of set function evaluations. Although \citet{gotovos2015sampling} proposed a Gibbs sampling approach for marginal inference with a polynomial mixing time in~$|\gD|$, to enforce ${|\gS| = k}$, it needs modification, involving swapping data points at each step. But this leads to considerable mixing times, especially when the current subset has a close-to-optimal $f(\gS)$ value. We propose two scalable alternatives for data exploration with varying exploration-to-exploitation ratios and leave the extension of \citet{gotovos2015sampling} to future work.

\subsubsection{Stochastic-Greedy Exploration (SGE)} \label{sec:stochastic_greedy}
The first method we use for data exploration involves identifying multiple subsets with high function values. We then train the downstream model on these selected subsets, changing the subsets every $R$ epochs. This approach emphasizes exploitation over exploration, as it primarily focuses on subsets with high function values. To select $n$ subsets $\gS_1, \gS_2, \cdots, \gS_n$ from the dataset $\gD$ with high set function values, we employ the stochastic greedy algorithm~\citep{mirzasoleiman2015lazier} to maximize the set function $f$ and repeat the maximization $n$ times.

\begin{align}
\small
    \gS_1, \gS_2, \cdots, \gS_n \xleftarrow{} \operatorname{SGE}(f, \gD, k)
    \label{eq:stochastic_data_exploration}
\end{align} 

The randomness of the stochastic greedy algorithm allows us to choose a different subset with an approximate guarantee of $\gO(1 - \frac{1}{e} - \epsilon) $ every time. Due to space constraints, a detailed pseudocode of the "SGE" is provided in Algorithm~\ref{alg:sge} in Appendix~\ref{app:pseudocode}. In our experiments, we use an $\epsilon$ value of 0.01 for stochastic greedy maximization.

\subsubsection{Weighted Random Exploration (WRE)} \label{sec:weighted_random}
In the WRE approach, we explore data by creating a multinomial probability distribution $\vp$ across the entire dataset $\gD$. Every $R$ epochs, we sample a subset $\gS$ of size $k$ from this distribution, without replacement. We employ a weighted random sampling method~\citep{Efraimidis2016}, whereby each data sample is assigned a weight equal to its normalized set function gain during greedy maximization. Specifically, we greedily maximize the set function $f$ over the whole dataset $\gD$ and designate the set function gains associated with each data sample $e$ at its point of greedy inclusion as its importance score $g_e$. Here, if $\gS$ is the subset selected thus far and $e$ is the next optimal data sample to be added greedily, the set function gain value of $e$ is computed as $f(\gS \cup e) - f(\gS)$.

\begin{align}
\small
    \vg = [g_1, g_2, \cdots, g_m] \xleftarrow{} \operatorname{GreedySampleImportance}(f, \gD)
    \label{eq:importance_scores}
\end{align}

As illustrated in \eqref{eq:taylor_softmax}, we normalize the importance scores $\vg$ and construct the probability distribution $\vp$ over the training set $\gD$ by employing the second-order Taylor-Softmax function~\citep{taylor-softmax} over the importance scores.

\begin{align}
\small
    \vp &= \operatorname{Taylor-Softmax}(\vg) = \Big[\frac{1 + g_i + 0.5 g_i^2}{\sum_{j = 1}^{m}1 + g_j + 0.5 g_j^2}\Big]_{i=1}^{m} 
    \label{eq:taylor_softmax}
\end{align}
When $f$ is submodular, the diminishing returns property of submodular functions means elements chosen in early iterations have greater set function gain than those selected later. The probability distribution $\vp$ generated assigns higher probability to more informative samples. Depending on the set function, this could mean better representativeness or diversity. Importantly, sampling from $\vp$ allows the exploration of less informative samples while frequently selecting more informative ones. Once $\vp$ is established, selecting new subsets from the multinomial distribution is as quick as random subset selection. We use $\vp$ to sample new subsets of size $k$ every $R$ epochs (without replacement). For the detailed pseudocode of the greedy sample importance estimation, please refer to Algorithm~\ref{alg:greedyimp} in Appendix~\ref{app:pseudocode}, as space constraints prevent inclusion here.


\begin{wrapfigure}{l}{0.5\textwidth}
  \vspace{-4.5mm}
  \centering
  \begin{subfigure}[b]{0.26\textwidth}
    \includegraphics[width=\linewidth]{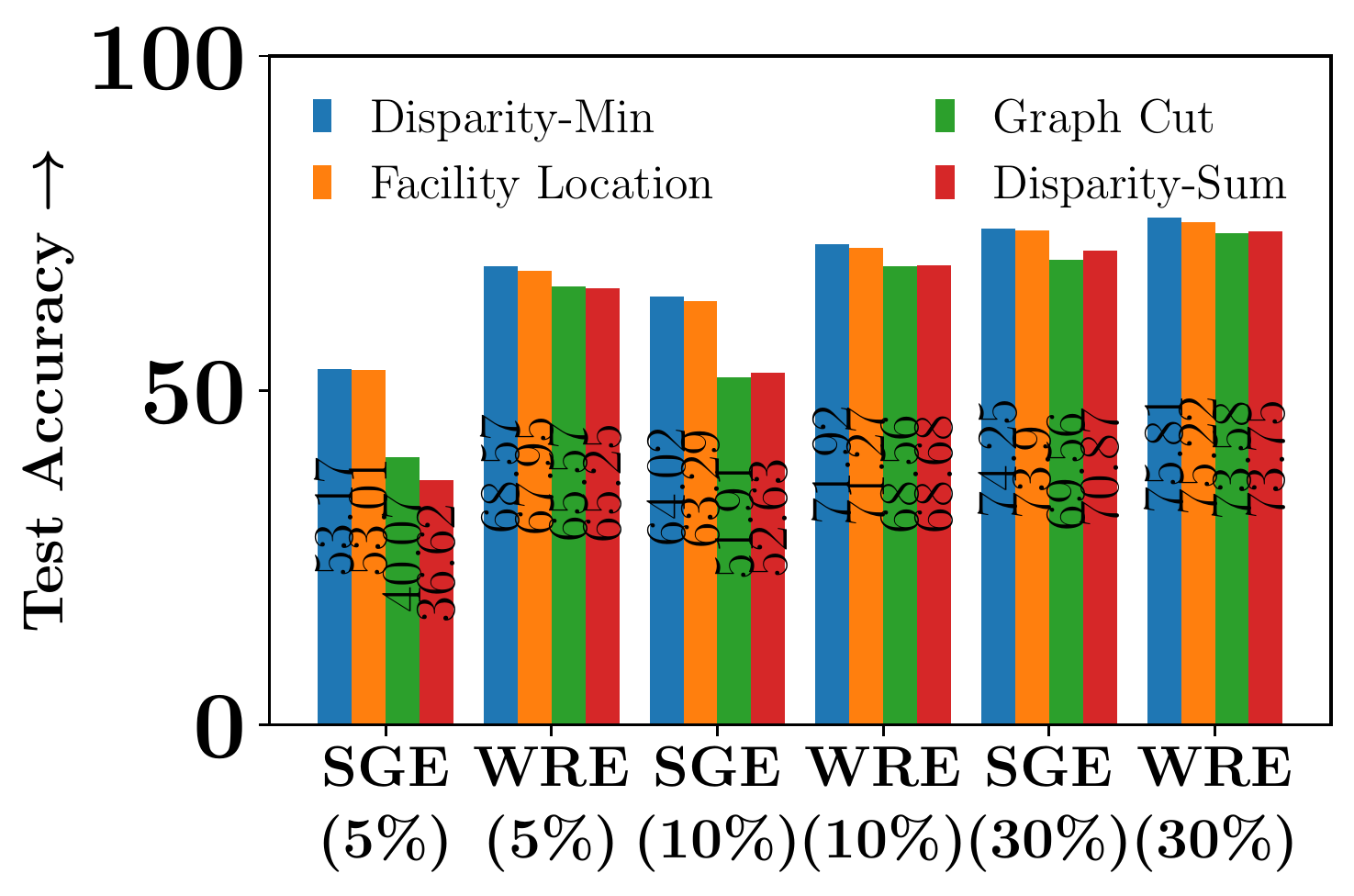}
    \caption{\footnotesize Comparison of data exploration approaches}
    \label{subfig:data_exploration_comp}
  \end{subfigure}
  \hfill 
  \begin{subfigure}[b]{0.21\textwidth}
    \includegraphics[width=\linewidth]{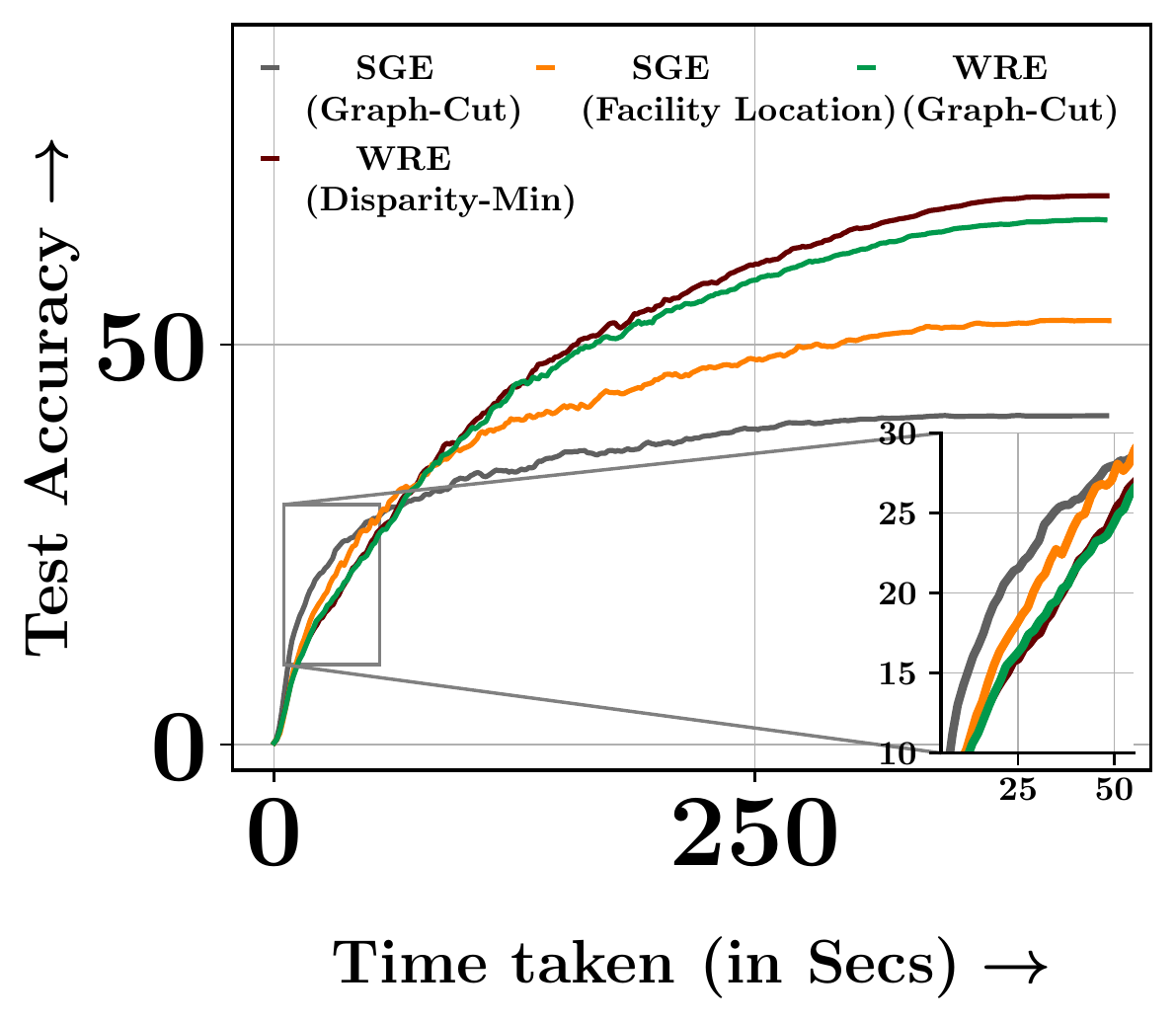}
    \caption{\footnotesize ResNet18 convergence on 5\% \textsc{Cifar100} subset}
    \label{subfig:data_exploration_convergence}
  \end{subfigure}
  \caption{\footnotesize{Figure (a) illustrates the performance of the ResNet18 model on various subset sizes and set functions using SGE and WRE approaches on \textsc{Cifar100}. Figure (b) demonstrates the convergence of the model on a 5\% subset of \textsc{Cifar100} employing SGE with Graph Cut and WRE with Disparity-Min function.}}
  \vspace{-3mm}
  \label{fig:data_exploration_submod_comp}
\end{wrapfigure}

\noindent \textbf{SGE vs. WRE: }Sub-figure \ref{subfig:data_exploration_comp} demonstrates that training the ResNet18 model using the Weighted Random Exploration (WRE) approach, which emphasizes exploration, yields better performance than both the Stochastic-Greedy Exploration (SGE) approach, which prioritizes exploitation, and fixed data subsets (pure exploitation). Furthermore, using Disparity-Min as a set function outperforms other set functions in both exploration strategies. However, we made an important empirical observation: SGE with the Graph-cut function achieves superior initial model convergence in the early iterations, as shown in sub-figure \ref{subfig:data_exploration_convergence}. This sub-figure emphasizes the benefit of early-stage model convergence of SGE using Graph-cut, focusing on easier samples, compared to WRE with Disparity-Min, focusing on more challenging samples, and both WRE with Graph-cut and SGE using Facility Location, targeting easier samples, on the \textsc{Cifar100} dataset. Early convergence is particularly beneficial in hyper-parameter tuning or neural-architecture search, where early-stage performance assessment is critical. This trend of superior early convergence with SGE using Graph-cut is consistent across multiple datasets and subset sizes, as shown in Figures \ref{fig:sge_gc_vs_sge_fl} and \ref{fig:sge_gc_vs_wre_gc} in the Appendix. We provide a detailed explanation in Appendix sections \ref{app:sge_gc_vs_sge_fl} and \ref{app:sge_gc_vs_wre_gc}, explaining why SGE with Graph-cut results in superior initial convergence compared to SGE with Facility Location and WRE with Graph-cut, even though all these approaches aim to select easy or representative samples.

\subsubsection{Developing a Curriculum of Easy-to-Hard Subsets: } \label{sec:curriculum}
Sub-figure \ref{subfig:data_exploration_convergence} suggests that for optimal model convergence with \model\ throughout training, we should construct an easy-to-hard curriculum, building on the empirical success of such approaches \citep{learningeasythingsfirst, powerofcurriculumlearning, dynamicinstancehardness}.
This is accomplished by initially using SGE with graph-cut, followed by WRE with disparity-min in later iterations. The curriculum design involves training the model for a fraction $\kappa$ of the total number of epochs with SGE and the graph-cut function, followed by WRE with disparity-min for the remaining epochs. The hyper-parameter $\kappa$ specifies the fraction of epochs dedicated to stochastic exploration with the graph-cut function. Our experiments determined that setting $\kappa = \frac{1}{6}$ yielded optimal results following hyper-parameter tuning. The use of disparity-min in WRE ensures the selection of subsets that include both easy and difficult samples, with a higher probability of selecting difficult samples, and mitigates the model's catastrophic forgetting of easy samples. The results of hyper-parameter tuning for $\kappa$ are presented in Appendix \ref{app:kappa_tuning}. Furthermore, we demonstrate the advantage of curriculum-based data exploration in enhancing model convergence and performance through an ablation study, as detailed in Appendix \ref{app:curriculum_ablation}.

\subsection{\model{} Framework}\label{sec:implementation}
As mentioned earlier, the \model{} training process follows a curriculum of subsets, transitioning from representative to diverse subsets. Figure~\ref{fig:pipeline} illustrates the \model{} training pipeline, while Figure~\ref{fig:hp_pipeline} in the Appendix depicts the hyper-parameter tuning pipeline with \model{}. Detailed pseudocode for the \model{} algorithm can be found in Algorithm~\ref{alg:mainalg} in Appendix~\ref{app:pseudocode}, as space constraints preclude its inclusion in the main paper. For hyper-parameter search and scheduling algorithms, we used Wandb~\citep{wandb}, SUBMODLIB~\citep{SUBMODLIB} for submodular optimization, and CORDS~\citep{Killamsetty_CORDS_COResets_and_2021} for baseline subset selection methods.

\paragraph{Class-wise Data Partitioning:} The set functions we experiment with in this study necessitate a similarity kernel $\gK$ of size $m \times m$, which must be computed over the entire dataset of size $m$. This could lead to substantial memory requirements for computation and storage as the dataset size $m$ increases, making it computationally prohibitive in some instances. To address this issue, we partition the data based on class labels and select multiple subsets from each class (for SGE) or create a probability distribution over each class (for WRE), considering only data instances from each class separately. For example, in a balanced dataset of size $m$ with $c$ classes, class-wise partitioning reduces memory requirements by a factor of $c^2$. By default, we use class-wise partitioning and curriculum-based data exploration with \model{}.

\section{Experimental Results}
\label{sec:exp_results}
Our experimental objective is to emphasize the efficiency and stability of \model{} in both model training and hyper-parameter tuning. Each experiment is repeated five times, and for clarity, we present only the mean test accuracies in our plots. A comprehensive table, including both the mean test accuracy and standard deviations, can be found in Appendices \ref{app:std_results} and \ref{app:hp_std_results}. To ensure a fair comparison, we use the same random seed for all trials across all methods. Implementation specifics, datasets, and baselines utilized in each scenario will be detailed in the following subsections.

\begin{figure}[!htbp]
\centering
\includegraphics[width = 13cm]{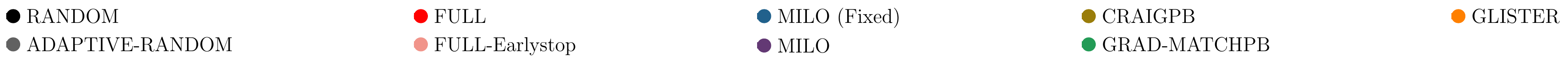}
\centering
\hspace{-0.6cm}
\begin{subfigure}[b]{0.225\textwidth}
\centering
\includegraphics[width=3.1cm]{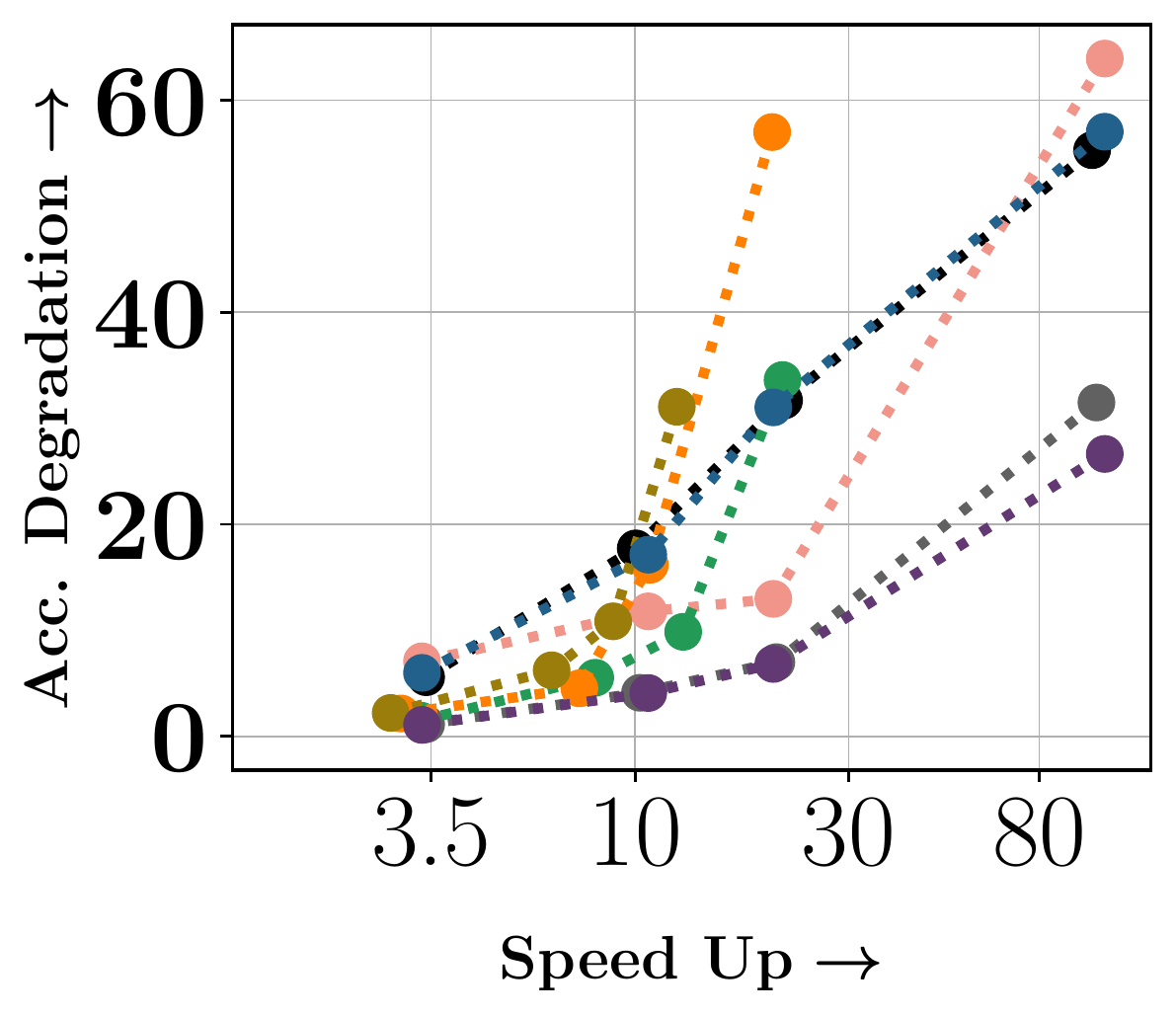}
\caption*{(a) \scriptsize CIFAR10 (ResNet18)}
\phantomcaption
\label{fig:CIFAR10_ResNet18}
\end{subfigure}
\begin{subfigure}[b]{0.225\textwidth}
\centering
\includegraphics[width=3.1cm]{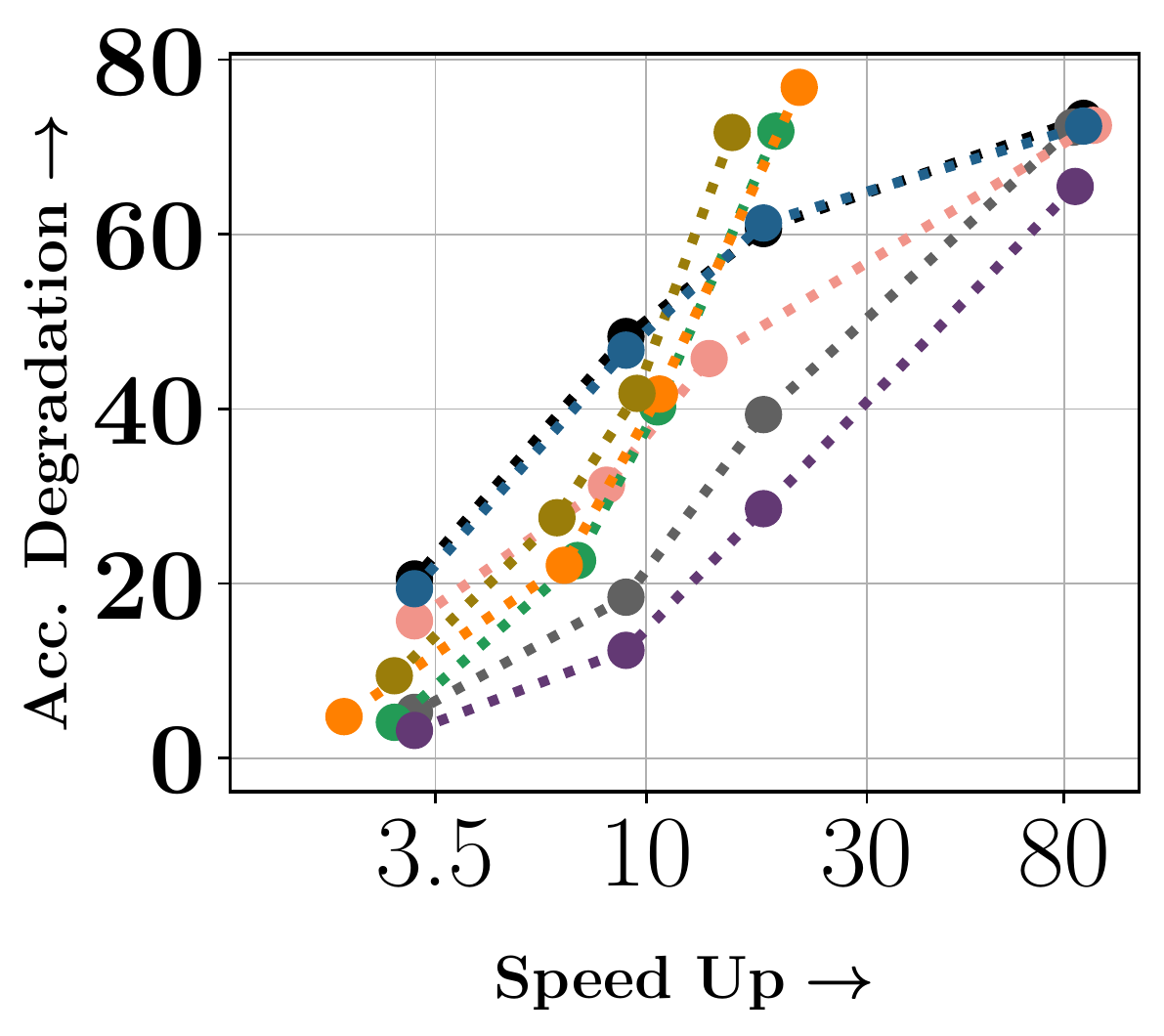}
\caption*{(b) \scriptsize CIFAR100 (ResNet101)}
\phantomcaption
\label{fig:CIFAR100_ResNet101}
\end{subfigure}
\begin{subfigure}[b]{0.225\textwidth}
\centering
\includegraphics[width=3.1cm]{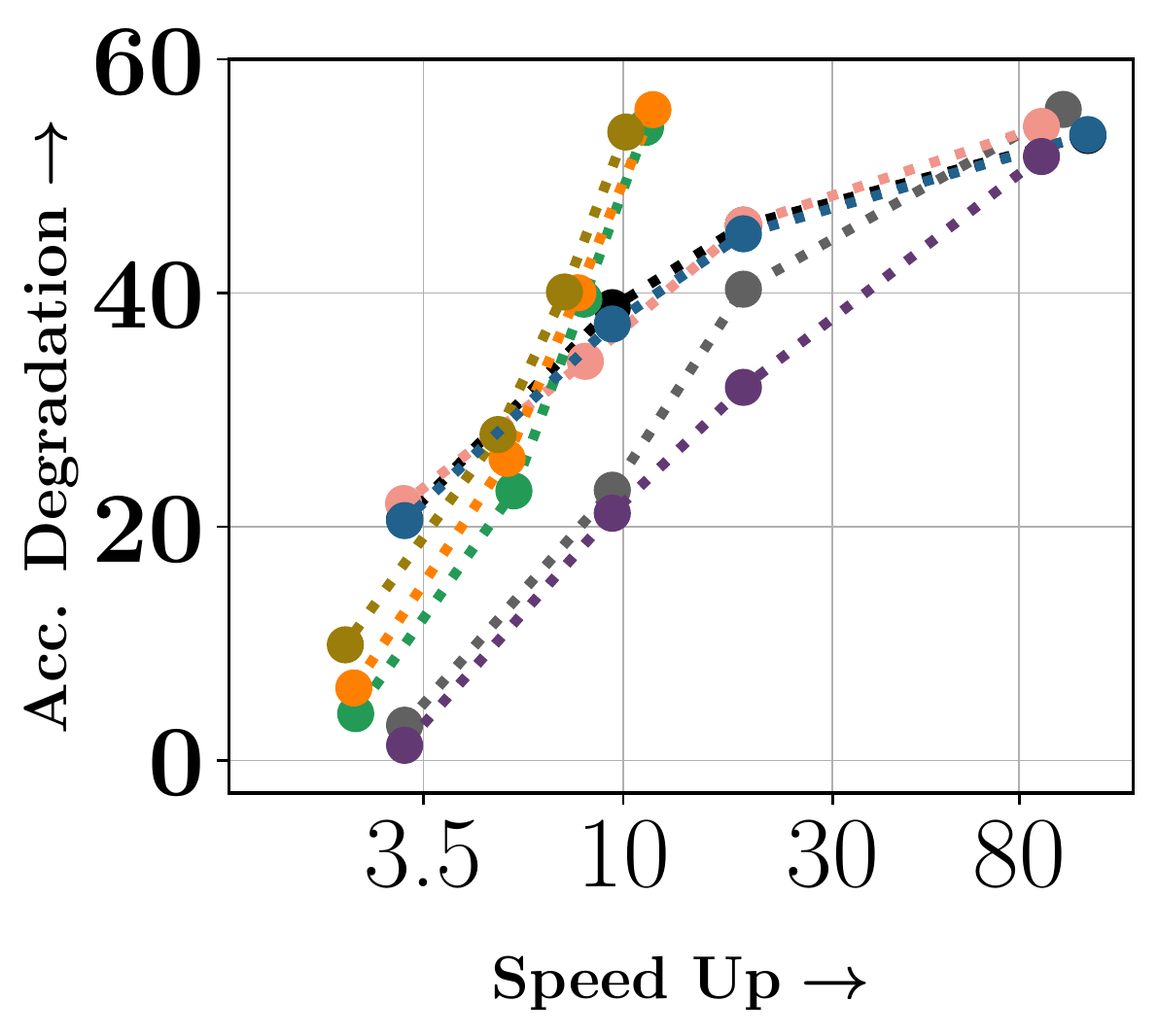}
\caption*{(c) \scriptsize TinyImagenet (ResNet101)}
\phantomcaption
\label{fig:tiny_imagenet_resnet101}
\end{subfigure}
\begin{subfigure}[b]{0.225\textwidth}
\centering
\includegraphics[width=3.1cm]{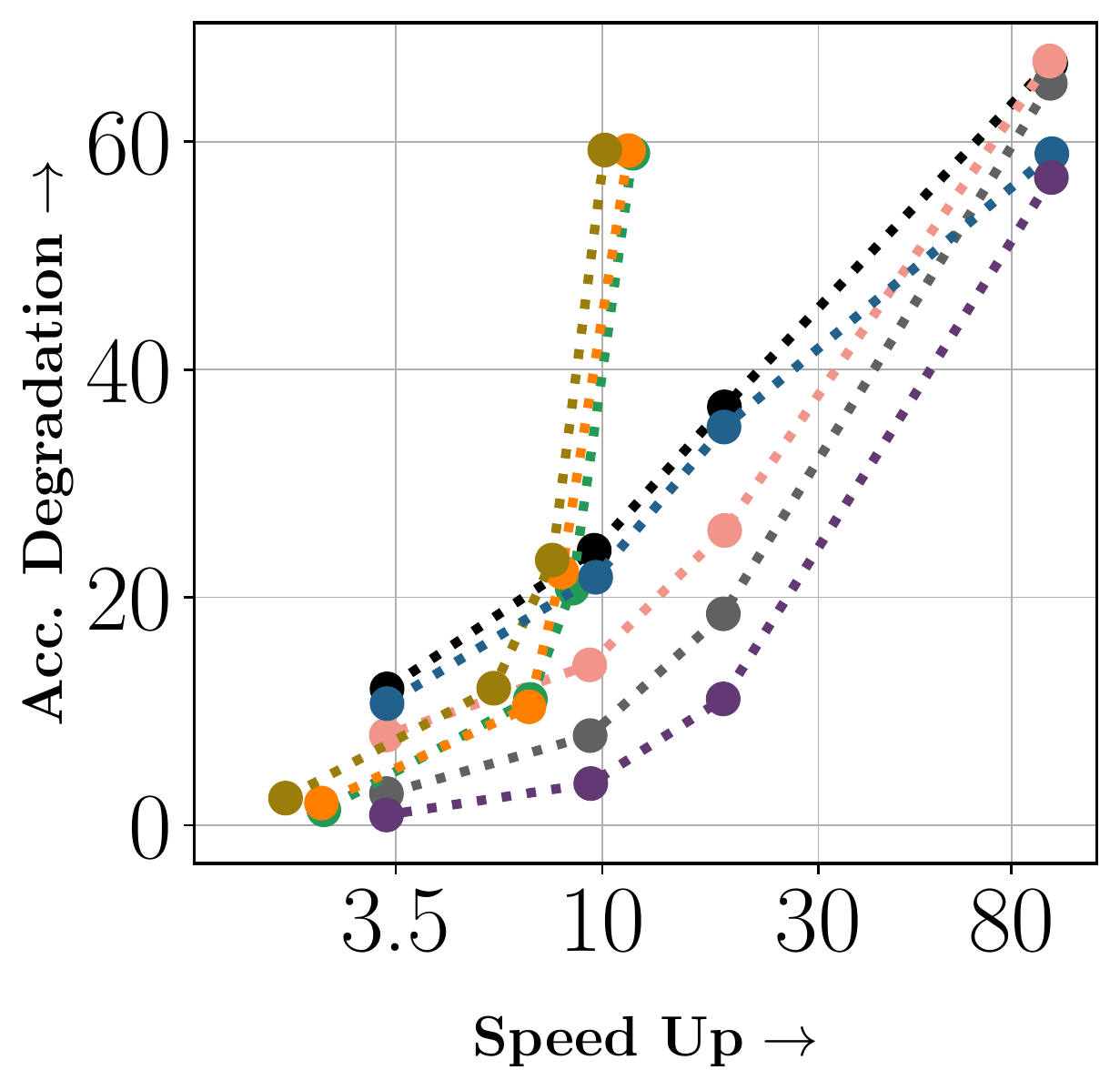}
\caption*{(d) \scriptsize ImageNet (ResNet50)}
\phantomcaption
\label{fig:imagenet_resnet50}
\end{subfigure}
\hspace{-0.6cm}
\begin{subfigure}[b]{0.225\textwidth}
\centering
\includegraphics[width=3.1cm]{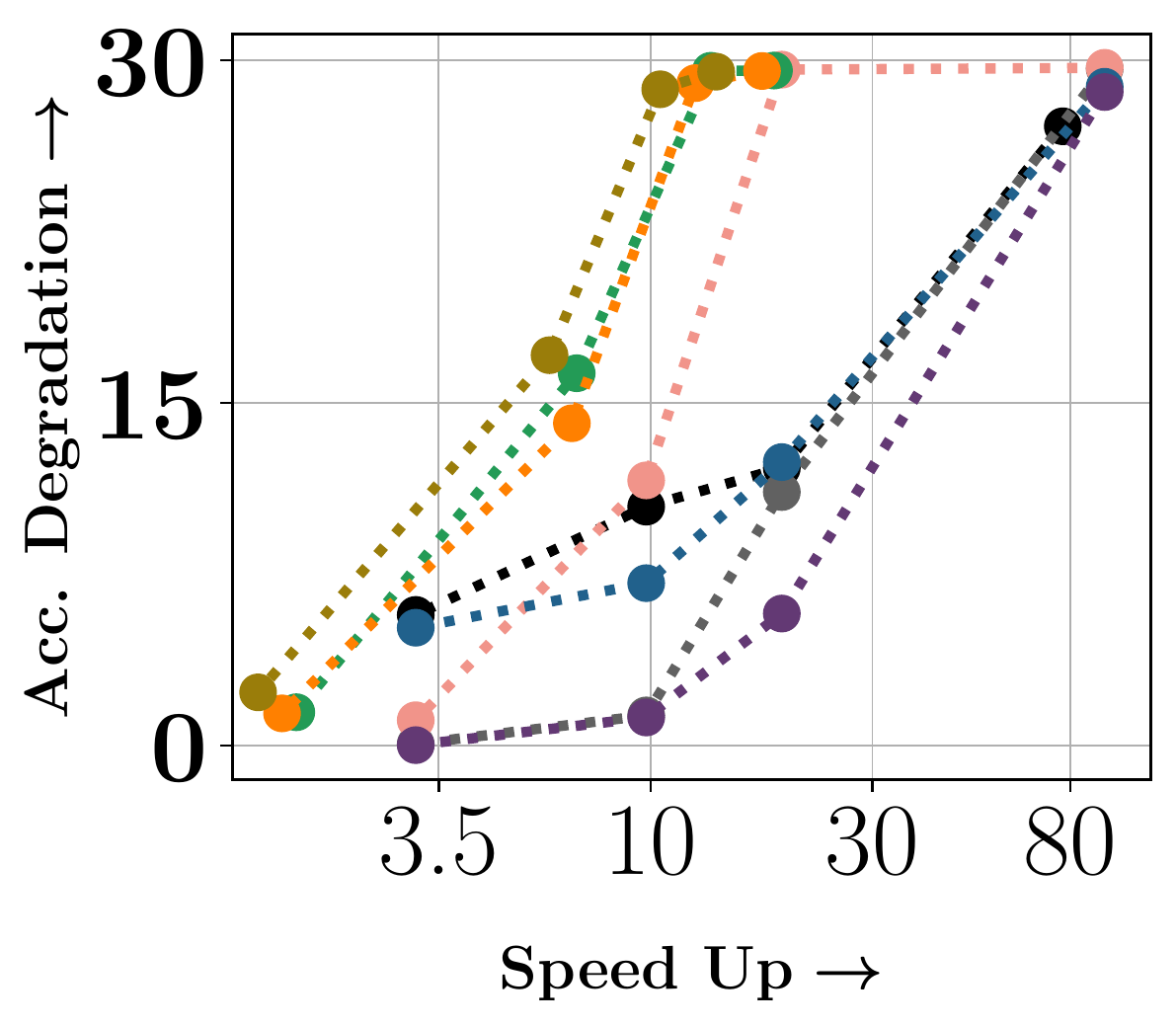}
\caption*{(e) \scriptsize Rotten Tomatoes (LSTM)}
\phantomcaption
\label{fig:rt_lstm}
\end{subfigure}
\begin{subfigure}[b]{0.23\textwidth}
\centering
\includegraphics[width=3.1cm]{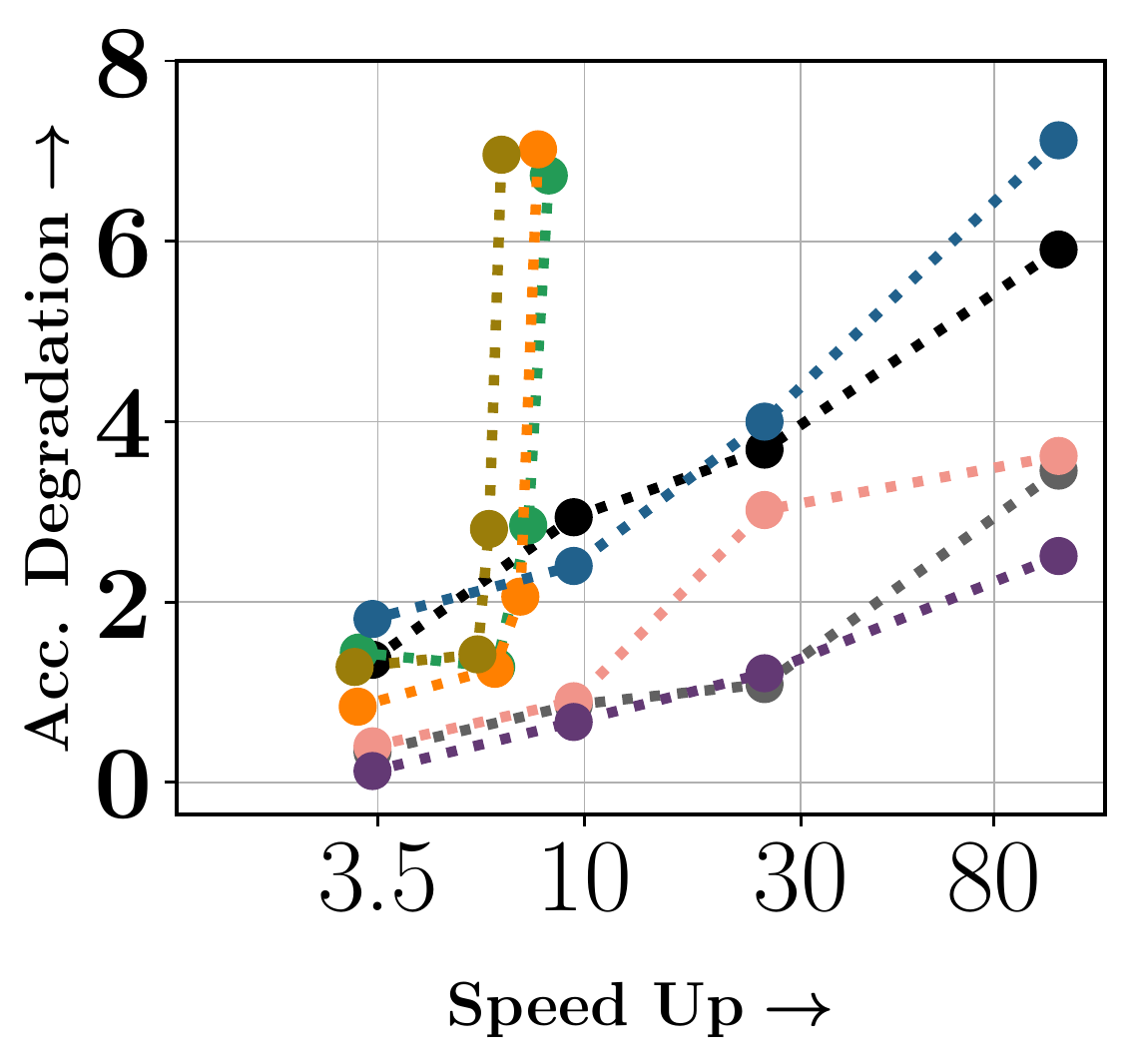}
\caption*{(f) \scriptsize IMDB (BERT+MLP)}
\phantomcaption
\label{fig:imdb_bert}
\end{subfigure}
\begin{subfigure}[b]{0.225\textwidth}
\centering
\includegraphics[width=3.1cm]{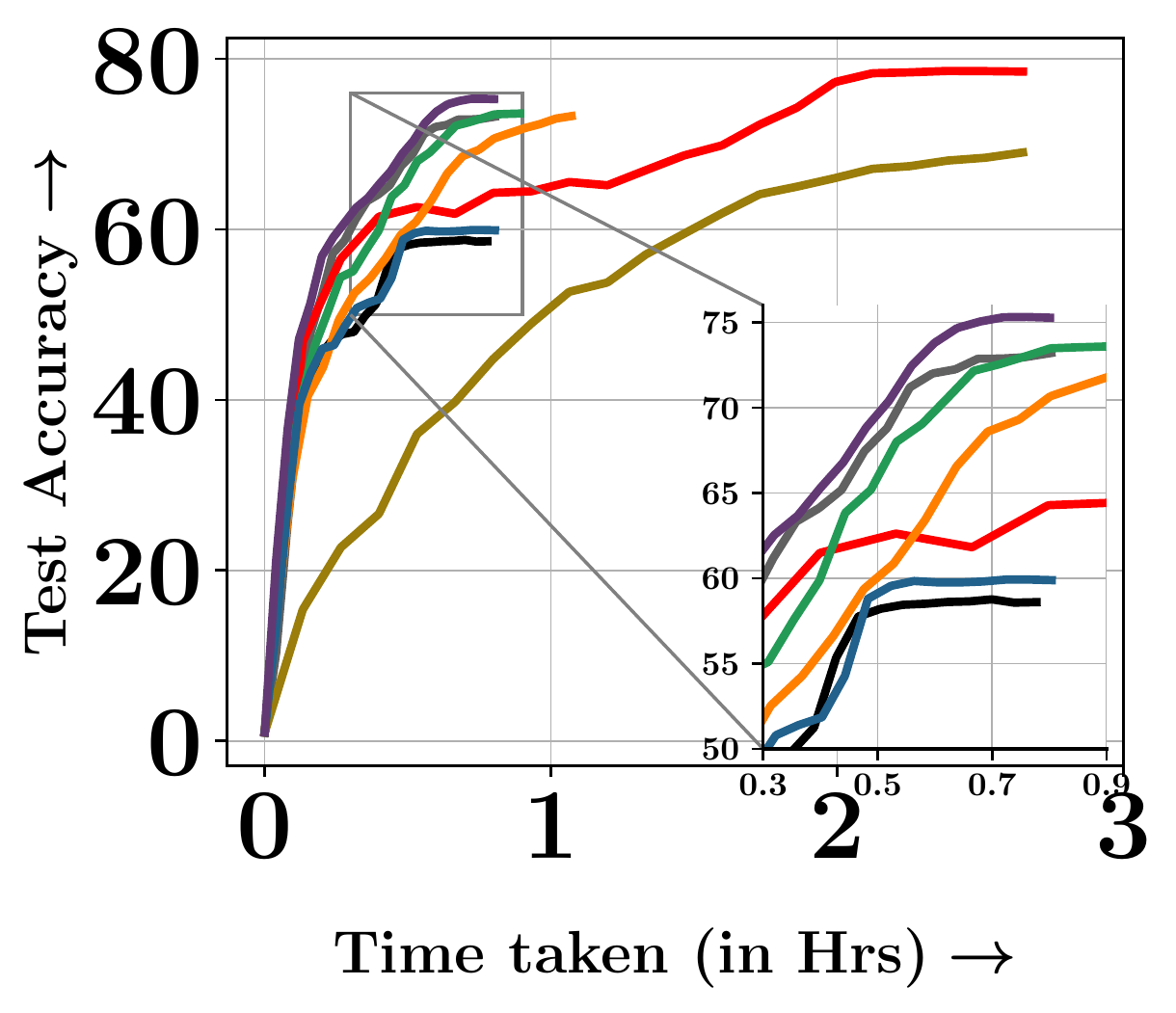}
\caption*{(g) \scriptsize CIFAR100 Convergence}
\phantomcaption
\label{fig:cifar100_resnet101_convergence}
\end{subfigure}
\begin{subfigure}[b]{0.225\textwidth}
\centering
\includegraphics[width=3.1cm]{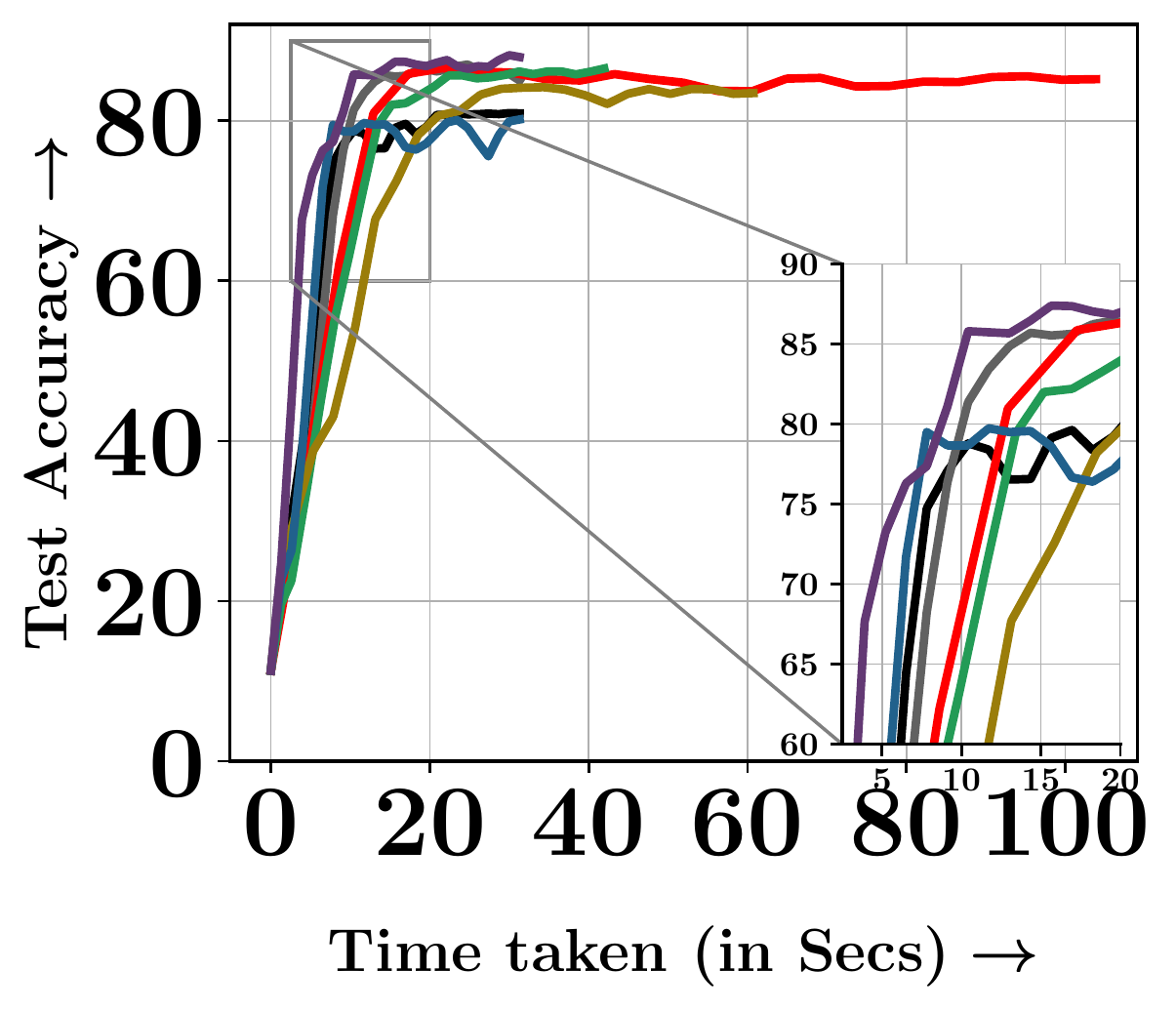}
\caption*{(h) \scriptsize Trec6 (LSTM) Convergence}
\phantomcaption
\label{fig:trec6_lstm_convergence}
\end{subfigure}
\caption{\footnotesize{Comparison of \model\ with baselines for model training using various subset sizes on seven different datasets with different models. The x-axis shows the accuracy degradation compared to full data training, and the y-axis shows the speedup achieved by using the subset selection approach. Smaller subsets are on the right, and larger ones are on the left. \model\ significantly outperforms existing baselines in terms of accuracy degradation and speedup tradeoff compared to full data training. \textbf{The bottom-right corner of each plot shows the best speedup-accuracy tradeoff region.} Plots (g) and (h) show the model convergence with time. \textit{Again, we see that \model\ achieves much faster convergence than all baselines and full training}.}}
\vspace{-6mm}
\label{fig:dss_general_experiments}
\end{figure}

\paragraph{Subset Selection Baselines:}Our experiments aim to showcase the performance of \model{} in model training and hyperparameter tuning scenarios. In \textit{single model training experiments}, we compare \model{} with several strategies. These include: \textsc{Random}, which randomly samples a subset of the same size as \model{} from the training data; \textsc{Adaptive-Random}, which adaptively samples a random subset of the same size as \model{} from the training data every $R$ epochs; \textsc{Full}, which uses the entire training data for model training and tuning; \textsc{Full-EarlyStop}, where early stopping is applied to full training to match the time or energy consumption of \model{}; and adaptive gradient-based subset selection strategies for efficient learning, where a new subset is selected every $R$ epochs. These include \textsc{CraigPB}, a faster per-batch version of \textsc{Craig}~\citep{mirzasoleiman2020coresets} as discussed in \citet{killamsetty2021grad}, \textsc{Glister}~\citep{killamsetty2021glister}, and \textsc{Grad-MatchPB}, the per-batch version of \textsc{Grad-Match}~\citep{killamsetty2021grad}. In \textit{hyper-parameter tuning experiments,} we adopt the experimental setup of \textsc{Automata}~\citep{killamsetty2022automata}, an efficient hyperparameter tuning framework that uses \textsc{Grad-MatchPB}, replacing the subset selection strategy with \model{}. To assess the efficacy of \model{} for hyperparameter tuning, we compare it with \textsc{Random}, \textsc{Full}, \textsc{Adaptive-Random}, and \textsc{Automata} (\textsc{Grad-MatchPB}) as subset selection baselines. We also present results from a variant of \model{}, denoted \textsc{Milo} (Fixed), which uses a fixed subset for model training, selected by maximizing the disparity-min function.

\noindent \textbf{Datasets, Model Architecture, and Experimental Setup: } 
In our experiments, we utilized various vision and text datasets, namely \textsc{Cifar100}, \textsc{Cifar10}~\citep{Krizhevsky09learningmultiple}, \textsc{TinyImageNet}~\citep{Le2015TinyIV}, \textsc{ImageNet}~\citep{russakovsky2015imagenet}, \textsc{Trec6}~\citep{trec1, trec2}, \textsc{Imdb}~\citep{maas-EtAl:2011:ACL-HLT2011}, and \textsc{Rotten Tomatoes}~\citep{Pang+Lee:05a}. For datasets without pre-specified validation sets, we created a new validation set by splitting the original training set into a 90\% training set and a 10\% validation set. Appendix~\ref{app:datasets} provides detailed information regarding dataset sizes and splits. For the text datasets, we employed the LSTM model sourced from PyTorch, using trainable GloVe embeddings of 300 dimensions as input. We also utilized the BERT+MLP model, which combines a \textsc{Bert-Base} model with a two-layer MLP for classification. Regarding the vision datasets, we used the ResNet18, ResNet50, and ResNet101 models. We trained the ResNet models for 200 epochs with a batch size of 128, except for \textsc{ImageNet}, where we trained for 90 epochs. The LSTM model was trained for 24 epochs with a batch size of 16, while the BERT model was trained for 12 epochs with a batch size of 16. For image datasets, both \model\ and the baselines employed Nesterov's accelerated SGD optimizer with a learning rate of 0.05, weight decay of 5e-4, momentum of 0.9, and a cosine annealing learning rate scheduler, except for \textsc{ImageNet} where a cyclic learning rate scheduler was used. For text datasets with the LSTM model, we utilized the Adam optimizer with a learning rate of 0.001. For fine-tuning the BERT model, we employed the AdamW optimizer with a learning rate of 5e-5. The hyperparameter search spaces used for the hyperparameter tuning experiments are outlined in Appendix~\ref{app:experimentaldetails}. All experiments were conducted on 80GB A100 GPUs.

\noindent \textbf{Pre-trained Transformer Models as Feature Encoders: } 
We utilized the \textsc{Dino-ViTb16} model~\citep{Caron2021EmergingPI} from HuggingFace~\citep{huggingface} as the feature encoder for vision datasets. The final layer CLS token embedding output served as the feature representation. For text datasets, we employed the pre-trained all-distilroberta-v1 model from the sentence transformers package~\citep{reimers-2019-sentence-bert}, computing the average of the final layer embeddings of all words in the sentence for the feature representation. An evaluation of the performance of pre-trained vision transformers and language models as feature encoders for subset selection can be found in Appendix~\ref{app:feature_encoders}. Furthermore, we provide empirical evidence in Appendix~\ref{app:specialized} demonstrating that the pre-trained transformer models used in this study can effectively generalize to specialized or unseen domain datasets.

\noindent \textbf{Efficacy of \model{} for Efficient Model Training: }
Figure~\ref{fig:dss_general_experiments} illustrates the accuracy-efficiency tradeoff comparison for various subset selection methods used in model training. The performance is compared across different subset sizes of the training data, including 1\%, 5\%, 10\%, and 30\%. In our experiments, both the \model{} and \textsc{Adaptive-Random} used a $R$ value of 1, implying subset selection at each epoch. It was empirically observed that an $R$ value of 1 enhances the performance of the \model{}. An ablation study for the $R$ value can be found in Appendix~\ref{app:R_analysis}. To ensure efficiency comparable to other adaptive baselines—namely $\textsc{CraigPB}$, $\textsc{GradMatchPB}$, and $\textsc{Glister}$—we used an $R$ value of 10 for vision experiments and 3 for text experiments. Accuracy degradation versus speedup plots, both in relation to full training, are provided in the sub-figures \ref{fig:CIFAR10_ResNet18}, \ref{fig:CIFAR100_ResNet101}, \ref{fig:tiny_imagenet_resnet101}, \ref{fig:imagenet_resnet50}, \ref{fig:rt_lstm}, and \ref{fig:imdb_bert}. Our results highlight \model{}'s optimal speedup-accuracy tradeoff, making it a greener choice considering CO2 emissions. Using ResNet18 on CIFAR10, \model{} achieves 3.34x and 10.69x speedups with 1.03\% and 4.07\% performance losses, respectively. Similarly, it achieves a 3.34x speedup with a mere 0.9\% performance loss using ResNet50 on ImageNet. When employed with ResNet101 on TinyImageNet and CIFAR100, \model{} yields approximately 3.2x speedup with 1.30\% and 3.18\% performance losses, respectively. \model\ achieves even higher speedup gains of around 10x with performance losses of 2.30\% and 1.23\% on TREC6 and Rotten Tomatoes datasets, respectively. For finetuning the BERT+MLP model on the IMDB dataset, \model{} attains a remarkable 24.94x speedup with just a 1.20\% performance loss. Furthermore, our model notably surpasses the \textsc{Adaptive-Random} baseline, with this advantage being particularly evident on text datasets. On more complex vision datasets, the gap between \model{} and \textsc{Adaptive-Random} widens, indicating the former's superior performance. In sub-figures \ref{fig:cifar100_resnet101_convergence} and \ref{fig:trec6_lstm_convergence}, the \model\ is shown to achieve faster convergence compared to all other methods on CIFAR100 and TREC6 datasets using 30\% subsets. Appendix~\ref{app:experimental_results} provides further efficient training results on additional datasets and model architectures, thereby demonstrating the \model\'s generalizability.

\begin{figure}[t]
\centering
\includegraphics[width = 13cm]{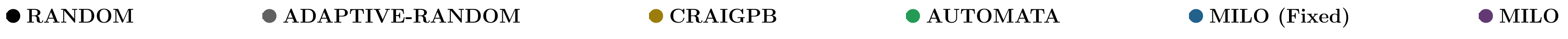}
\centering
\hspace{-0.6cm}
\begin{subfigure}[b]{0.225\textwidth}
\centering
\includegraphics[width=3.1cm, height=2.5cm]{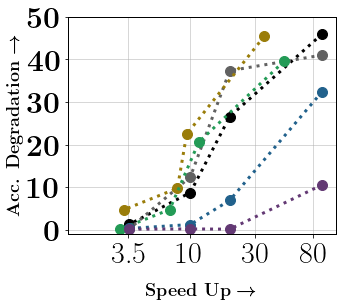}
\caption*{(a) \scriptsize TREC6(Random,HB)}
\phantomcaption
\label{fig:trec6_random_hb}
\end{subfigure}
\begin{subfigure}[b]{0.225\textwidth}
\centering
\includegraphics[width=3.1cm, height=2.5cm]{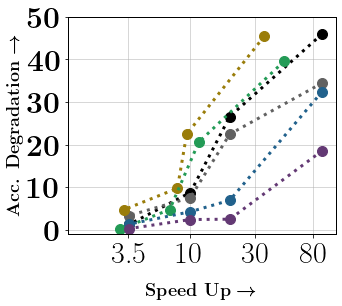}
\caption*{(b) \scriptsize TREC6(TPE, HB)}
\phantomcaption
\label{fig:trec6_tpe_hb}
\end{subfigure}
\begin{subfigure}[b]{0.225\textwidth}
\centering
\includegraphics[width=3.1cm, height=2.5cm]{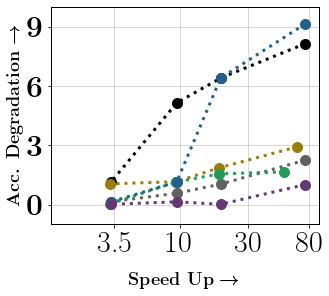}
\caption*{(c) \scriptsize CIFAR10(Random,HB)}
\phantomcaption
\label{fig:cifar10_random_hb}
\end{subfigure}
\begin{subfigure}[b]{0.225\textwidth}
\centering
\includegraphics[width=3.1cm, height=2.5cm]{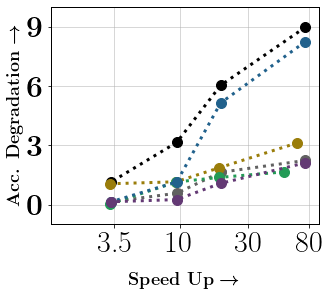}
\caption*{(d) \scriptsize CIFAR10(TPE, HB)}
\phantomcaption
\label{fig:cifar10_tpe_hb}
\end{subfigure}
\caption{\footnotesize{Comparison of \model\ with baselines for hyper-parameter tuning using subset sizes of 1\%, 5\%, 10\%, and 30\% is shown in sub-figures (a-d) for TREC6 and CIFAR10 datasets for combinations of Random Search and Hyperband Scheduler and TPE and Hyperband Scheduler. \emph{The scatter plots show that \model{} achieves the best speedup-accuracy tradeoff in almost every case  \textbf{(bottom-right corner of each plot indicates the best speedup-accuracy tradeoff region})}.}}
\vspace{-6mm}
\label{fig:tuning_experiments}
\end{figure}

\noindent \textbf{Efficacy of \model{} for Hyper-parameter Tuning: }
The effectiveness of \model{} in preserving the original hyper-parameter ordering is evaluated in Appendix~\ref{app:hp_ordering_retention}. We found that \model\ retains the hyper-parameter ordering better than the baseline methods. Figure~\ref{fig:tuning_experiments} illustrates the accuracy-efficiency trade-off among various subset selection methods for hyper-parameter tuning. We evaluated the performance for different subset sizes of the training data: 1\%, 5\%, 10\%, and 30\% using combinations of Random Search~\citep{randomsearch} and Hyperband Scheduler~\citep{hyperband}, as well as TPE~\citep{tpe} and Hyperband Scheduler~\citep{hyperband}. Sub-figures \ref{fig:trec6_random_hb}, \ref{fig:trec6_tpe_hb}, \ref{fig:cifar10_random_hb}, and \ref{fig:cifar10_tpe_hb} display the accuracy degradation vs. speedup plots, both w.r.t full data tuning. \model{} consistently outperforms the baselines, even with smaller subset sizes. Results indicate that \model\ achieves the best tradeoff between speedup and accuracy for hyper-parameter tuning. Specifically, \model{} achieves a 75$\times$ and 20$\times$ speedup on \textsc{Cifar10} and TREC6 datasets, respectively, with a negligible performance loss of about $0.1\%$. 


\section{Conclusion}

We introduce \model{}, a novel, model-agnostic subset selection framework for efficient model training and tuning. \model{} not only rivals the efficiency of random subset selection but also outperforms current state-of-the-art strategies in model convergence. Empirically, \model{} expedites model training and hyper-parameter tuning by 3x to 10x and 20x to 75x, respectively, with minimal performance degradation. Consequently, \model{} offers a significant societal advantage by facilitating faster, energy-efficient model training and tuning, thus reducing CO2 emissions. Although \model{} employs pre-trained transformer models, it does not rely heavily on them. These models can be replaced with smaller proxy models, though the preprocessing times will be slightly longer. However, \model{} does present some challenges, most notably the requirement for a large amount of memory to construct similarity kernels, even with class-wise partitioning strategies. In the future, we will investigate feature-based submodular functions to avoid the need for similarity kernel construction, as well as potential biases introduced when training on data subsets, to further improve the \model{} framework.

\bibliographystyle{plainnat}
\bibliography{refs}

\appendix
\onecolumn
\newpage
\begin{center}
    \Huge{Supplementary Material}
\end{center}
\addcontentsline{toc}{section}{Appendix} 
\part{Appendix} 
\parttoc 
\newpage

\section{Code} \label{app:code}
The code of \model{} is available at the following link: \url{https://anonymous.4open.science/r/MILO-282B/}.

\section{Licenses} \label{app:licenses}
We release the code repository of \model{} with MIT license, and it is available for everybody to use freely. We use the popular deep learning framework~\citep{pytorch} for implementation of \model{} framework, wandb~\citep{wandb} for hyper-parameter search and scheduling algorithms, SUBMODLIB~\citep{SUBMODLIB} for submodular functions and submodular maximization, and CORDS~\citep{Killamsetty_CORDS_COResets_and_2021} for subset selection strategies. We use TREC6~\citep{trec1, trec2}, IMDB~\citep{maas-EtAl:2011:ACL-HLT2011}, Rotten Tomatoes~\citep{Pang+Lee:05a}, \textsc{Cifar10}~\citep{Krizhevsky09learningmultiple},  \textsc{Cifar100}~\citep{Krizhevsky09learningmultiple}, \textsc{TinyImageNet}~\citep{Le2015TinyIV}, \textsc{ImageNet}~\citep{russakovsky2015imagenet}, \textsc{OrganCMNIST}~\citep{medmnistv1, medmnistv2}, \textsc{DermaMNIST}~\citep{medmnistv1, medmnistv2}, and AdeCorpusV2~\citep{GURULINGAPPA2012885} datasets.  The license of the TREC6~\citep{trec1, trec2} dataset is CC0:Public Domain.  IMDB~\citep{maas-EtAl:2011:ACL-HLT2011} dataset is released with a non-commercial license. The licenses of Rotten Tomatoes and AdeCorpusV2 datasets are unknown. But we used the publicly available version from the "HuggingFace Datasets" package. \textsc{Cifar10}, \textsc{Cifar100}, and \textsc{TinyImageNet} are released under MIT license. ImageNet~\citep{russakovsky2015imagenet} is released under a non-commercial license. \textsc{OrganCMNIST}~\citep{medmnistv1, medmnistv2} and \textsc{DermaMNIST}~\citep{medmnistv1, medmnistv2} datasets are released under Creative Commons Attribution 4.0 International license (CC BY 4.0). Furthermore, all the datasets and pre-trained models used in this work are publicly available. In addition, the datasets used do not contain any personally identifiable information.

\section{\model{} Algorithm Pseudocode} \label{app:pseudocode}
We give the pseudo-code of \model{} algorithm for model-training in Algorithm~\ref{alg:mainalg}. 

\begin{algorithm}
 \SetCustomAlgoRuledWidth{0.45\textwidth}
 \LinesNotNumbered
 \DontPrintSemicolon
\footnotesize{
\KwIn{Train set: ${\mathcal D}$; initial params: $\theta_{0}$; learning rate: $\alpha$; total epochs: $T$; selection interval: $R$; SGE Fraction: $\kappa$; Training Loss: $L_T$, Mini-batch size: $B$, Subset size: $k$} \;
\SetKwBlock{Begin}{function}{end function}{
Initialize Graph-Cut function: $f_1(\gS) = \sum_{i \in \gD} \sum_{j \in \gS} s_{ij} - 0.4 \sum_{i \in \gS} \sum_{j \in \gS} s_{ij}$ \;
Initialize Disparity-Min function: $f_2(\gS) = \min_{i, j \in \gS, i \neq j} (1 - s_{ij})$ \;
\textcolor{blue}{*** Check if dataset is pre-processed already ***} \;
\If{$\operatorname{is\_preprocessed}(\gD)$}
    {
    \textcolor{blue}{*** Retrieve pre-selected subsets and pre-constructed probability from metadata ***} \;
    $\gS_0, \gS_R, \gS_{\kappa T - R}, \vp = \operatorname{loadmetadata}(\gD)$ \;}
\Else
    {
     \textcolor{blue}{*** Stochastic Greedy Exploration with Graph-Cut ***} \;
     $\gS_{0}, \gS_{R}, \gS_{\kappa T - R} = \operatorname{SGE}(f_1, \gD, k)$  \;
     \textcolor{blue}{*** Weighted Random Exploration with Disparity-Min ***} \;
     $\vg = [g_0, g_1, \cdots, g_{|\gD|}] = \operatorname{GreedySampleImportance}(f_2, \gD)$ \;
     $\vp = \Big[\frac{1 + g_i + 0.5 g_i^2}{\sum_{j = 1}^{|\gD|}1 + g_j + 0.5 g_j^2}\Big]_{i=1}^{|\gD|} $ \;
     \textcolor{blue}{***Stored selected subsets and constructed probability as metadata***} \;
     $\operatorname{storemetadata}(\gD, \gS_0, \gS_R, \gS_{\kappa T - R}, \vp )$ \;
    }
\textcolor{blue}{*** Model training with subset curriculum ***} \;
Initialize the subset: $\gS = \gS_0$ \;
\textcolor{blue}{*** Model training on representative/easy subsets ***} \;
\For{epochs $t$ in $0, \cdots, \kappa T-1$}
    {
        \If{$(t \mbox{ mod } R == 0)$}
        {
            Update the Subset: $\gS = \gS_t$ \;
        }    
        $\theta_{t+1} = \mbox{mini-BatchSGD}(\gS, \alpha, L_T, B, \mbox{Epochs} = 1)$ \;
    }
\textcolor{blue}{*** Model training on diverse/hard subsets ***}} \;
\For{epochs $t$ in $\kappa T, \cdots, T-1$}
    {
        \If{$(t - \kappa T \mbox{ mod } R == 0)$}
        {
            Sample subset $\gS_t$ from dataset $\gD$ using the probability $\vp$ without replacement \;
            Update the Subset: $\gS = \gS_t$ \;
        }
        $\theta_{t+1} = \mbox{mini-BatchSGD}(\gS, \alpha, L_T, B, \mbox{Epochs} = 1)$ \;
    }
\Return{Final model parameters $\theta_{T}$}}
\caption{\model\ Algorithm}
\label{alg:mainalg}
\end{algorithm}

\begin{algorithm}
 \SetCustomAlgoRuledWidth{0.45\textwidth}
 \LinesNotNumbered
 \DontPrintSemicolon
\footnotesize{
\KwIn{Train set: ${\mathcal D}$; Subset Size: $k$; Set function: $f$; $\epsilon = 1e-2$; Number of Subsets: $n$} \;
\SetKwBlock{Begin}{function}{end function}{
\textcolor{blue}{*** Sampling n subsets using stochastic-greedy ***} \;
\For{$i \in 0, \cdots n-1$}
{
    Initialize empty subset: $\gS_i = \emptyset$ \;
    Set random subset size for the stochastic-greedy algorithm: $s = \frac{|\gD|}{k}\log(\frac{1}{\epsilon})$ \;
    \For{$j \in 0, \cdots, k-1$}
    {
        \textcolor{blue}{*** Sample a random subset by sampling $s$ elements from $\gD \setminus \gS_i$ ***} \;
        $R \leftarrow \operatorname{sample}(\gD \setminus \gS_i, s)$ \;
        $e = \underset{e \in R}{\operatorname{arg\hspace{0.7mm}max\hspace{0.7mm}}} f(A \cup e) - f(A)$ \;
        $\gS_i = \gS_i \cup \{e\}$\;
    }
}}
\Return{$\gS_0, \cdots, \gS_{n-1}$}}
\caption{SGE}
\label{alg:sge}
\end{algorithm}

\begin{algorithm}
\SetCustomAlgoRuledWidth{0.45\textwidth}
\LinesNotNumbered
\DontPrintSemicolon
\footnotesize{
\KwIn{Train set: ${\mathcal D}$; Subset Size: $k$; Set function: $f$}
\SetKwBlock{Begin}{function}{end function}{
Initialize empty subset: $\gS = \emptyset$ \;
Initialize gains vector of dimension $|\gD|$ to zeros $\vg = \overset{\rightarrow}{0}$ \;
\textcolor{blue}{*** Calculate greedy informative gains for each element ***}} \;
\For{$j \in 0, \cdots, |\gD|-1$}
{
    $g = \underset{e \in \gD \setminus \gS}{\max} f(A \cup e) - f(A)$ \;
    $e = \underset{e \in \gD \setminus \gS}{\operatorname{arg\hspace{0.7mm}max\hspace{0.7mm}}} f(A \cup e) - f(A)$ \;
    $\vg[e] = g$ \;
}
\Return{gain vector $\vg$}
}
\caption{GreedySampleImportance Algorithm}
\label{alg:greedyimp}
\end{algorithm}

\section{Instantiations of Different Submodular Functions}
\label{app:submodfuncs}
\subsection{Representation Functions}
\subsubsection{Facility Location}
Given, a dataset $\gD = \{(x_i, y_i)\}_{i=1}^{n}$ with n data samples, the facility location objective can be given as following:

\begin{align}
    f(\gS) = \sum_{i \in \gD} \max_{j \in \gS} s_{ij} 
    \label{eq:fl}
\end{align}

where $s_{ij}$ denotes the similarity of data samples $i$ and $j$. For each data point $i$ in the ground set $\gD$, we compute the most representative samples $j$ from the subset $\gS$ which is closest to it and add these similarities for all data points. Note that due to the sum-max formulation involved in facility location, having one representative data sample from each cluster in your dataset is enough for maximizing the facility location value. This further prevents the selection of more samples in the subset from more dense regions.

\subsubsection{Graph-Cut}
Given, a dataset $\gD = \{(x_i, y_i)\}_{i=1}^{n}$ with n data samples, the graph-cut objective can be given as following:

\begin{align}
    f(\gS) = \sum_{i \in \gD} \sum_{j \in \gS} s_{ij} - \lambda \sum_{i \in \gS} \sum_{j \in \gS} s_{ij}
    \label{eq:gc}
\end{align}

The $\lambda$ parameter in the above equation controls the trade-off between diversity and representation. When $\lambda$ is large, the graph cut function also incorporates diversity into its model. In our experiments, we set $\lambda=0.4$, thereby making the graph-cut function model representation more and making it monotone-submodular. Note that due to the sum-sum formulation involved in graph-cut, selecting more samples from dense regions in the center leads to the maximization of the graph-cut function value. Thus, graph-cut can result in the selection of subsets consisting of easy samples from very dense regions in the dataset.

\subsection{Diversity Functions}
\subsubsection{Disparity-Sum}
Given, a dataset $\gD = \{(x_i, y_i)\}_{i=1}^{n}$ with n data samples, the disparity-sum objective can be given as following:

\begin{align}
    f(\gS) = \sum_{i \in \gS} \sum_{j \in \gS} (1 - s_{ij})
    \label{eq:disp_sum}
\end{align}

Note that in the above equation, disparity-sum purely concentrates on selected samples that are maximally different from each other in the dataset without any care for the representation.

\subsubsection{Disparity-Min}
Given, a dataset $\gD = \{(x_i, y_i)\}_{i=1}^{n}$ with n data samples, the disparity-min objective can be given as following:

\begin{align}
    f(\gS) = \min_{i, j \in \gS, i \neq j} (1 - s_{ij})
    \label{eq:disp_min}
\end{align}

In the above equation, $1 - s_{ij}$ can be considered as a distance measure, and maximization of the disparity-min objective results in the maximization of the minimum distance between the samples in the selected subset. Even though the disparity-min objective is not submodular, it is proven to perform empirically well using conventional greedy approaches~\citep{dasgupta-etal-2013-summarization}.

\section{Assessing EL2N Scores of Subset Selection with Different Set Functions}
\label{app:el2n}
In this section, we assess the difficulty of data samples by measuring the hardness of subsets selected using different set functions. This evaluation is done using the EL2N metric~\citep{EL2N}. Tables~\ref{tab:el2n_cifar10} and \ref{tab:el2n_cifar100} present the Average Sample and Median Sample scores for the selected subsets.

\begin{table}[!ht]
    \centering
    \scalebox{0.9}{
    \begin{tabular}{|l|l|l|l|l|l|}
    \hline
        Dataset & Model & Subset Fraction & Set Function & EL2N (mean) & EL2N (median) \\ \hline \hline
        CIFAR100 & ResNet18 & 0.01 & Graph Cut & 0.31932 & 0.25196815 \\ \cline{4-6}
        ~ & ~ & ~ & Facility Location & 0.5005 & 0.48838884 \\ \cline{4-6}
        ~ & ~ & ~ & Disparity Min & 0.9512 & 0.998251547 \\ \cline{4-6}
        ~ & ~ & ~ & Disparity Sum & 0.9432 & 0.9945051 \\ \cline{3-6}
        ~ & ~ & 0.05 & Graph Cut & 0.39428785 & 0.35422683 \\ \cline{4-6}
        ~ & ~ & ~ & Facility Location & 0.6082416 & 0.6329421 \\ \cline{4-6}
        ~ & ~ & ~ & Disparity Min & 0.8688732 & 0.94560076 \\ \cline{4-6}
        ~ & ~ & ~ & Disparity Sum & 0.8725095 & 0.947182 \\ \cline{3-6}
        ~ & ~ & 0.1 & Graph Cut & 0.44762444 & 0.4188764 \\ \cline{4-6}
        ~ & ~ & ~ & Facility Location & 0.67536277 & 0.74037623 \\ \cline{4-6}
        ~ & ~ & ~ & Disparity Min & 0.8125279 & 0.8964435 \\ \cline{4-6}
        ~ & ~ & ~ & Disparity Sum & 0.8283565 & 0.90549386 \\ \cline{3-6}
        ~ & ~ & 0.3 & Graph Cut & 0.534334 & 0.5372032 \\ \cline{4-6}
        ~ & ~ & ~ & Facility Location & 0.68066484 & 0.73706484 \\ \cline{4-6}
        ~ & ~ & ~ & Disparity Min & 0.79167026 & 0.89193624 \\ \cline{4-6}
        ~ & ~ & ~ & Disparity Sum & 0.7429611 & 0.81957316 \\ 
        \hline \hline
    \end{tabular}}
    \caption{Table presents the Mean and Median EL2N scores~\citep{EL2N} for the CIFAR100 subsets of different sizes selected using various set functions.}
    \label{tab:el2n_cifar100}
\end{table}

\begin{table}[!ht]
    \centering
    \scalebox{0.9}{
    \begin{tabular}{|l|l|l|l|l|l|}
    \hline
        Dataset & Model & Subset Fraction & Set Function & EL2N (mean) & EL2N (median) \\ \hline \hline
        CIFAR10 & ResNet18 & 0.01 & Graph Cut & 0.053498093 & 0.093978405 \\ \cline{4-6}
        ~ & ~ & ~ & Facility Location & 0.24479946 & 0.32504702 \\ \cline{4-6}
        ~ & ~ & ~ & Disparity Min & 0.3455502 & 0.34030083 \\ \cline{4-6}
        ~ & ~ & ~ & Disparity Sum & 0.34296998 & 0.3437602 \\ \cline{3-6}
        ~ & ~ & 0.05 & Graph Cut & 0.093978405 & 0.039304152 \\ \cline{4-6}
        ~ & ~ & ~ & Facility Location & 0.32504702 & 0.21770108 \\ \cline{4-6}
        ~ & ~ & ~ & Disparity Min & 0.34030083 & 0.24009448 \\ \cline{4-6}
        ~ & ~ & ~ & Disparity Sum & 0.3437602 & 0.24489668 \\ \cline{3-6}
        ~ & ~ & 0.1 & Graph Cut & 0.13287988 & 0.060230725 \\ \cline{4-6}
        ~ & ~ & ~ & Facility Location & 0.33706713 & 0.2388236 \\ \cline{4-6}
        ~ & ~ & ~ & Disparity Min & 0.37538758 & 0.24573568 \\ \cline{4-6}
        ~ & ~ & ~ & Disparity Sum & 0.36741473 & 0.24983436 \\ \cline{3-6}
        ~ & ~ & 0.3 & Graph Cut & 0.2021475 & 0.11052186 \\ \cline{4-6}
        ~ & ~ & ~ & Facility Location & 0.33232126 & 0.23153576 \\ \cline{4-6}
        ~ & ~ & ~ & Disparity Min & 0.38120695 & 0.26727863 \\ \cline{4-6}
        ~ & ~ & ~ & Disparity Sum & 0.35034057 & 0.25104147 \\ \hline \hline
    \end{tabular}}
    \caption{Table presents the Mean and Median EL2N scores~\citep{EL2N} for the CIFAR10 subsets of different sizes selected using various set functions.}
    \label{tab:el2n_cifar10}
\end{table}

The results demonstrate that subsets selected using Facility Location and GraphCut exhibit lower hardness scores compared to those chosen by Disparity Min and Disparity Sum. This finding provides strong support for the assertion that Facility Location and GraphCut tend to prefer subsets that are easier, while Disparity Min and Disparity Sum tend to select subsets that are harder. Moreover, the results reveal an expected pattern in the disparity of EL2N scores between diversity-based set functions and representation-based set functions. Specifically, this disparity is more pronounced for smaller subset sizes and gradually diminishes as the subset size increases. 

Another noteworthy finding is that the hardness scores for samples in the CIFAR10 dataset are lower than those in the CIFAR100 dataset. This observation offers a plausible explanation for why a straightforward baseline method like \textsc{Adaptive-Random} performs comparably well to \model{} on the CIFAR10 dataset. This finding further bolsters our claim, as stated in Section~\ref{sec:exp_results}, that \model{} outperforms \textsc{Adaptive-Random} as the dataset complexity increases. These additional insights contribute to the overall strength of our arguments and provide a more comprehensive understanding of the interplay between set functions, hardness scores, and dataset complexity.

\section{Additional Dataset Details} \label{app:datasets}
\begin{table}[!htbp]
    \centering
    \begin{tabular}{|l|l|l|l|l|}
    \hline
        Dataset & \#Classes & \#Train & \#Validation & \#Test \\ \hline \hline
        TREC6 & 6 & 4907 & 545 & 500 \\ \hline
        IMDB & 2 & 22500 & 25000 & 25000\\ \hline
        Rotten Tomatoes & 2 & 8530 & 1066 & 1066\\ \hline
        AdeCorpusV2 & 2 & 23516 & & \\ \hline \hline
        \hline
    \end{tabular}
    \caption{Number of classes, Number of instances in Train, Validation and Test split in Text datasets\label{tab:textdatasplits}}
\end{table}
\subsection{Details of Text Datasets}\label{app:txt_datasets}
We conducted experiments on three text datasets: TREC6 \citep{trec1, trec2}, IMDB \citep{maas-EtAl:2011:ACL-HLT2011}, and Rotten Tomatoes \citep{Pang+Lee:05a}. TREC6 is a question classification dataset comprising open-domain, fact-based questions categorized into broad semantic groups: ABBR (Abbreviation), DESC (Description and Abstract Concepts), ENTY (Entities), HUM (Human Beings), LOC (Locations), and NYM (Numeric Values). Both IMDB and Rotten Tomatoes datasets are sentiment classification datasets, with reviews sourced from their respective platforms. For specialized domain datasets, we utilized the AdeCorpusV2 dataset~\citep{GURULINGAPPA2012885}, which contains data on adverse drug reactions. 

The AdeCorpusV2 dataset was split such that 20\% of the original data was used for testing, and 10\% of the remaining data served as a validation set. The validation data for TREC6 and IMDB datasets were obtained by using 10\% of the training data. A seed value of 42 was used in the generator argument for the random\_split function in PyTorch. The number of classes and instances in each split of the text datasets are summarized in Table~\ref{tab:textdatasplits}.

\subsection{Details of Vision Datasets}\label{app:vision_datasets}
We performed experiments on \textsc{Cifar10}~\citep{Krizhevsky09learningmultiple}, \textsc{Cifar100}~\citep{Krizhevsky09learningmultiple}, and \textsc{TinyImageNet}~\citep{Le2015TinyIV} datasets. The \textsc{Cifar10}~\citep{Krizhevsky09learningmultiple} dataset contains 60,000 colored images of size 32$\times$32 divided into ten classes, each with 6000 images. \textsc{Cifar100}~\citep{Krizhevsky09learningmultiple} is also of size 32$\times$32 but contains 600 images per class and 100 classes. Both \textsc{Cifar10}~\citep{Krizhevsky09learningmultiple} and \textsc{Cifar100}~\citep{Krizhevsky09learningmultiple} have 50,000 training samples and 10,000 test samples distributed equally across all classes. \textsc{TinyImageNet}~\citep{Le2015TinyIV} dataset contains 120,000 colored images of size $64\times64$ from $200$ classes with $600$ images per each class. \textsc{ImageNet}~\citep{russakovsky2015imagenet} dataset contains 1281167 colored images of size $224\times224$ from $1000$ classes. For \textsc{ImageNet} dataset, we report the accuracy on the validation set as test accuracy in this work. For experiments on datasets from specialized domains, we use \textsc{OrganCMNIST}~\citep{medmnistv1, medmnistv2} and \textsc{DermaMNIST}~\citep{medmnistv1, medmnistv2} datasets. \textsc{OrganCMNIST} dataset contains AbdominalCT gray images of size $32\times32$ from $11$ classes. \textsc{DermaMNIST} dataset contains dermatoscope images of size $32\times32$ from $7$ classes.

\begin{table}[!htbp]
    \centering
    \begin{tabular}{|l|l|l|l|l|}
    \hline
        Dataset & \#Classes & \#Train & \#Validation & \#Test \\ \hline \hline
        \textsc{Cifar10} & 10 & 45000 & 5000 & 10000 \\ \hline
        \textsc{Cifar100} & 100 & 45000 & 5000 & 10000 \\ \hline
        \textsc{TinyImageNet} & 200 & 100000 & 10000 & 10000 \\ \hline
        \textsc{ImageNet} & 100 & 1281167   & 50000  &   \\ \hline
        \textsc{OrganCMNIST} & 11 &  13000 & 2392  &  8268 \\ \hline
        \textsc{DermaMNIST} & 7 & 7007 & 1003 & 2005 \\ \hline \hline
    \end{tabular}
    \caption{Number of classes, Number of instances in Train, Validation and Test split in Image datasets\label{tab:imagedatasplits}}
\end{table}

For both \textsc{Cifar10} and \textsc{Cifar100} datasets, $10\%$ of the training data is used for validation (seed value = 42). In Table~\ref{tab:imagedatasplits}, we summarize the number of classes and the number of instances in each split in the image datasets.

\section{Additional Experimental Details} \label{app:experimentaldetails}
\subsection{GPU Resources}\label{app:gpuresources}
We performed experiments on an 80GB A100 GPU cluster. To be fair in timing computation, we ran \model\ and all other baselines for a particular setting on the same GPU server.

\begin{figure}[!ht]
\centering
\includegraphics[width=0.95\textwidth]{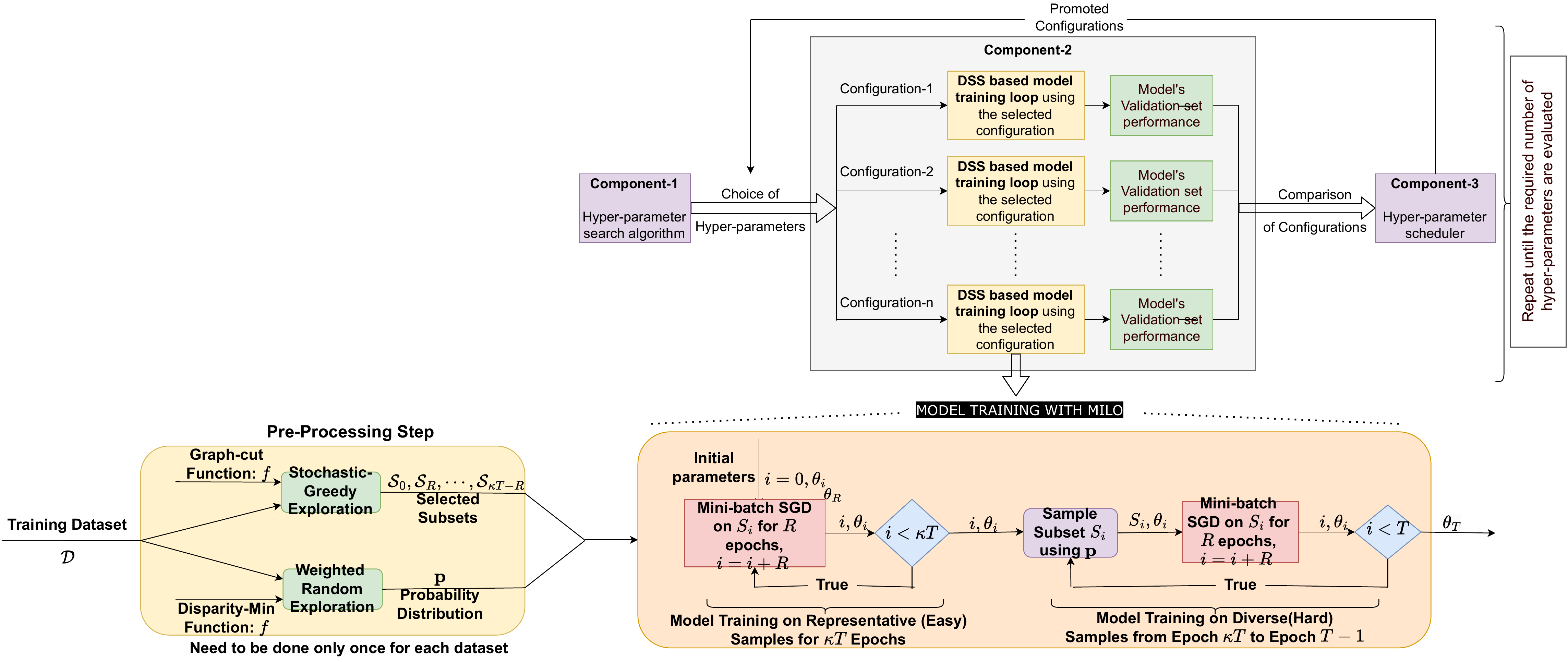}
\caption{Block Diagram of \model{} for hyper-parameter tuning where each individual configuration training is done using \model\  instead of training on the full dataset.}
\label{fig:hp_pipeline}
\end{figure}

\subsection{Experimental Details for Model Training}\label{app:trn_experimentaldetails}
We compare \model\ with \textsc{Random}: randomly sample a fixed subset of the same size subset used by \model\ from the training data, \textsc{Adaptive-Random}: adaptively sample a random subset of the same size subset used by \model\ from the training data every $R$ epochs, \textsc{Full}: using the entire training data for model training and tuning, \textsc{Full-EarlyStop}: where we do an early stop to full training to match the time taken (or energy used) by \model{}, and adaptive gradient-based subset selection strategies for efficient learning where a new subset is selected every $R$ epochs, namely \textsc{CraigPB}: the faster per-batch version of \textsc{Craig}~\citep{mirzasoleiman2020coresets} shown in \citet{killamsetty2021grad}, \textsc{Glister}~\citep{killamsetty2021glister}, \textsc{Grad-MatchPB}: the per-batch version of \textsc{Grad-Match}~\citep{killamsetty2021grad}. Our experiments use a $R$ value of 1 (i.e., subset selection every epoch) for \model{} and \textsc{Adaptive-Random}. We empirically observed that using $R=1$ results in better model performance with \model{}.  In order to achieve comparable efficiency with other adaptive baselines, including $\textsc{CraigPB}$, $\textsc{GradMatchPB}$, and $\textsc{Glister}$, we use an $R$ value of 10 for vision experiments and a $R$ value of 3 for text experiments.

\noindent \textbf{Sepcific Details of Text Experiments}
For text datasets, we use the LSTM model (from PyTorch) with trainable GloVe embeddings~\citep{pennington-etal-2014-glove} of dimension 300 as input and the BERT+MLP model consisting of \textsc{Bert-Base}~\citep{BERT} model with an added two layer MLP for classification. We train the LSTM model for 24 epochs and the BERT model for 12 epochs using a batch size of 16. For text datasets using the LSTM model, we use the Adam optimizer~\citep{ADAM} with a learning rate of 0.001. For BERT model finetuning, we use the AdamW optimizer~\citep{loshchilov2018decoupled} with a learning rate of 5e-5.

\noindent \textbf{Specific Details of Image Experiments}
For vision datasets, we use the ResNet18~\citep{He2016DeepRL} and ResNet101~\citep{He2016DeepRL} models. 
We train the ResNet~\citep{He2016DeepRL} models for 200 epochs using a batch size of 128. For training ResNet models with \model\ and baselines, we use Nesterov's accelerated SGD optimizer with a learning rate of 0.05, weight decay of 5e-4, momentum of 0.9, and a cosine annealing~\citep{loshchilov2017sgdr} learning rate scheduler for all the experiments.


\subsection{Experimental Details for Hyper-Parameter Tuning}\label{app:hp_experimentaldetails}
We follow the experimental setup of \textsc{Automata}~\citep{killamsetty2022automata}, an efficient hyperparameter tuning framework using \textsc{Grad-MatchPB} and replace the subset selection strategy with \model\. Figure~\ref{fig:hp_pipeline} gives a pictorial depiction of the hyper-parameter tuning pipeline with \model\. As discussed in \textsc{Automata}~\citep{killamsetty2022automata}, the hyper-parameter tuning pipeline contains three major components: a) hyper-parameter search algorithms - that returns the list of hyper-parameter configurations that need to be evaluated; b) configuration evaluations involving subset based model training runs - where for each configuration, we train a model using the selected configuration on the subsets selected by \model\ , c) hyper-parameter schedulers - that decides the resources allocated for each configuration evaluation and what configurations to be discarded early. We evaluate the effectiveness of \model\ for hyperparameter tuning, by performing subset-based configuration evaluations using \textsc{Random}, \textsc{Full}, \textsc{Adaptive-Random}, and \textsc{Automata}(\textsc{Grad-MatchPB}) as subset selection baselines. 
 For tuning with \textsc{Full} datasets, the entire dataset is used to train the model during hyperparameter tuning. But for other baselines, we use a subset of the dataset to train various models during tuning. In addition, the subset selection techniques used are adaptive, which means that the model is trained on a different subset every few epochs. We empirically observed that using $R=1$ results in better model performance with \model{}.  In order to achieve comparable efficiency with other adaptive baselines, including $\textsc{CraigPB}$, and $\textsc{GradMatchPB}$, we use an $R$ value of 10 for vision experiments and a $R$ value of 3 for text experiments. We experiment with the combination of a) Random Search~\citep{randomsearch} and Hyperband~\citep{hyperband} scheduler; and b)TPE Search~\citep{tpe} and Hyperband~\citep{hyperband}. We only test with Hyperband~\citep{hyperband} scheduler because we use Wandb~\citep{wandb} for running hyper-parameter optimization algorithms, and wandb provides the support for Hyperband~\citep{hyperband} only. We use Wand for hyper-parameter to accurately compute the tuning times, which we found difficult with Ray~\citep{liaw2018tune} due to its high internal communication times.

\noindent \textbf{Specific Details of Text Experiments}
For text datasets, we use the LSTM model (from PyTorch) with trainable GloVe embeddings~\citep{pennington-etal-2014-glove} of dimension 300 as input. The hyper-parameter space for experiments on text datasets includes learning rate, hidden size \& number of layers of LSTM, and batch size of training. In some experiments (with TPE search algorithm) where the best configuration among 108 configurations are found, the hyper-parameter space is learning rate: [0.001,0.1], LSTM hidden size: \{64,128,256, 512\}, number of layers in LSTM: \{1, 2\}, batch size: \{16,32,64\}.

\noindent \textbf{Specific Details of Image Experiments}
For vision datasets, we use the ResNet18~\citep{He2016DeepRL} model. The hyper-parameter search space for tuning experiments on image datasets includes a choice between the Momentum method and Nesterov Accelerated Gradient method, a choice of learning rate scheduler and their corresponding parameters, and four different group-wise learning rates, $lr_1$ for layers of the first group, $lr_2$ for layers of intermediate groups, $lr_3$ for layers of the last group of ResNet model, and $lr_4$ for the final fully connected layer. For the learning rate scheduler, we change the learning rates during training using either a cosine annealing schedule or decay it linearly by $\gamma$ after every 20 epochs. Best configuration for most experiments is selected from $108$ configurations where the hyper-parameter space is $lr_1$: [0.001, 0.01], $lr_2$: [0.001, 0.01], $lr_3$: [0.001, 0.01], $lr_4$: [0.001, 0.01], Nesterov: \{True, False\}, learning rate scheduler: \{Cosine Annealing, Linear Decay\}, $\gamma$: [0.05, 0.5]. 

\section{Additional Experimental Results} \label{app:experimental_results}
\subsection{Results on Specialized Domain Datasets} \label{app:specialized}
In \model{}, pre-trained models serve a key role in capturing pairwise interactions between data samples in zero-shot setting. The resulting pairwise similarity matrix is then employed for subset selection and probability distribution construction. Notably, these pre-trained models (\textsc{all-distilroberta-v1}, \textsc{DINO (CLS)}), which have been previously trained on natural images or a general text corpus, have demonstrated a certain degree of success in recognizing these interactions, even within specialized domain samples. This can be attributed to the training approach of the pre-trained transformer models utilized in this work. They are trained using either a contrastive or denoising objective, which enables them to identify low-level features in examples from specialized domains, thereby facilitating the capturing of interactions between data samples.

Empirically, we observed that pre-trained transformer models generalize well to datasets from different domains for subset selection. This was demonstrated through a series of experiments on specialized domain datasets. For vision datasets, we conducted experiments on the \textsc{OrganCMNIST} dataset using the EfficientNetB0 model~\citep{tan2020efficientnet}, and the \textsc{DermaMNIST} dataset using the MobileNet model~\citep{howard2017mobilenets}, for subset sizes of 5\% and 10\%. The EfficientNet model was suitably adapted for single-channel images. For text datasets, we carried out experiments on the AdeCorpusV2 dataset using an LSTM model for subset sizes of 5\% and 10\%. We adhered to the training details outlined in Appendix~\ref{app:experimentaldetails}.

\begin{figure}[!ht]
  \centering
  \includegraphics[width = 13cm]{figures/legend_notbold.pdf}
  \centering
  \begin{subfigure}[b]{0.31\columnwidth}
    \includegraphics[width=\linewidth]{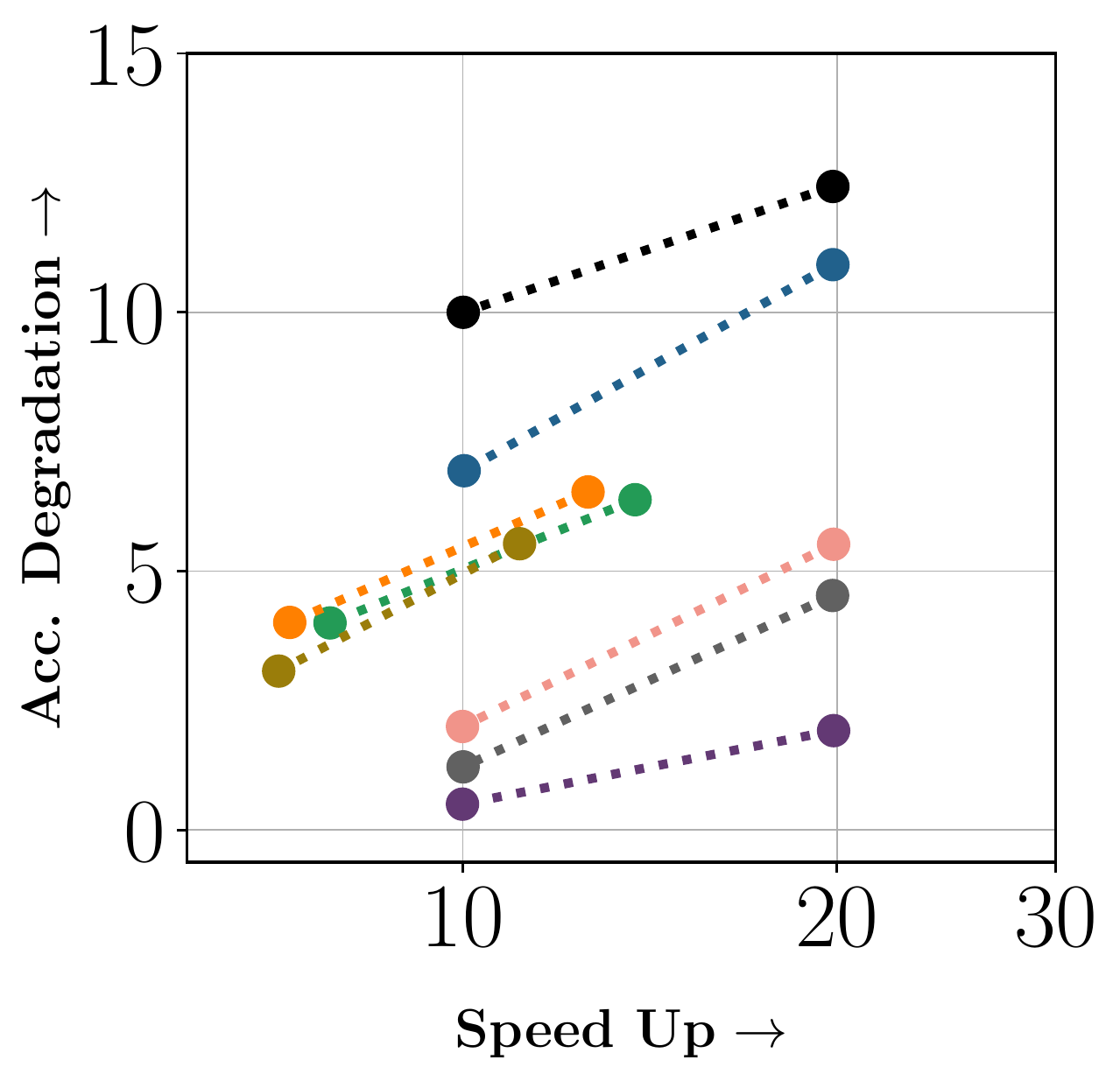}
    \caption{\scriptsize AdeCorpusV2 (LSTM)}
    \label{subfig:adecorpus_lstm}
  \end{subfigure}
  \begin{subfigure}[b]{0.30\columnwidth}
    \includegraphics[width=\linewidth]{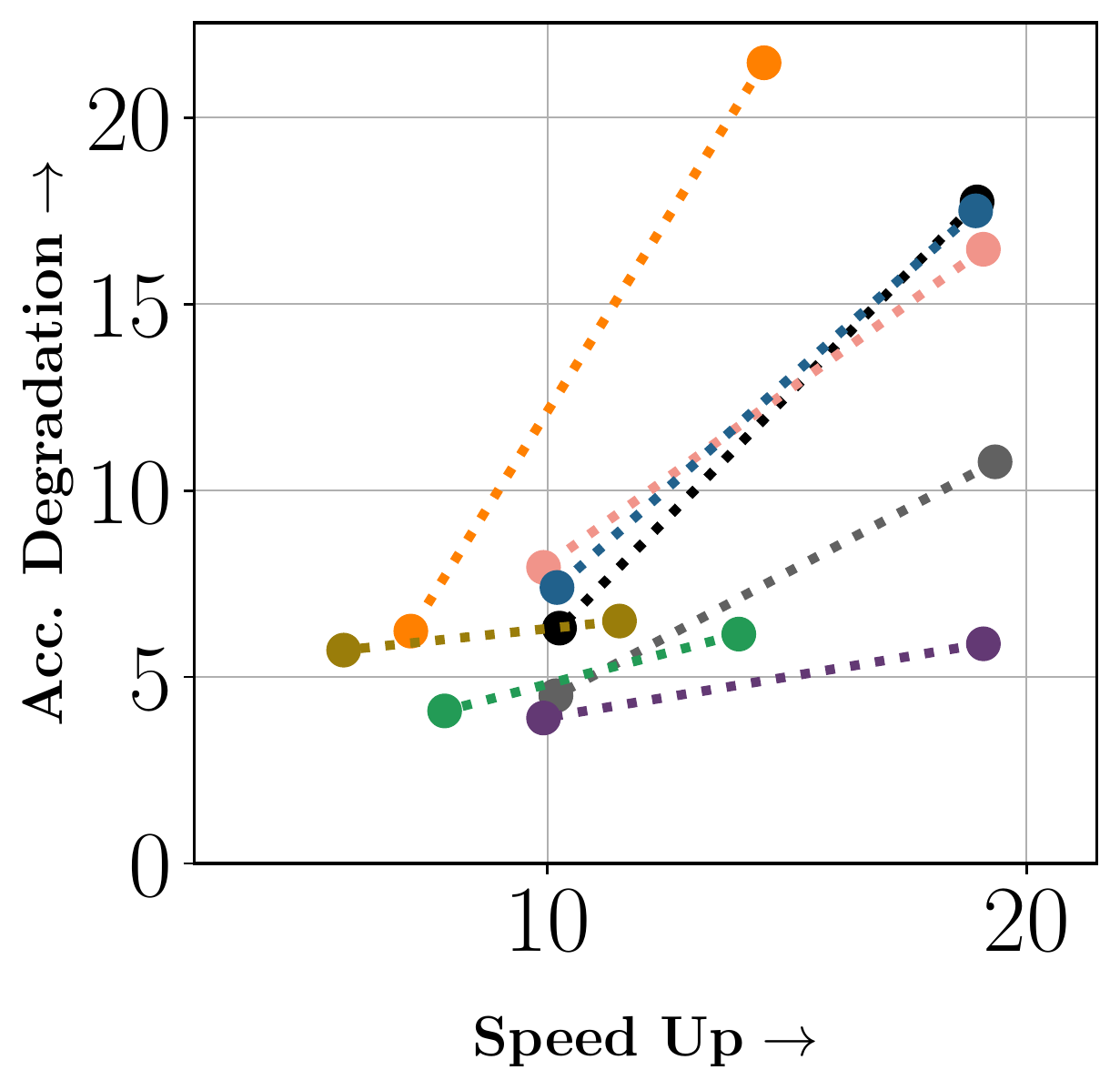}
    \caption{\scriptsize \textsc{OrganCMNIST} (EfficientNetB0)}
    \label{subfig:organcmnist_efficientnet}
  \end{subfigure}
  \begin{subfigure}[b]{0.31\columnwidth}
      \includegraphics[width=\linewidth]{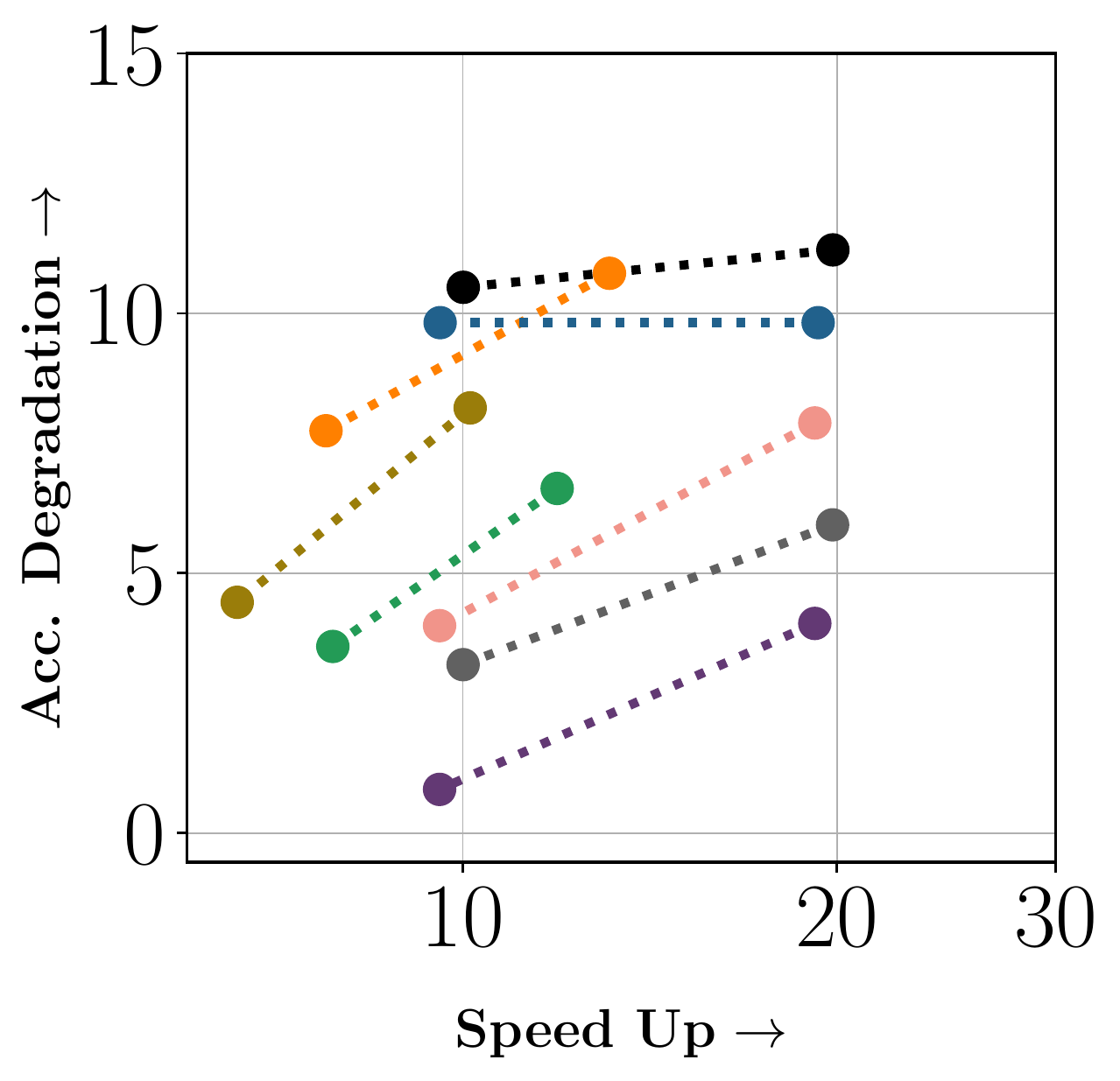}
    \caption{\scriptsize \textsc{DermaMNIST} (MobileNet)}
    \label{subfig:dermamnist_mobilenet}
  \end{subfigure}
  \caption{\footnotesize{Comparison of \model{} to baselines using 5\%, 10\% subsets of specialized domain datasets for model training with general pre-trained transformer models as feature encoders.}}
    \label{fig:specialized_domain}
\end{figure}

Figure~\ref{fig:specialized_domain} presents scatter plots that demonstrate the trade-off between accuracy degradation and computational speed-up for \model{}. These results were obtained using general pre-trained transformer models as feature encoders and compared against baseline models on specialized domain datasets. The subsets utilized for these models constituted 5\% and 10\% of the full dataset, using \textsc{all-distilroberta-v1} as the text feature encoder and \textsc{DINO (CLS)} as the image feature encoder. The results clearly show that \model{} consistently outperforms the baselines in achieving an optimal balance between efficiency and performance, even when general pre-trained models are used.

Moreover, the risk of model overfitting to a small subset and not generalizing well is significantly higher when a fixed data subset is used. However, \model{} allows the downstream model to explore the entire dataset through Stochastic-Greedy Exploration (SGE) and Weighted Random Exploration (WRE). This approach reduces the risk of model overfitting to a small data subset and improves generalization performance. Our empirical findings suggest that \model{} closely approximates optimal full data training performance and experiences minimal performance degradation compared to the baselines. We argue, therefore, that \model{} offers a viable approach for training downstream models that exhibit strong generalization, even when using pre-trained encoders.

\subsection{Results using Proxy Model} \label{app:proxy_model}
As demonstrated in Appendix~\ref{app:specialized}, the pre-trained models (\textsc{all-distilroberta-v1}, \textsc{DINO (CLS)}) utilized in this work have exhibited generalization capabilities to datasets from unseen or specialized domains. However, there might be cases where these models' performance is suboptimal, which can be conveniently assessed through linear probing accuracies. In these instances, a smaller proxy model, such as ResNet18, can be trained to convergence and used as a feature encoder for \model{}, potentially improving performance.

Although training a proxy model increases pre-processing costs, it's worth noting that \model{}'s data pre-processing only needs to be performed once per dataset. Thus, despite an initial increase in pre-processing costs, these can be amortized over time when training multiple downstream models.

To demonstrate \model{}'s effectiveness when using proxy models, we conducted a series of experiments on the \textsc{Cifar100} and \textsc{OrganCMNIST} datasets. Using ResNet18 as a proxy model, and the inputs to the final fully connected layer as sample features, we generated subsets and probability distributions as part of the pre-processing using \model{}'s data exploration strategies. To evaluate the efficiency and generalizability of \model{} in the context of using a proxy model for training, we conducted experiments with ResNet101, EfficientNetB0, and DenseNet121 models. These models were trained using subsets of the \textsc{Cifar100} and \textsc{OrganCMNIST} datasets, selected by \model{} and the considered baselines, with subset sizes of 1\%, 5\%, 10\%, and 30\%.

\begin{figure}[!ht]
  \centering
  \includegraphics[width = 13cm]{figures/legend_notbold.pdf}
  \centering
  \begin{subfigure}[b]{0.31\columnwidth}
      \includegraphics[width=\linewidth]{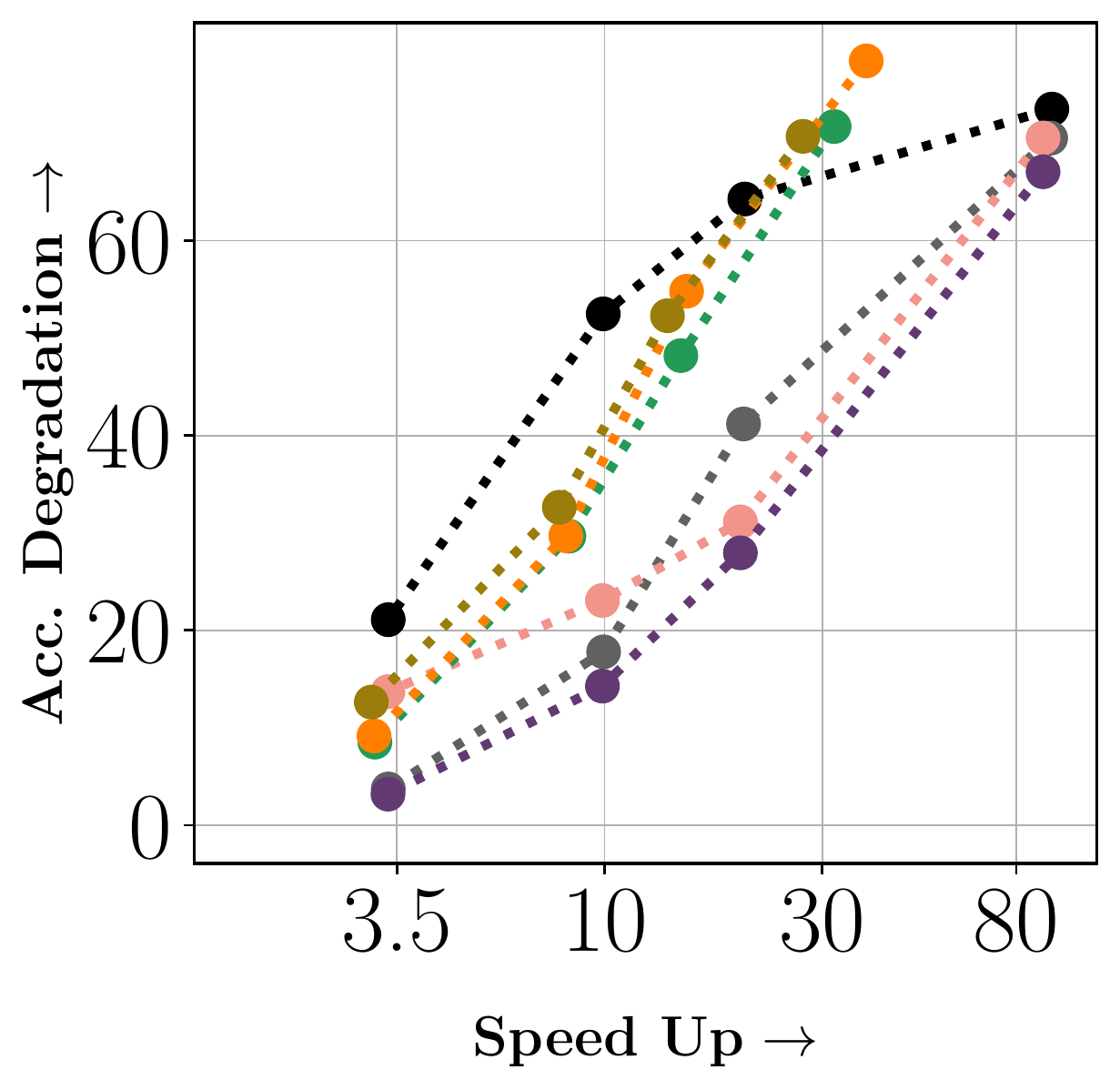}
    \caption{\scriptsize CIFAR100 (ResNet101)}
    \label{subfig:cifar100_proxy_resnet}
  \end{subfigure}
  \begin{subfigure}[b]{0.31\columnwidth}
    \includegraphics[width=\linewidth]{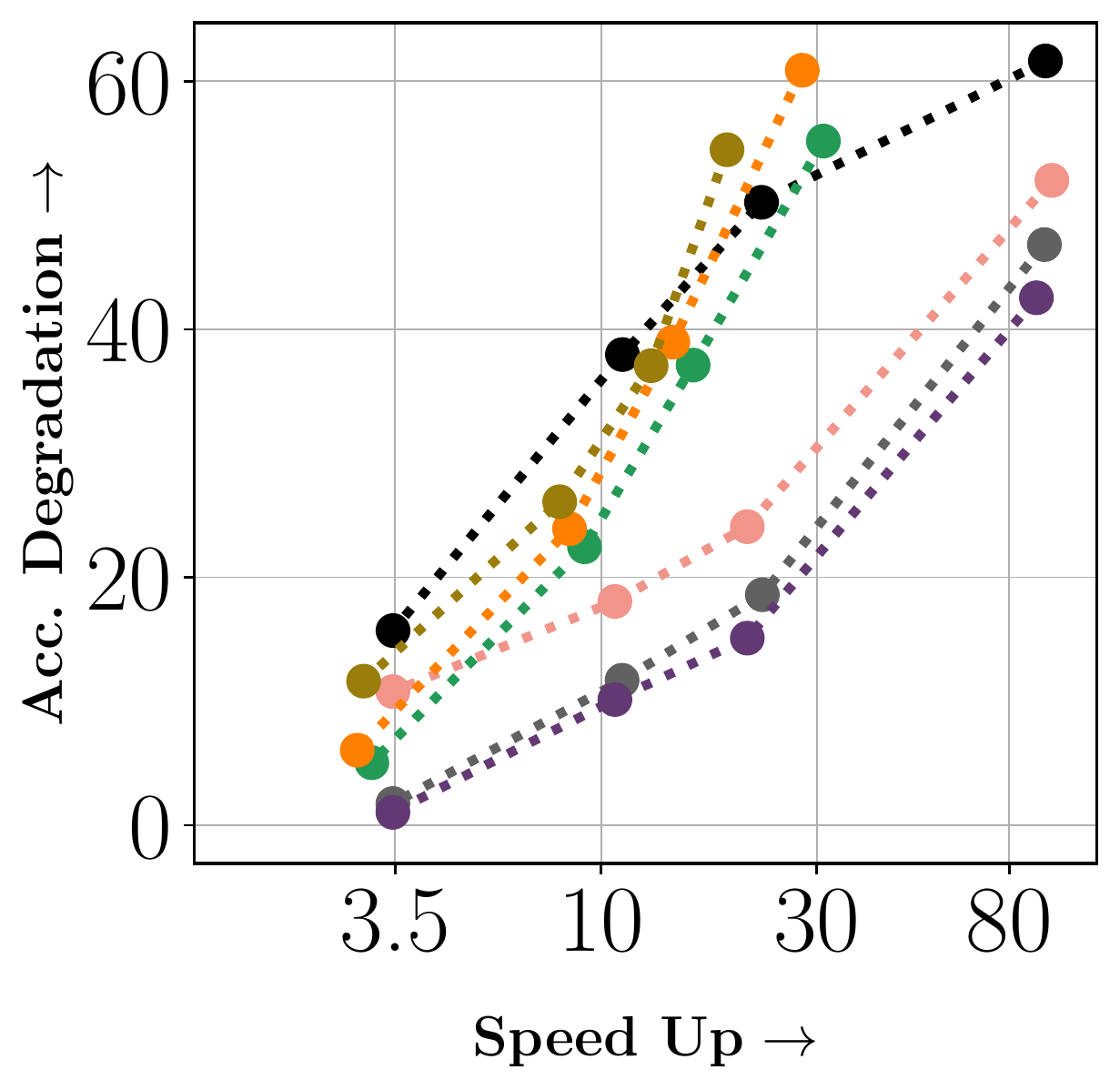}
    \caption{\scriptsize CIFAR100 (EfficientNetB0)}
    \label{subfig:cifar100_proxy_efficientnet}
  \end{subfigure}
  \begin{subfigure}[b]{0.30\columnwidth}
    \includegraphics[width=\linewidth]{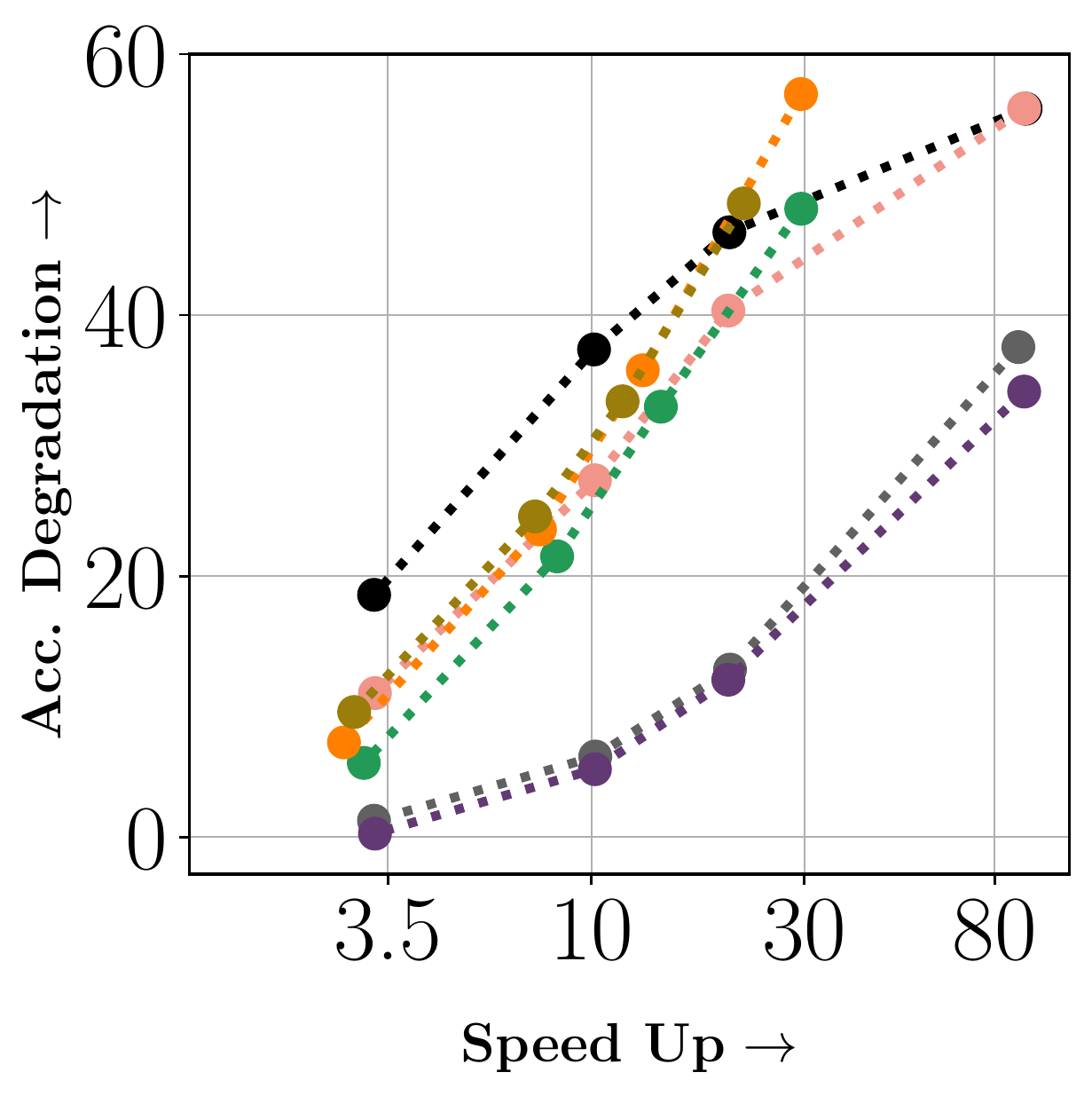}
    \caption{\scriptsize CIFAR100 (DenseNet121)}
    \label{subfig:cifar100_proxy_densenet}
  \end{subfigure}
  \centering
  \begin{subfigure}[b]{0.31\columnwidth}
      \includegraphics[width=\linewidth]{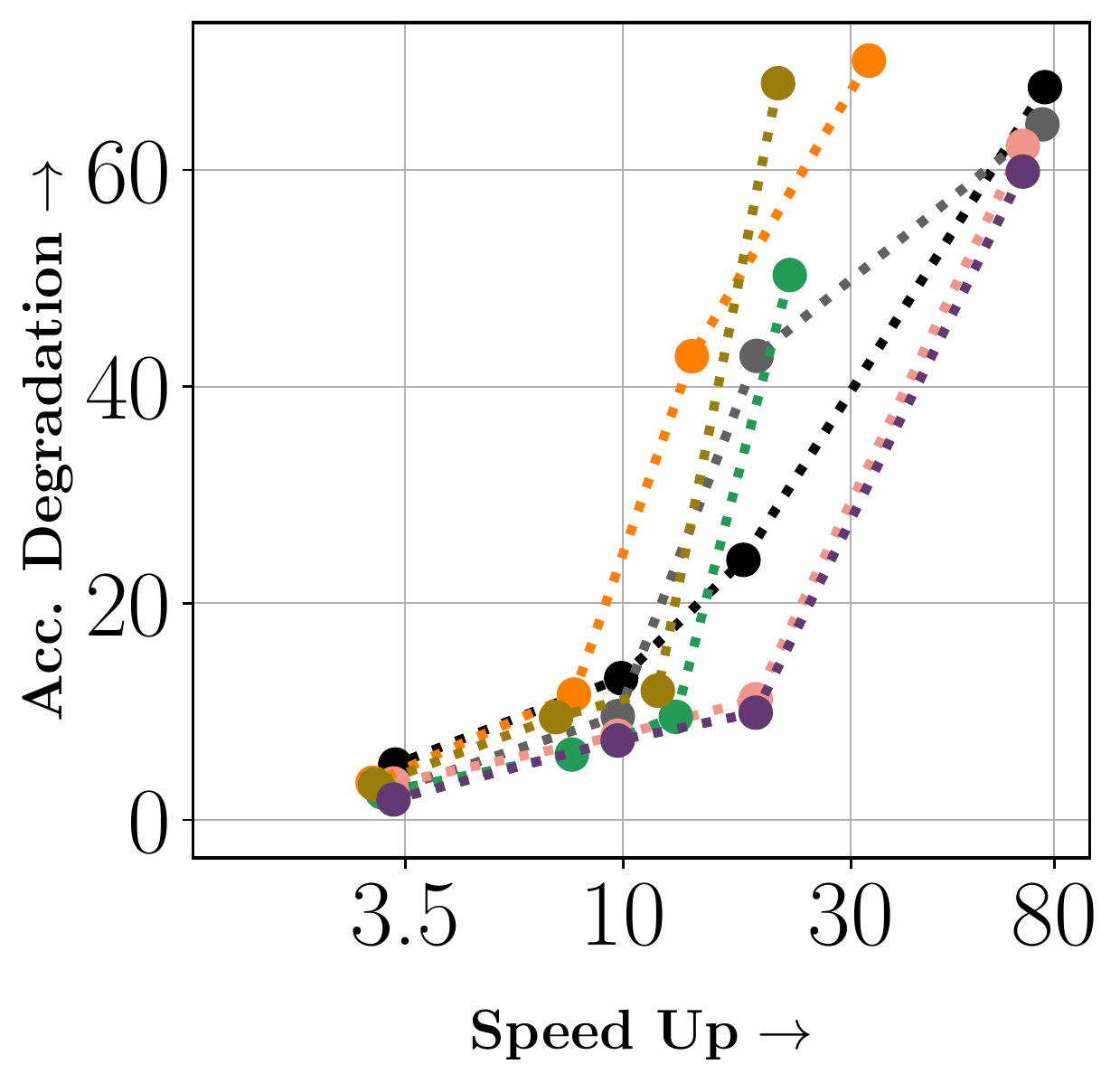}
    \caption{\scriptsize \textsc{OrganCMNIST} (ResNet101)}
    \label{subfig:organcmnist_proxy_resnet}
  \end{subfigure}
  \begin{subfigure}[b]{0.31\columnwidth}
    \includegraphics[width=\linewidth]{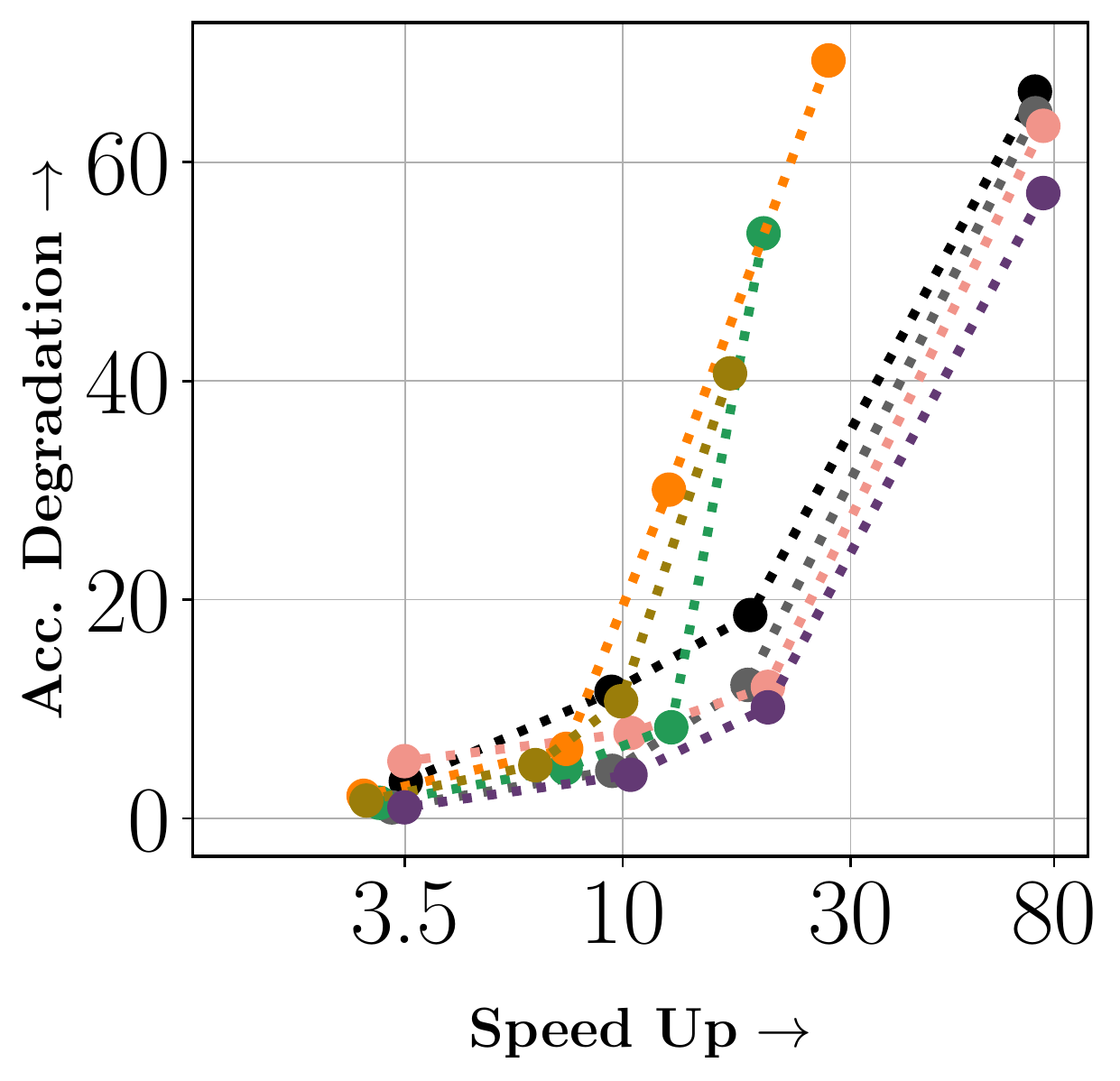}
    \caption{\scriptsize \textsc{OrganCMNIST} (EfficientNetB0)}
    \label{subfig:organcmnist_proxy_efficientnet}
  \end{subfigure}
  \begin{subfigure}[b]{0.30\columnwidth}
    \includegraphics[width=\linewidth]{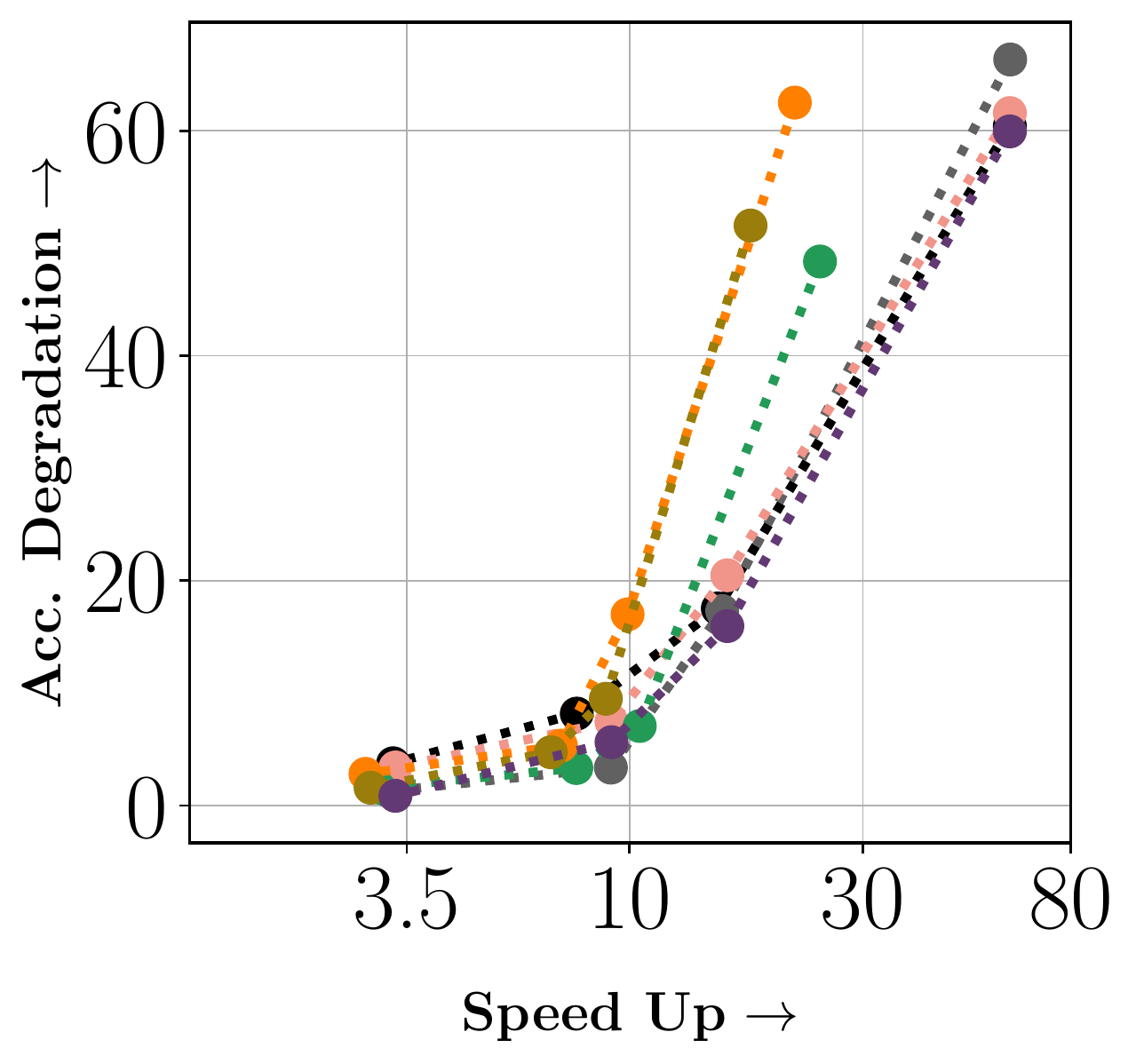}
    \caption{\scriptsize \textsc{OrganCMNIST} (DenseNet121)}
    \label{subfig:organcmnist_proxy_densenet}
  \end{subfigure}
  \caption{\footnotesize{Comparison of \model{} to baselines using 1\%, 5\%, 10\%, and 30\% subsets of \textsc{Cifar100} and \textsc{OrganCMNIST} datasets for model training with the ResNet18 proxy model as the feature encoder.}}
    \label{fig:proxy_models}
\end{figure}

Figure~\ref{fig:proxy_models} presents scatter plots illustrating the trade-off between accuracy degradation and computational speed-up for \model{}, using the ResNet18 proxy model as a feature encoder, and comparing it to baseline models on the \textsc{Cifar100} and \textsc{OrganCMNIST} datasets. The results clearly show that \model{} consistently outperforms the baselines in achieving an optimal balance between efficiency and performance, even when utilizing proxy models. These results indicate that \model{} is not heavily dependent on pre-trained transformer models and can effectively utilize any informative feature encoders. The primary conclusion drawn from these experiments is that the \model{} framework is highly adaptable and can be readily applied across a wide range of scenarios for efficient model training.

\subsection{Data Pre-processing Time} \label{app:preprocessing}
The preprocessing time for \model{}, utilizing pre-trained transformer models as feature encoders, on a machine equipped with 500 GB RAM and an A100 GPU with 80GB memory, for several datasets, is as follows:
\begin{enumerate}
    \item{\textsc{Cifar10 Dataset}: } 5 minutes
    \item{\textsc{Cifar100 Dataset}: } 5 minutes
    \item{\textsc{ImageNet Dataset}: } 20 minutes
\end{enumerate}

The preprocessing times for \model{}, using pre-trained transformer models as feature encoders, are relatively minimal compared to the full training times for models of varying sizes. Specifically, the preprocessing times amount to between 3\% and 5\% of the full training times for a medium-sized model like ResNet18, and less than 3\% of the full training time for a larger model like ResNet101 on the datasets considered. Importantly, this preprocessing step is independent of the model used for training and can be shared across any model being trained on these datasets. Therefore, it only needs to be done once per dataset and subset size.

However, when using proxy models, the preprocessing costs do increase because the training time of the proxy model also needs to be taken into account. For instance, when using a ResNet18 as a proxy model, the preprocessing times can equate to between 103\% and 105\% of the full training times. Despite this increase, these preprocessing costs can be amortized over multiple model trainings, maintaining the time-efficiency of \model{} for training models on large datasets.

\subsection{Test-Accuracies, Training times and Standard deviations for Model Training Experiments} \label{app:std_results}
Table~\ref{tab:vision_sel_results}, Table~\ref{tab:text_sel_results} shows the top-1 test accuracies and training times taken by \model{} and the other baselines considered for single model training on \textsc{Cifar10}, \textsc{Cifar100}, \textsc{TinyImageNet}, TREC6, IMDB, Rotten Tomaotes datasets for different subset sizes of 1\%, 5\%, 10\%, and 30\% respectively. Furthermore, Table~\ref{tab:vision_std_results}, Table~\ref{tab:text_std_results} gives the standard deviation numbers of \model{} and other baselines for single model training on \textsc{Cifar10}, \textsc{Cifar100}, \textsc{TinyImageNet}, TREC6, IMDB, Rotten Tomaotes datasets for different subset sizes of 1\%, 5\%, 10\%, and 30\% respectively. 

\begin{sidewaystable}[!htbp]
    \centering
    \scalebox{0.7}{
    \begin{tabular}{c c c|c c c c|c c c c} \hline \hline
\multicolumn{11}{c}{Model Training Results on Vision Datasets}\\ \hline
\multicolumn{3}{c|}{} & \multicolumn{4}{c|}{Top-1 Test accuracy of the Model(\%)} & \multicolumn{4}{c}{Model Training time(in hrs)} \\ 
\multicolumn{1}{c}{} & \multicolumn{1}{c}{} & \multicolumn{1}{c|}{Budget(\%)} & \multicolumn{1}{c}{1\%} & \multicolumn{1}{c}{5\%} & \multicolumn{1}{c}{10\%} & \multicolumn{1}{c|}{30\%} & \multicolumn{1}{c}{1\%} & \multicolumn{1}{c}{5\%} & \multicolumn{1}{c}{10\%} & \multicolumn{1}{c}{30\%} \\ \hline
\multicolumn{1}{c}{Dataset} & \multicolumn{1}{c}{Model}  &\multicolumn{1}{c|}{Selection Strategy} & \multicolumn{4}{c|}{} & \multicolumn{4}{c}{} \\ \hline
CIFAR10 &ResNet18 &\textsc{Full} (skyline for test accuracy)  &95.19 &95.19  &95.19  &95.19  &1.72736889& 1.72736889& 1.72736889& 1.72736889 \\ 
 & &\textsc{Random} (skyline for training time) &39.91& 63.52& 77.47& 89.62  &0.01646028& 0.08030361& 0.1722375& 0.50628667\\
 & &\textsc{Adaptive-Random} (skyline for training time) &63.71& 88.2& 91.09& 94.05&0.01632191& 0.08136123& 0.1776311& 0.5068371\\\cline{3-11}
 & &\textsc{Glister} &38.2& 79.02& 90.67& 93.04  &0.08538972& 0.16018167& 0.23015611& 0.57614889\\
 & &\textsc{CraigPB}  &64.11& 84.35& 88.97& 92.99 &0.13940694& 0.19342472& 0.26544833& 0.60756111\\
 & &\textsc{GradMatchPB}  &61.59& 85.34& 89.67& 93.71 &0.08093056& 0.13492361& 0.21210222& 0.52081639\\
 & &\textsc{Milo} (Fixed)  &38.17& 64.17& 78.06& 89.21 &0.01667131& 0.0821212& 0.172141& 0.50281731\\ 
 & &\textsc{Milo}  &68.56& 88.37& 91.11& 94.12&0.0163121& 0.0808317& 0.1702131& 0.50645219\\ \hline \hline
 CIFAR10 &ResNet101 &\textsc{Full} (skyline for test accuracy) & 95.01&  95.01&  95.01& 95.01 &2.607 &2.607 &2.607 &2.607\\ 
 & &\textsc{Random} (skyline for training time) &31.49 &54.35  &69.99& 88.51  &0.02820806& 0.14013056& 0.27079694& 0.85268889\\
 & &\textsc{Adaptive-Random} (skyline for training time) &21.72 &80.25  &88.13  &94.2  &0.0270875& 0.13807778& 0.2795975& 0.82271222\\\cline{3-11}
 & &\textsc{Glister} &10.46& 62.41& 83.79& 92.47& 0.12346444& 0.24328889& 0.38844556& 1.05459139\\
 & &\textsc{CraigPB} &33.74& 74.07& 85.4& 92.44 &0.16501306& 0.27757139& 0.41062028& 0.9307625\\
 & &\textsc{GradMatchPB}  &27.23& 71.71& 86.35& 93.39& 0.13171083& 0.23819667& 0.36952333& 0.89125889\\
 & &\textsc{Milo} (Fixed)  &31.14& 52.73& 66.53& 87.87 &0.0269& 0.1321025& 0.28677639& 0.850435\\ 
 & &\textsc{Milo}  &43.37 &81.63  &88.72  &94.24  &0.026376&  0.1321091& 0.28635171& 0.8503878\\ \hline \hline
 CIFAR100 &ResNet18 &\textsc{Full} (skyline for test accuracy)  &77.03 &77.03  &77.03  &77.03  &1.52 &1.52 &1.52 &1.52\\ 
 & &\textsc{Random} (skyline for training time) &8.937& 21.74& 35.03& 61.93& 0.0152& 0.0836& 0.1554& 0.449\\
 & &\textsc{Adaptive-Random} (skyline for training time) &29.99& 61.86& 69.14& 74.82& 0.016& 0.0757& 0.1532& 0.4491\\\cline{3-11}
 & &\textsc{Glister} &2.72& 52.17& 65.72& 72.59& 0.0872& 0.1553& 0.2362& 0.5717\\
 & &\textsc{CraigPB}  &23.38& 51.56& 59.6& 70.25& 0.1344& 0.193& 0.2679& 0.5568\\
 & &\textsc{GradMatchPB}  &24.4& 54.42& 65.93& 73.57& 0.07873& 0.1384& 0.2125& 0.4876\\
  & &\textsc{Milo} (Fixed)  &9.133& 22.57& 36.03& 61.97& 0.0183& 0.0797& 0.1576& 0.449\\
 & &\textsc{Milo}  &35.34 &64.4  &69.28  &74.95  &0.0173& 0.082& 0.1506& 0.4442\\ \hline \hline
 CIFAR100 &ResNet101 &\textsc{Full} (skyline for test accuracy)  &78.46& 78.46& 78.46& 78.46& 2.6034& 2.6034& 2.6034& 2.6034\\ 
 & &\textsc{Random} (skyline for training time) &5.18& 17.81& 30.17& 57.93& 0.02952& 0.14524& 0.2879& 0.825375\\
 & &\textsc{Adaptive-Random} (skyline for training time) &6.199& 39.11& 60.02& 73.21& 0.03112& 0.14523& 0.2879 &0.825175
\\\cline{3-11}
 & &\textsc{Glister} &1.634& 36.76& 56.37& 73.7& 0.12156& 0.24377& 0.39149& 1.17212\\
 & &\textsc{CraigPB}  &6.8& 36.68& 50.92& 69.02& 0.1697& 0.2727& 0.40612& 0.912357\\
 & &\textsc{GradMatchPB}  &6.653& 38.23& 55.82& 74.35& 0.136& 0.246& 0.3665& 0.91235\\
 & &\textsc{Milo} (Fixed)  &6.064&17.17&31.73&59.05&0.02952&0.14523&0.28797&0.82537\\
 & &\textsc{Milo}  &12.99 &49.89  &66.11  &75.28  &0.030778& 0.145234& 0.28797& 0.8253\\ \hline \hline
 TinyImageNet &ResNet18 &\textsc{Full} (skyline for test accuracy)  &52.44 &52.44  &52.44  &52.44  &15.411 &15.411 &15.411 &15.411\\ 
 & &\textsc{Random} (skyline for training time) &3.21 &13  &19.61  &35.68  &0.171 &0.874 &1.82 &4.99\\
 & &\textsc{Adaptive-Random} (skyline for training time) &0.62 &27.34  &38.73  &50.3  &0.168 &0.832 &1.82 &5.12\\\cline{3-11}
 & &\textsc{Glister} &0.9667& 25.53& 36.41& 46.86& 1.375& 2.061& 2.823& 5.968\\
 & &\textsc{CraigPB}  &1.31 &22.31  &34.21  &4.31  &1.98 &2.56 &3.241 &6.42\\
 & &\textsc{GradMatchPB}  &3.21 &28.41  &35.64  &50.34  &1.68 &2.02 &2.674 &5.954\\
 & &\textsc{Milo} (Fixed)  &4.517 &13.83  &20.03  &35.79  &0.1795 &0.79 &1.82 &4.89\\
 & &\textsc{Milo}  &16.33 &31.7  &39.99  &50.56  &0.1701 &0.81 &1.81 &4.97\\ \hline \hline
 TinyImageNet &ResNet101 &\textsc{Full} (skyline for test accuracy)  &56.32&56.32&56.32&56.32& 17.2387& 17.2387& 17.2387 &17.2387\\ 
 & &\textsc{Random} (skyline for training time) &2.84&10.52&17.57&35.68& 0.150245& 0.91737& 1.826483&5.437941\\
 & &\textsc{Adaptive-Random} (skyline for training time) &0.6067&15.99&33.19&53.3& 0.1710 &0.91737& 1.82648& 5.43794\\\cline{3-11}
 & &\textsc{Glister} &0.6255&16.3&30.47&50.11& 1.47556&2.18824&3.17468&7.099\\
 & &\textsc{CraigPB}  &2.54&16.23&28.43&46.42&1.7016&2.348&3.32867&7.42342\\
 & &\textsc{GradMatchPB}  &2.15&16.84&33.25&52.31&1.53448&2.1209&3.0624&7.03117\\
 & &\textsc{Milo} (Fixed)  &2.76&11.25&18.97&35.79&0.150245&0.91737&1.82648&5.4379\\
 & &\textsc{Milo}  &4.64&24.39&35.16&55.02& 0.19173&0.91737&1.82648&5.4379\\ \hline \hline
\end{tabular}}
    \caption{Data Selection Results for \textsc{Cifar10}, \textsc{Cifar100} and \textsc{TinyImageNet} datasets}
    \label{tab:vision_sel_results}
\end{sidewaystable}

\begin{table}[!htbp]
    \centering
    \scalebox{0.7}{
    \begin{tabular}{c c c|c c c c} \hline \hline
\multicolumn{7}{c}{Test Accuracy Standard Deviation Results on Vision Datasets}\\ \hline
\multicolumn{3}{c|}{} & \multicolumn{4}{c|}{Test Accuracy Standard Deviation(for 5 runs)}\\ 
\multicolumn{1}{c}{} & \multicolumn{1}{c}{} & \multicolumn{1}{c|}{Budget(\%)} & \multicolumn{1}{c}{1\%} & \multicolumn{1}{c}{5\%} & \multicolumn{1}{c}{10\%} & \multicolumn{1}{c|}{30\%}\\ \hline
\multicolumn{1}{c}{Dataset} & \multicolumn{1}{c}{Model}  &\multicolumn{1}{c|}{Selection Strategy} & \multicolumn{4}{c|}{}  \\ \hline
CIFAR10 &ResNet18 &\textsc{Full} (skyline for test accuracy)  &0.43 &0.43 &0.43 &0.43 \\ 
 & &\textsc{Random} (skyline for training time) &0.187 &0.127 &0.173 &0.154 \\
 & &\textsc{Adaptive-Random} (skyline for training time) &1.981 &0.218  &1.87  &0.0837 \\\cline{3-7}
 & &\textsc{Glister} &0.76 &1.872  &5.8721  &0.863  \\
 & &\textsc{CraigPB}  &9.31 &4.831  &4.313  &0.2841  \\
 & &\textsc{GradMatchPB}  &2.041 &1.0841  &0.7631  &0.0736 \\
 & &\textsc{Milo} (Fixed)  &1.471 &4.31 &4.12 &2.191 \\
 & &\textsc{Milo}  &0.762 &1.712  &0.0821  &0.037  \\ \hline \hline
CIFAR10 &ResNet101 &\textsc{Full} (skyline for test accuracy)  &0.3606 &0.3606 &0.3606 &0.3606 \\ 
 & &\textsc{Random} (skyline for training time) &0.141 &0.1527  &0.1424  &0.077 \\
 & &\textsc{Adaptive-Random} (skyline for training time) &1.295 &0.198  &1.28  &0.063  \\\cline{3-7}
 & &\textsc{Glister} &0.51& 10.27& 5.56& 0.37  \\
 & &\textsc{CraigPB}  &3.804& 5.996& 2.729& 0.2051  \\
 & &\textsc{GradMatchPB}  &2.086& 0.5374& 0.02& 0.16\\
 & &\textsc{Milo} (Fixed)  &3.84& 5.64& 6.13& 1.01\\
 & &\textsc{Milo}  &2.093& 3.691& 0.03& 0.02  \\ \hline \hline
CIFAR100 &ResNet18 &\textsc{Full} (skyline for test accuracy)  &0.3986 &0.3986  &0.3986  &0.3986 \\ 
 & &\textsc{Random} (skyline for training time) &0.2848& 0.9871& 0.637& 0.7168 \\
 & &\textsc{Adaptive-Random} (skyline for training time) &0.539& 0.2546& 0.176& 0.3915\\\cline{3-7}
 & &\textsc{Glister} &0.516& 1.32& 0.1473& 0.08  \\
 & &\textsc{CraigPB}  &0.5798& 0.8132& 0.3323& 0.1626 \\
 & &\textsc{GradMatchPB}  &0.7637& 0.3536& 1.039& 0.6293 \\
  & &\textsc{Milo} (Fixed)  &0.049& 0.56& 0.69& 0.84\\
 & &\textsc{Milo} &0.8211 &0.3353  &0.1041  &0.39 \\ \hline \hline
 CIFAR100 &ResNet101 &\textsc{Full} (skyline for test accuracy)  &0.31 &0.31  &0.31  &0.31 \\ 
 & &\textsc{Random} (skyline for training time) &0.2716 &1.213  &1.273  &0.891 \\
 & &\textsc{Adaptive-Random} (skyline for training time) &0.831 &0.628  &0.583  &0.846  \\\cline{3-7}
 & &\textsc{Glister} &0.631& 1.172&  0.31&  1.23  \\
 & &\textsc{CraigPB}  &0.831 &1.731  &0.347  &0.541  \\
 & &\textsc{GradMatchPB}  &0.642 & 0.24 &0.53  &0.62 \\
 & &\textsc{Milo} (Fixed)  &0.192 &0.0931 &0.041 &0.47 \\
 & &\textsc{Milo}  &1.013 &0.764  &0.453  &0.036  \\ \hline \hline
 TinyImageNet &ResNet18 &\textsc{Full} (skyline for test accuracy)  &0.51 &0.51  &0.51  &0.51 \\ 
 & &\textsc{Random} (skyline for training time) &0.53 &0.872  &0.41  &0.73 \\
 & &\textsc{Adaptive-Random} (skyline for training time) &0.41 &0.763  &0.31  &0.63  \\\cline{3-7}
 & &\textsc{Glister} &0.125& 0.45& 0.45& 0.17\\
 & &\textsc{CraigPB}  &2.31 &0.47  &0.51  &0.21  \\
 & &\textsc{GradMatchPB}  &0.73 &0.23  &0.46 &0.53 \\
 & &\textsc{Milo} (Fixed)  &0.32 &0.43 &0.24 &0.53 \\
 & &\textsc{Milo}  &0.53 &0.41  &0.32  &0.41  \\ \hline \hline
 TinyImageNet &ResNet101 &\textsc{Full} (skyline for test accuracy)  &0.34 &0.34  &0.34  &0.34 \\ 
 & &\textsc{Random} (skyline for training time) &0.65 &0.43  &0.41  &0.53 \\
 & &\textsc{Adaptive-Random} (skyline for training time) &0.64 &0.31  &0.24  &0.41  \\\cline{3-7}
 & &\textsc{Glister} &0.53 &0.741  &0.452  &0.371  \\
 & &\textsc{CraigPB}  &0.872 &0.735  &0.472  &0.531  \\
 & &\textsc{GradMatchPB}  &0.764 &0.31  &0.42  &0.53 \\
 & &\textsc{Milo} (Fixed)  &0.736 &0.24 &0.43 &0.45\\
 & &\textsc{Milo}  &0.41 &0.24  &0.41  &0.431  \\ \hline \hline
\end{tabular}}
    \caption{Standard Results for \textsc{Cifar10}, \textsc{Cifar100} and \textsc{TinyImageNet} datasets}
    \label{tab:vision_std_results}
\end{table}

\begin{sidewaystable}[!htbp]
    \centering
    \scalebox{0.7}{
    \begin{tabular}{c c c|c c c c|c c c c} \hline \hline
\multicolumn{11}{c}{Model Training Results on Text Datasets}\\ \hline
\multicolumn{3}{c|}{} & \multicolumn{4}{c|}{Top-1 Test accuracy of the Model(\%)} & \multicolumn{4}{c}{Model Training time(in secs)} \\ 
\multicolumn{1}{c}{} & \multicolumn{1}{c}{} & \multicolumn{1}{c|}{Budget(\%)} & \multicolumn{1}{c}{1\%} & \multicolumn{1}{c}{5\%} & \multicolumn{1}{c}{10\%} & \multicolumn{1}{c|}{30\%} & \multicolumn{1}{c}{1\%} & \multicolumn{1}{c}{5\%} & \multicolumn{1}{c}{10\%} & \multicolumn{1}{c}{30\%} \\ \hline
\multicolumn{1}{c}{Dataset} & \multicolumn{1}{c}{Model}  &\multicolumn{1}{c|}{Selection Strategy} & \multicolumn{4}{c|}{} & \multicolumn{4}{c}{} \\ \hline
TREC6 & LSTM &\textsc{Full} (skyline for test accuracy)  &86.6&86.6&86.6&86.6&101.835&101.835&101.835&101.835\\ 
 & &\textsc{Random} (skyline for training time) &40.2&62.93&72.27&80.93& 1.61&5.83&10.185&31.23\\
 & &\textsc{Adaptive-Random} (skyline for training time) &28.13&61.0&81.53&85.03& 1.623&5.82&10.171&31.46\\\cline{3-11}
 & &\textsc{Glister} &24.52&65.45&84.21&86.21&12.621&17.825&19.452&46.742\\
 & &\textsc{CraigPB}  &25.53&52.87&83.47&85.43&33.83&39.841&43.957&60.742\\
 & &\textsc{GradMatchPB}  &23.04&69.93&85.0&87.0&8.621&14.363&18.553&43.848\\
 & &\textsc{Milo} (Fixed)  &34.2&66.07&69.26&81.05& 1.614&5.72&10.174&31.22\\
 & &\textsc{Milo}  &46.87&76.13&84.3&87.97& 1.613&5.81&10.182&31.21\\ \hline \hline
IMDB &LSTM &\textsc{Full} (skyline for test accuracy)  & 89.03&89.03&89.03&89.03&613.553&613.553&613.553&613.553\\ 
 & &\textsc{Random} (skyline for training time) &50.01&50.01&79.14&84.42&6.789&37.07&63.86&192.587\\
 & &\textsc{Adaptive-Random} (skyline for training time) &49.99&60.04&88.13&88.29&6.673&37.14&63.88&193.632\\\cline{3-11}
 & &\textsc{Glister} &50.02&61.25&85.21&86.81&110.242&137.256&166.839&300.512\\
 & &\textsc{CraigPB}  &50.02&66.73&87.54&87.7&118.553&137.256&166.839&295.567\\
 & &\textsc{GradMatchPB}  &50.02&58.99&84.66&85.75&99.116&137.256&166.839&347.057\\
 & &\textsc{Milo} (Fixed)  &50.03&66.54&81.3&85.06&6.791&37.06&63.86&192.59\\
 & &\textsc{Milo}  &50.02&69.15&88.45&88.56&6.81&37.1&63.89&192.57\\ \hline \hline
 Rotten Tomatoes &LSTM &\textsc{Full} (skyline for test accuracy)  &79.54&79.54&79.54&79.54&177.656&177.656&177.656&177.656\\ 
 & &\textsc{Random} (skyline for training time) &52.43&67.39&69.08&73.82&2.305&9.274&18.159&56.883\\
 & &\textsc{Adaptive-Random} (skyline for training time) &49.91&68.44&78.22&79.47&1.8730&9.258&18.163&56.914\\\cline{3-11}
 & &\textsc{Glister} &50.01&50.54&65.43&78.13&10.23&14.23&26.24&110.31\\
 & &\textsc{CraigPB}  &50.04&50.81&62.45&77.21&12.84&16.94&29.31&124.21\\
 & &\textsc{GradMatchPB}  &49.99&50.0&63.24&78.08&9.641&13.184&25.623&102.789\\
 & &\textsc{Milo} (Fixed)  &50.7&67.13&72.43&74.38&1.87&9.28&18.2&56.90\\
 & &\textsc{Milo}  &50.93&73.76&78.31&79.52&1.89&9.74&18.34&56.781\\ \hline \hline
 IMDB &BERT+MLP &\textsc{Full} (skyline for test accuracy)  &93.27&93.27&93.27&93.27&5695.329&5695.329&5695.329&5695.329\\ 
 & &\textsc{Random} (skyline for training time) &51.25&228.316&601.866&1672.21&87.362&89.58&90.33&91.91\\
 & &\textsc{Adaptive-Random} (skyline for training time) &51.25&228.316&601.866&1672.21&89.813&92.18&92.41&92.93\\\cline{3-11}
 & &\textsc{Glister} &86.25&91.21&92.01&92.43&721.43&789.21&898.31&1802.74\\
 & &\textsc{CraigPB}  &86.31&90.46&91.85&91.99&868.42&925.32&981.321&1831.31\\
 & &\textsc{GradMatchPB}  &86.54&90.42&91.99&91.83&682.987&758.215&892.21&1792.43\\
 & &\textsc{Milo} (Fixed)  &86.15&89.27&90.87&91.458&51.25&228.316&601.866&1672.21\\
 & &\textsc{Milo}  &90.76&92.06&92.6&93.144&51.25&228.316&601.866&1672.21\\ \hline \hline
\end{tabular}}
    \caption{Model Training Results for TREC6, IMDB and Rotten Tomatoes datasets}
    \label{tab:text_sel_results}
\end{sidewaystable}

\begin{table}[!htbp]
    \centering
    \scalebox{0.7}{
    \begin{tabular}{c c c|c c c c} \hline \hline
\multicolumn{7}{c}{Test Accuracy Standard Deviation on Text Datasets}\\ \hline
\multicolumn{3}{c|}{} & \multicolumn{4}{c|}{Standard deviation of the Model(for 5 runs)}\\ 
\multicolumn{1}{c}{} & \multicolumn{1}{c}{} & \multicolumn{1}{c|}{Budget(\%)} & \multicolumn{1}{c}{1\%} & \multicolumn{1}{c}{5\%} & \multicolumn{1}{c}{10\%} & \multicolumn{1}{c|}{30\%}\\ \hline
\multicolumn{1}{c}{Dataset} & \multicolumn{1}{c}{Model}  &\multicolumn{1}{c|}{Selection Strategy} & \multicolumn{4}{c|}{}  \\ \hline
TREC6 &LSTM &\textsc{Full} (skyline for test accuracy)  &0.77 &0.77  &0.77  &0.77 \\ 
 & &\textsc{Random} (skyline for training time) &3.41 &1.23  &0.57  &1.34 \\
 & &\textsc{Adaptive-Random} (skyline for training time) &3.2 &0.43  &1.31  &0.43  \\\cline{3-7}
 & &\textsc{Glister} &2.45 &0.74  &0.31  &0.54  \\
 & &\textsc{CraigPB}  &0.791 &0.462  &1.821  &0.861  \\
 & &\textsc{GradMatchPB}  &1.7265 &0.8762  &0.472  &0.876 \\
 & &\textsc{Milo} (Fixed)  &0.352 &0.62  &0.454  &0.726 \\
 & &\textsc{Milo}  &0.25 &0.36  &0.21  &0.31  \\ \hline \hline
IMDB &LSTM &\textsc{Full} (skyline for test accuracy)  &0.3 &0.3  &0.3  &0.3 \\ 
 & &\textsc{Random} (skyline for training time) &0.21 &0.31  &0.41  &1.2 \\
 & &\textsc{Adaptive-Random} (skyline for training time) &1.241 &0.21  &0.41  &0.21  \\\cline{3-7}
 & &\textsc{Glister} &0.862 &0.21  &0.21  &0.1  \\
 & &\textsc{CraigPB}  &0.731 & 1.21 &0.451  &0.41  \\
 & &\textsc{GradMatchPB}  &0.817 &0.441  &0.41  &1.22 \\
 & &\textsc{Milo} (Fixed)  &1.51 &0.42  &0.34  &0.51 \\
 & &\textsc{Milo}  &0.45 &0.21  &0.104  &0.05  \\ \hline \hline

 Rotten Tomatoes &LSTM &\textsc{Full} (skyline for test accuracy)  &0.34 & 0.34 &0.34 &0.34 \\ 
 & &\textsc{Random} (skyline for training time) &1.21 &1.41  &1.97  &0.31 \\
 & &\textsc{Adaptive-Random} (skyline for training time) &0.0871 &0.16  &0.31  &0.21  \\\cline{3-7}
 & &\textsc{Glister} &0.761 &0.43  &0.34  &0.31  \\
 & &\textsc{CraigPB}  &0.826 &0.12  &0.43  &0.41  \\
 & &\textsc{GradMatchPB}  &0.52 &0.2 & 0.21 &0.12 \\
 & &\textsc{Milo} (Fixed)  &0.7761 &0.31  &0.1  &0.31 \\
 & &\textsc{Milo}  &0.71 &0.41  &0.1  &0.21  \\ \hline \hline
  IMDB &BERT+MLP &\textsc{Full} (skyline for test accuracy)  &0.14 &0.14  &0.14  &0.14 \\ 
 & &\textsc{Random} (skyline for training time) &0.38 &0.981  &1.31  &0.21 \\
 & &\textsc{Adaptive-Random} (skyline for training time) &0.21 & 0.24 &0.31  &0.51  \\\cline{3-7}
 & &\textsc{Glister} &0.871 &0.31  &0.53  &0.72  \\
 & &\textsc{CraigPB}  &0.87 &0.66  &0.2  &0.731  \\
 & &\textsc{GradMatchPB}  &0.54 &0.53  &0.23  &0.61 \\
 & &\textsc{Milo} (Fixed)  &0.21 &0.21  &0.12 &0.3 \\
 & &\textsc{Milo}  &0.43 &0.12  &0.3  &0.12  \\ \hline \hline
\end{tabular}}
    \caption{Standard Deviation Results for TREC6, IMDB and Rotten Tomatoes datasets}
    \label{tab:text_std_results}
\end{table}

\subsection{Hyper-parameter Ordering Retention Analysis} \label{app:hp_ordering_retention}
We evaluate the effectiveness of \model\ and the considered subset selection baselines in preserving original hyper-parameter ordering by full data tuning. In a nutshell, we are trying to determine whether the original order of the hyper-parameters is maintained, despite using small subsets for model training runs involved in hyper-parameter tuning. To examine this, we experiment on the  Trec6~\citep{trec1, trec2} dataset using an LSTM model. Hyper-parameter search for the TREC6 dataset includes a grid search over 108 configurations of the learning rate, optimizer, LSTM hidden size, training batch size, and the number of final fully connected layers. Table~\ref{tab:kendall_tau} shows the Kendall Tau correlation values between the hyper-parameter ordering obtained using 1\%, 5\%, and 10\% subsets selected by \model{}, \textsc{Random}, \textsc{Adaptive-Random}, \textsc{Automata}, and \textsc{CraigPB} and Full data hyper-parameter ordering on TREC6~\citep{trec1, trec2} dataset. Results in Table~\ref{tab:kendall_tau} demonstrate that \model{} is more effective than the baselines considered in preserving hyper-parameter ordering even when using small subsets.

\begin{table}[!ht]
\centering
\begin{tabular}{c c c c| c} \hline \hline
\multicolumn{5}{c}{Hyper-parameter Ordering Retention Capabilit}\\ \hline
\hline
\multicolumn{4}{c|}{} & \multicolumn{1}{c}{Kendall Tau Values} \\ 
\multicolumn{1}{c}{Dataset} & \multicolumn{1}{c}{Model}  &\multicolumn{1}{c}{Budget} & \multicolumn{1}{c|}{Strategy} & \multicolumn{1}{c}{} \\ \hline
TREC6 & LSTM &  1\%   & \model{} & \textbf{0.4321}\\
& & & \textsc{Random} & 0.0679 \\
& & & \textsc{Adaptive-Random} & 0.313  \\
& & & \textsc{Automata} & 0.3484 \\
& & & \textsc{CraigPB} & 0.325 \\ \cline{3-5}
& &  5\%   & \model{} & \textbf{0.521}  \\
& & & \textsc{Random} & 0.199 \\
& & & \textsc{Adaptive-Random} & 0.440\\
& & & \textsc{Automata} &  0.4764\\
& & & \textsc{CraigPB} & 0.4521 \\ \cline{3-5}
& &  10\%   & \model{} & \textbf{0.6342}\\
& & & \textsc{Random} & 0.2605\\
& & & \textsc{Adaptive-Random} & 0.5313\\
& & & \textsc{Automata} & 0.4742 \\
& & & \textsc{CraigPB} & 0.4531 \\\hline
\end{tabular}
\caption{Mean test set accuracy of ResNet18 trained on \textsc{Cifar100} dataset for subset sizes of 10\%, and 30\% selected using \model{} for 200 epochs for different values of $R$.}
\label{tab:kendall_tau}
\end{table}

\subsection{Test-Accuracies, Training times, and Standard deviations for Hyper-Parameter Tuning Experiments} \label{app:hp_std_results}
Table~\ref{tab:hp_tuning_results} shows the top-1 test accuracies and tuning times taken by \model{} and the other baselines considered for single model training on \textsc{Cifar10}, TREC6 datasets for different subset sizes of 1\%, 5\%, 10\%, and 30\% respectively. 

\begin{sidewaystable}[!htbp]
    \centering
    \scalebox{0.7}{
    \begin{tabular}{c c c c|c c c c|c c c c} \hline \hline
\multicolumn{12}{c}{Hyper-parameter Tuning Results}\\ \hline
\multicolumn{4}{c|}{} & \multicolumn{4}{c|}{Top-1 Test accuracy of the Model(\%)} & \multicolumn{4}{c}{Tuning time(in hrs)} \\ 
\multicolumn{1}{c}{} & \multicolumn{1}{c}{} & \multicolumn{1}{c}{} & \multicolumn{1}{c|}{Budget(\%)} & \multicolumn{1}{c}{1\%} & \multicolumn{1}{c}{5\%} & \multicolumn{1}{c}{10\%} & \multicolumn{1}{c|}{30\%} & \multicolumn{1}{c}{1\%} & \multicolumn{1}{c}{5\%} & \multicolumn{1}{c}{10\%} & \multicolumn{1}{c}{30\%} \\ \hline
\multicolumn{1}{c}{Dataset} & \multicolumn{1}{c}{Model}  & \multicolumn{1}{c}{Search+Scheduler}  &\multicolumn{1}{c|}{Selection Strategy} & \multicolumn{4}{c|}{} & \multicolumn{4}{c}{} \\ \hline
TREC6 & LSTM & Random+HB&\textsc{Full} (skyline for test accuracy)  &86.8&86.8&86.8&86.8&0.64&0.64&0.64&0.64 \\ 
 & & &\textsc{Random} (skyline for training time) &40.8&60.2 &78.0&85.5&0.01&0.03&0.07&0.18\\
 & & &\textsc{Adaptive-Random} (skyline for training time) 45.76&49.43&74.32&86.6&0.01&0.03&0.06&0.18\\ \cline{4-12}
 & & &\textsc{CraigPB}  &41.33&64.33&77.0&82.0&0.02&0.07&0.08&0.19\\
 & & &\textsc{Automata}  &47.3&66.2&82.0&86.6&0.01&0.06&0.09&0.21 \\
 & & &\textsc{Milo} (Fixed)  &54.32&79.83&85.51&86.43&0.01&0.04&0.06&0.18 \\
 & & &\textsc{Milo} &76.2&86.6&86.54&86.6&0.01&0.03&0.06&0.18\\ \hline \hline
 
 TREC6 & LSTM & TPE+HB&\textsc{Full} (skyline for test accuracy)  &84.4&84.4&84.4&84.4&0.64&0.64&0.64&0.64\\ 
 & & &\textsc{Random} (skyline for training time) &40.8&60.2&78&85.5&0.01&0.03&0.06&0.18\\
 & & &\textsc{Adaptive-Random} (skyline for training time) &52.32&64.32&79.2&83.42&0.01&0.04&0.06&0.18 \\\cline{4-12}
 & & &\textsc{CraigPB}  &41.33&64.33&77&82&0.02&0.07&0.08&0.19\\
 & & &\textsc{Automata}  &47.2&66.2&82&86.6&0.01&0.06&0.09&0.21\\
 & & &\textsc{Milo} (Fixed)  &54.32&79.83&82.51&85.43&0.01&0.03&0.06&0.18\\
 & & &\textsc{Milo} &68.2 &84.26 &84.34 &86.46 &0.01&0.03&0.06&0.18\\ \hline \hline
 
CIFAR10 & ResNet18 & Random+HB&\textsc{Full} (skyline for test accuracy)  &95.35&95.35&95.35&95.35&36.11&36.11&36.11&36.11 \\ 
 & & &\textsc{Random} (skyline for training time) &87.21&88.95&90.21&94.21&0.4779&1.8681&3.7917&10.900\\
 & & &\textsc{Adaptive-Random} (skyline for training time) &93.1&94.32&94.78&95.21&0.47790&1.86814&3.79172&10.9002\\\cline{4-12}
 & & &\textsc{CraigPB}  &92.43&93.5&94.2&94.3&0.5434&1.92925&3.83061&11.06829\\
 & & &\textsc{Automata}  &93.7&93.8&94.21&95.31&0.67548&1.91481&3.81394&11.0127 \\
 & & &\textsc{Milo} (Fixed)  &86.21&88.95&94.21&95.21&0.47790&1.868144&3.791&10.909\\
 & & &\textsc{Milo} &94.34&95.34&95.21&95.34&0.4778&1.8644&3.796&10.9889\\ \hline \hline
 
 CIFAR10 & ResNet18 & TPE+HB&\textsc{Full} (skyline for test accuracy)  &95.24&95.24&95.24&95.24&36.12&36.12&36.12&36.12 \\ 
 & & &\textsc{Random} (skyline for training time) &86.38&89.32&92.21&94.21&0.48&1.87&3.79&10.9 \\
 & & &\textsc{Adaptive-Random} (skyline for training time) &93.12&93.72&94.78&95.21&0.49&1.89&3.78&11.1\\\cline{4-12}
 & & &\textsc{CraigPB}  & 92.23&93.5&94.2&94.3&0.54&1.93&3.83&11.07 \\
 & & &\textsc{Automata}  & 93.7&93.98&94.21&95.31&0.68&1.91&3.81&11.01\\
 & & &\textsc{Milo} (Fixed)  & 87.11&90.21&94.21&95.21&0.48&1.87&3.79&10.9\\
 & & &\textsc{Milo} & 93.24&94.28&95.1&95.21&0.52&1.89&3.77&10.89\\ \hline \hline
\end{tabular}}
    \caption{Hyper-parameter Tuning Results}
    \label{tab:hp_tuning_results}
\end{sidewaystable}

\section{Ablation Studies} \label{app:ablation}
\subsection{Performance Comparison of Existing Pre-trained Transformer Models as Feature Encoders}
\label{app:feature_encoders}
\begin{figure}[!ht]
  \centering
  \begin{subfigure}[b]{0.48\columnwidth}
    \includegraphics[width=\linewidth]{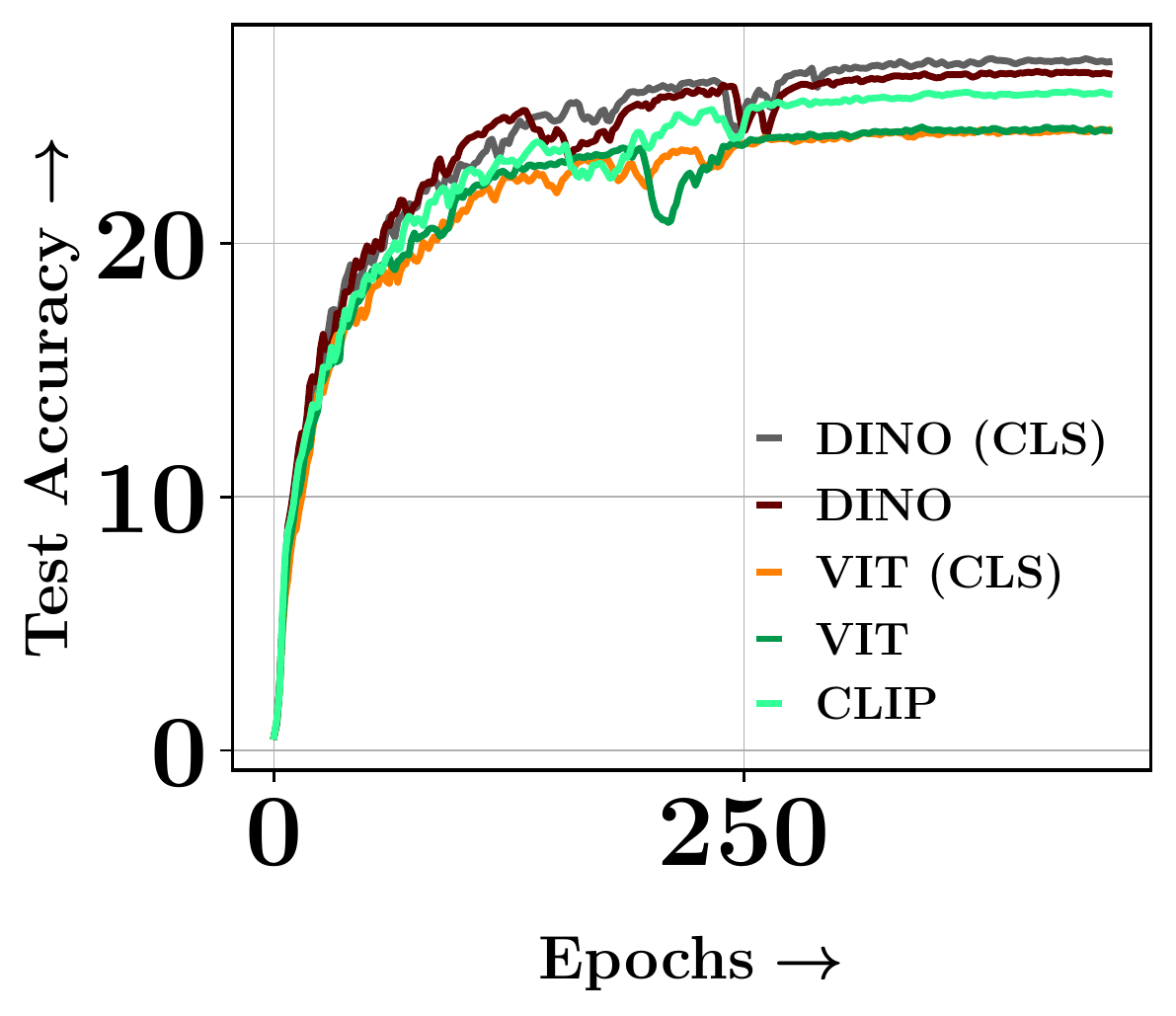}
    \caption{Comparison of Vision Transformers for Subset Selection}
    \label{subfig:vision_transformers_comp}
  \end{subfigure}
  \hfill 
  \begin{subfigure}[b]{0.48\columnwidth}
    \includegraphics[width=\linewidth]{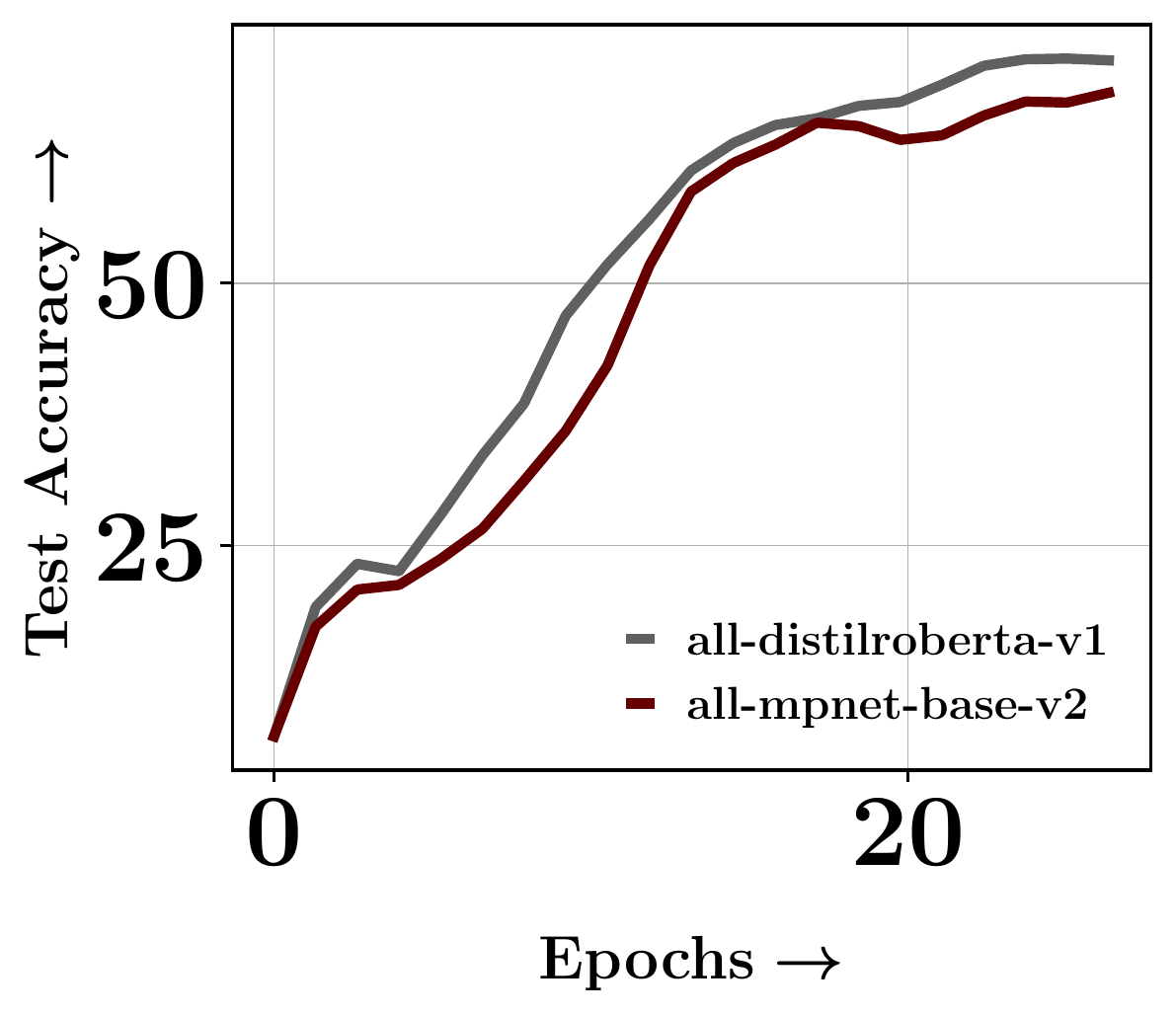}
    \caption{Comparison of Language Models for Subset Selection}
    \label{subfig:language_models_comp}
  \end{subfigure}
  \caption{\footnotesize{Sub-figure (a) shows the performance of the ResNet18 model trained on a fixed 5\% subset of the \textsc{Cifar100} dataset selected by maximizing Facility Location function using different pre-trained vision transformer models. Sub-figure (b) shows the performance of the LSTM model trained on a fixed 5\% subset of the TREC6 dataset selected by maximizing the Facility Location function using different pre-trained language models.}}
    \label{fig:feature_encoder_comp}
\end{figure}
In this ablation study, we evaluate the effectiveness of various existing pre-trained transformer models as feature encoders for subset selection.

\noindent \textbf{Optimal Feature Encoders for Vision Datasets: } For vision datasets, we test with \textsc{Dino-ViTb16}~\citep{Caron2021EmergingPI}, \textsc{vit-large-patch16-224-in21k}~\citep{dosovitskiy2021an}, and \textsc{clip-ViT-L-14}~\citep{CLIP} models available from the HuggingFace~\citep{huggingface} as the feature encoder. For CLIP model, we use CLIP VIT's final projection layer output embedding as the feature representation. For both DINO and \textsc{ViT} models, we test with two different types of embeddings as feature vectors: a) we use the final layer CLS token embedding output as the feature representation; b) we use the average of the final layer output embeddings of the constituent patches as the feature representation. We denote the DINO model using CLS token output as a feature representation vector by \textsc{DINO (CLS)}. Similarly, we denote the \textsc{ViT} model using CLS token output as a feature representation vector by \textsc{ViT (CLS)}. Sub-figure~\ref{subfig:vision_transformers_comp} shows the performance of the ResNet18 model trained on a fixed 5\% subset of the \textsc{Cifar100} dataset selected by maximizing the facility location function using different pre-trained vision transformer models as the feature encoder. In this experiment, we used the facility location set function because when using fixed subsets of small subset sizes, facility location yielded optimal results (See Sub-figure~\ref{subfig:0.1_submod_comparison}). The results presented in Sub-figure~\ref{subfig:vision_transformers_comp} indicate that the DINO model using the final layer CLS token embedding as the feature representation gave the best model performance. Hence, in our experiments, we use the DINO model and compute the feature representations for images by using the final layer CLS token output embedding.

\noindent \textbf{Optimal Feature Encoders for Text Datasets: } For text datasets, we test with \textsc{all-distilroberta-v1}~\citep{liu2020roberta}, and \textsc{all-mpnet-base-v2}~\citep{mpnet} models available from the the Sentence Transformers~\citep{reimers-2019-sentence-bert} as the feature encoder. We employ pre-trained models from the Sentence Transformers package, as they have been fine-tuned to produce improved representations that accurately compute the similarity between phrases for natural language inference (NLI) tasks. Similar to the SBERT~\citep{reimers-2019-sentence-bert} work, we compute the sentence representations by taking the average of the final layer output embeddings of all the constituent tokens. Sub-figure~\ref{subfig:vision_transformers_comp} shows the performance of the LSTM model trained on a fixed 5\% subset of the TREC6 dataset selected by maximizing the facility location function using different pre-trained language models as the feature encoder. In this experiment, we used the facility location set function because when using fixed subsets of small subset sizes, facility location yielded optimal results (See Sub-figure~\ref{subfig:0.1_submod_comparison}). Results presented in the Sub-figure~\ref{subfig:language_models_comp} show that \textsc{all-distilroberta-v1} resulted in better model performance compared to \textsc{all-mpnet-base-v2}. Hence, in our experiments, we use the \textsc{all-distilroberta-v1} as the feature encoder for text datasets.

\subsection{Optimal Similarity Metric Analysis}
In this experiment, we aim to analyze the performance of different similarity metrics for the computation of similarity between data samples. As part of our experiment, we evaluate the following similarity metrics:

\noindent \emph{Cosine Similarity: }
Given two data samples $e_1$, $e_2$ having feature representations $r_1$, $r_2$, the similarity between the data samples using the cosine-similarity is as follows:
\begin{align}
    \operatorname{Cosine-Similarity}(r_1, r_2) = 0.5 + 0.5 \cdot \frac{r_1 \cdot r_2}{||r_1|| ||r_2||}
\end{align}

We perform additive scaling of the cosine similarity to obtain non-negative similarity values.

\noindent \emph{Dot Product: }
Given two data samples $e_1$, $e_2$ having feature representations $r_1$, $r_2$, the similarity between the data samples using the dot-product is $r_1 \cdot r_2$. We also perform additive scaling to the obtained dot product values to ensure that all the pair-wise similarity values are non-negative.

\noindent \emph{RBF Kernel: }
Given two data samples $e_1$, $e_2$ having feature representations $r_1$, $r_2$, the similarity between the data samples using the RBF kernel is as follows:
\begin{align}
    \operatorname{RBF-Kernel}(r_1, r_2) = \exp{(-\frac{||r_1 - r_2||^2}{kw * mean\_dist})}
\end{align}

In the above equation, the parameter $kw$ is a hyper-parameter that controls the saturation of the similarity between data samples and the parameter $mean\_dist$ is the mean distance of the pair-wise distance between all the samples in the dataset. In the case of small $kw$ values, the similarity between samples of data drops dramatically with even a small increase in distance between them. In contrast with high $kw$ values, the similarity between samples of data drops slowly with an increase in distance between them. We also test with different $kw$ values of 0.01, 0.05, 0.1, 0.5, and 1 in this experiment.

\noindent \textbf{Similarity Metric for Vision Datasets: } Table~\ref{tab:vision_similarity} shows the test accuracies of a ResNet18 model trained on a 5\% fixed \textsc{Cifar100} subset selected by maximizing the facility location function using the DINO model with final layer CLS token outputs as feature representations using different similarity metrics.  Results demonstrate that with DINO model using cosine-similarity resulted in a better-performing model compared to other similarity metrics. Hence in our work for vision experiments, we use cosine-similarity as the similarity metric. We do not show results with dot-product in Table~\ref{tab:vision_similarity} as the CLS token embedding outputs of the DINO model are normalized because of which both dot-product and cosine similarity gives the same similarity values.

\begin{table}[!ht]
    \centering
    \begin{tabular}{|c|c|}
    \hline
        \textbf{Similarity Metric} & \textbf{Test Accuracy} \\ \hline
        Cosine Similarity & \textbf{28.68 $\pm$ 0.1103} \\ \hline
        RBF Kernel ($kw$=0.01) & 27.14 $\pm$ 0.2051 \\ \hline
        RBF Kernel ($kw$=0.05) & 27.88 $\pm$ 0.8132 \\ \hline
        RBF Kernel ($kw$=0.1) & 28.11 $\pm$ 0.1499 \\ \hline
        RBF Kernel ($kw$=0.5) & 27.61 $\pm$ 0.601 \\ \hline
        RBF Kernel ($kw$=0.1) & 27.62 $\pm$ 0.6576 \\ \hline
    \end{tabular}
    \caption{Comparison of accuracies of ResNet18 model trained on a 5\% fixed \textsc{Cifar100} subset selected by maximizing the facility location function using the DINO model with CLS token outputs as feature representations using different similarity metrics.}
    \label{tab:vision_similarity}
\end{table}

\noindent \textbf{Similarity Metric for Text Datasets: } Table~\ref{tab:text_similarity} shows the test accuracies of the LSTM model trained on a 5\% fixed TREC6 subset selected by maximizing the facility location function using the \textsc{all-distilroberta-v1} as the feature encoder using different similarity metrics.  Results demonstrate that with \textsc{all-distilroberta-v1} model using cosine-similarity resulted in better performing model compared to other similarity meterics. Hence in our work for text experiments, we use cosine-similarity as the similarity metric. 

\begin{table}[!ht]
    \centering
    \begin{tabular}{|c|c|}
    \hline
        \textbf{Similarity Metric} & \textbf{Test Accuracy} \\ \hline
        Cosine Similarity & \textbf{72.1 $\pm$ 2.1508} \\ \hline
        Dot Product &  70.38 $\pm$ 4.5514\\ \hline
        RBF Kernel ($kw$=0.01) & 63.58 $\pm$ 9.8051 \\ \hline
        RBF Kernel ($kw$=0.05) & 67.58 $\pm$ 2.561 \\ \hline
        RBF Kernel ($kw$=0.1) & 66.92 $\pm$ 2.715 \\ \hline
        RBF Kernel ($kw$=0.5) & 70.7 $\pm$ 2.291 \\ \hline
        RBF Kernel ($kw$=0.1) & 70.54 $\pm$ 4.518 \\ \hline
    \end{tabular}
    \caption{Comparison of accuracies of LSTM model trained on a 5\% fixed TREC6 subset selected by maximizing the facility location function using the \textsc{all-distilroberta-v1} as the feature encoder using different similarity metrics.}
    \label{tab:text_similarity}
\end{table}

\subsection{SGE (Graph-Cut) vs SGE (Facility Location)} \label{app:sge_gc_vs_sge_fl}
\begin{figure}[!ht]
\centering
\hspace{-0.6cm}
\begin{subfigure}[b]{0.32\textwidth}
\centering
\includegraphics[width=4.5cm]{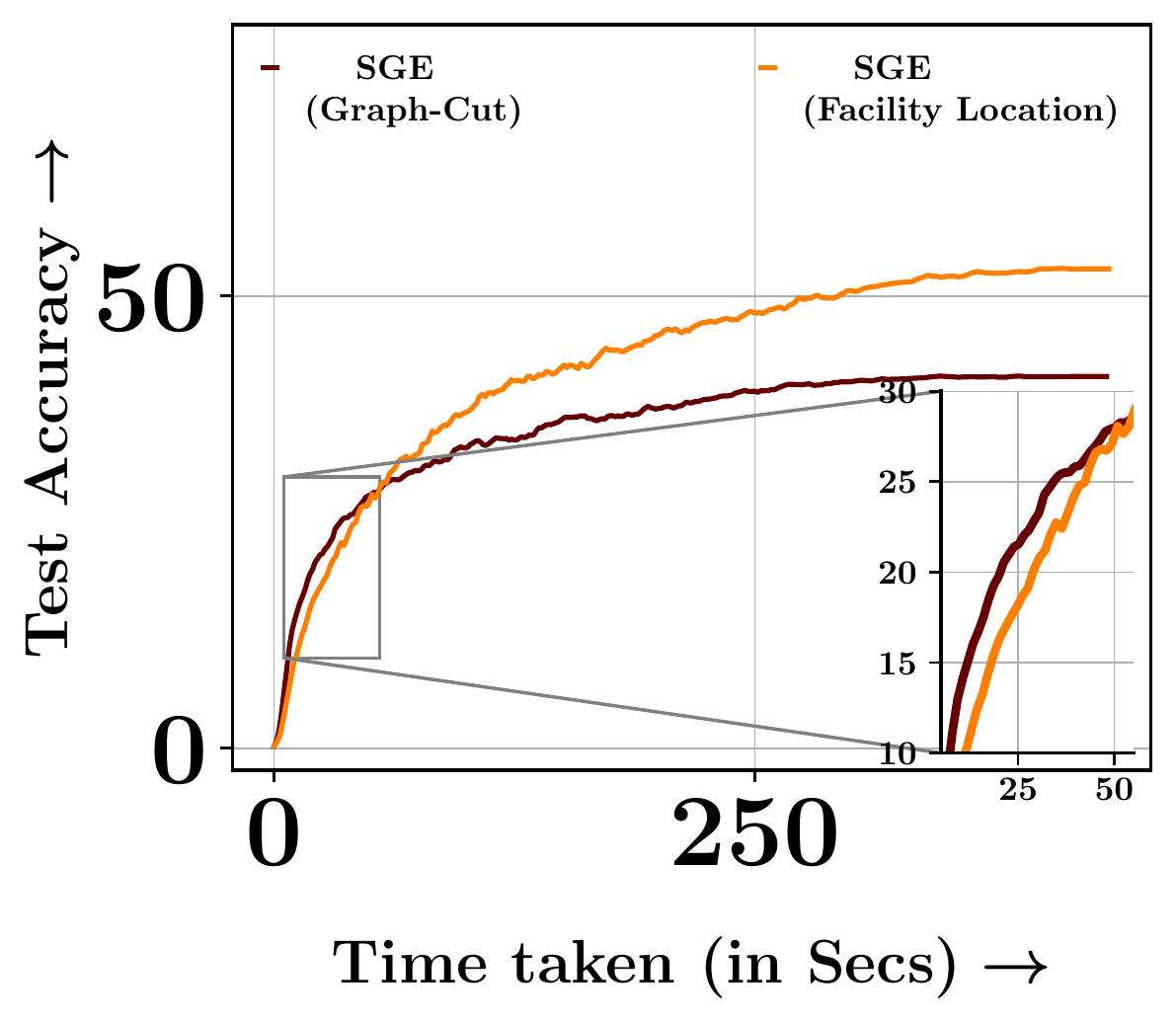}
\caption*{(a) \scriptsize \textsc{Cifar100}(5\%) \\ ResNet18}
\phantomcaption
\label{fig:cifar100_sge}
\end{subfigure}
\begin{subfigure}[b]{0.32\textwidth}
\centering
\includegraphics[width=4.5cm]{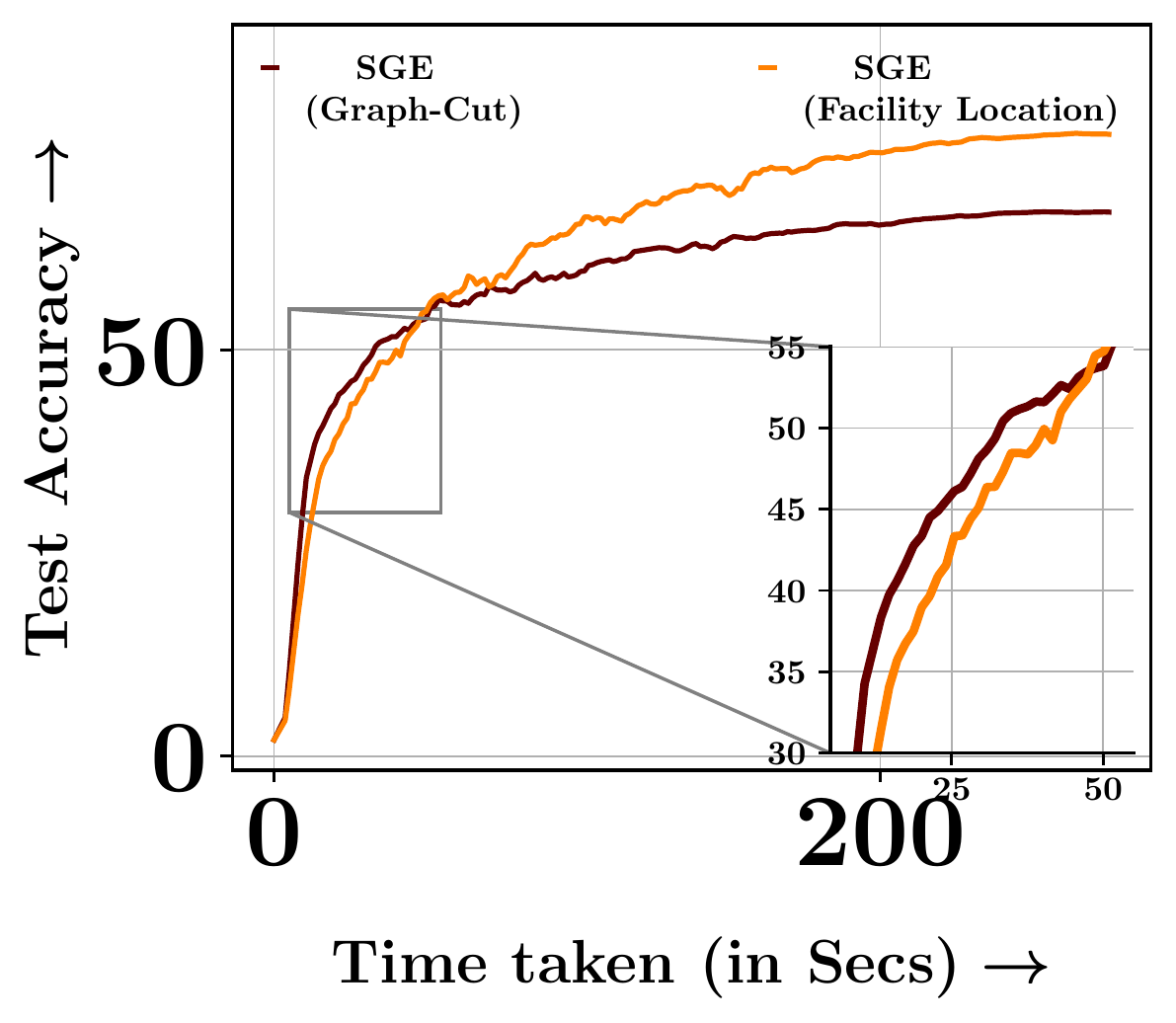}
\caption*{(b) \scriptsize \textsc{Cifar10}(5\%) \\ ResNet18}
\phantomcaption
\label{fig:cifar10_sge}
\end{subfigure}
\begin{subfigure}[b]{0.32\textwidth}
\centering
\includegraphics[width=4.5cm]{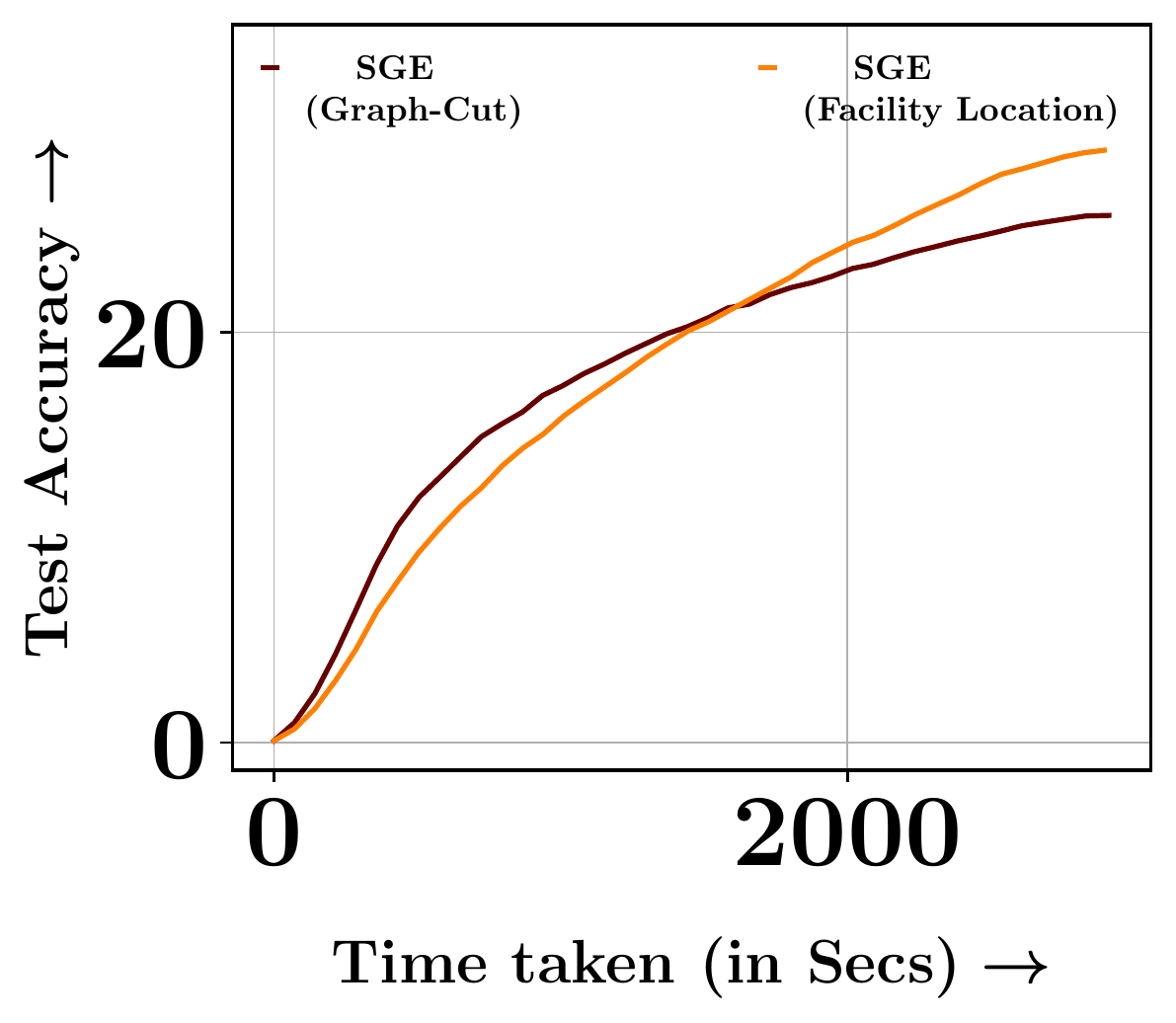}
\caption*{(c) \scriptsize \textsc{TinyImageNet}(5\%) \\ ResNet18}
\phantomcaption
\label{fig:tinyimagenet_sge}
\end{subfigure}
\begin{subfigure}[b]{0.32\textwidth}
\centering
\includegraphics[width=4.5cm]{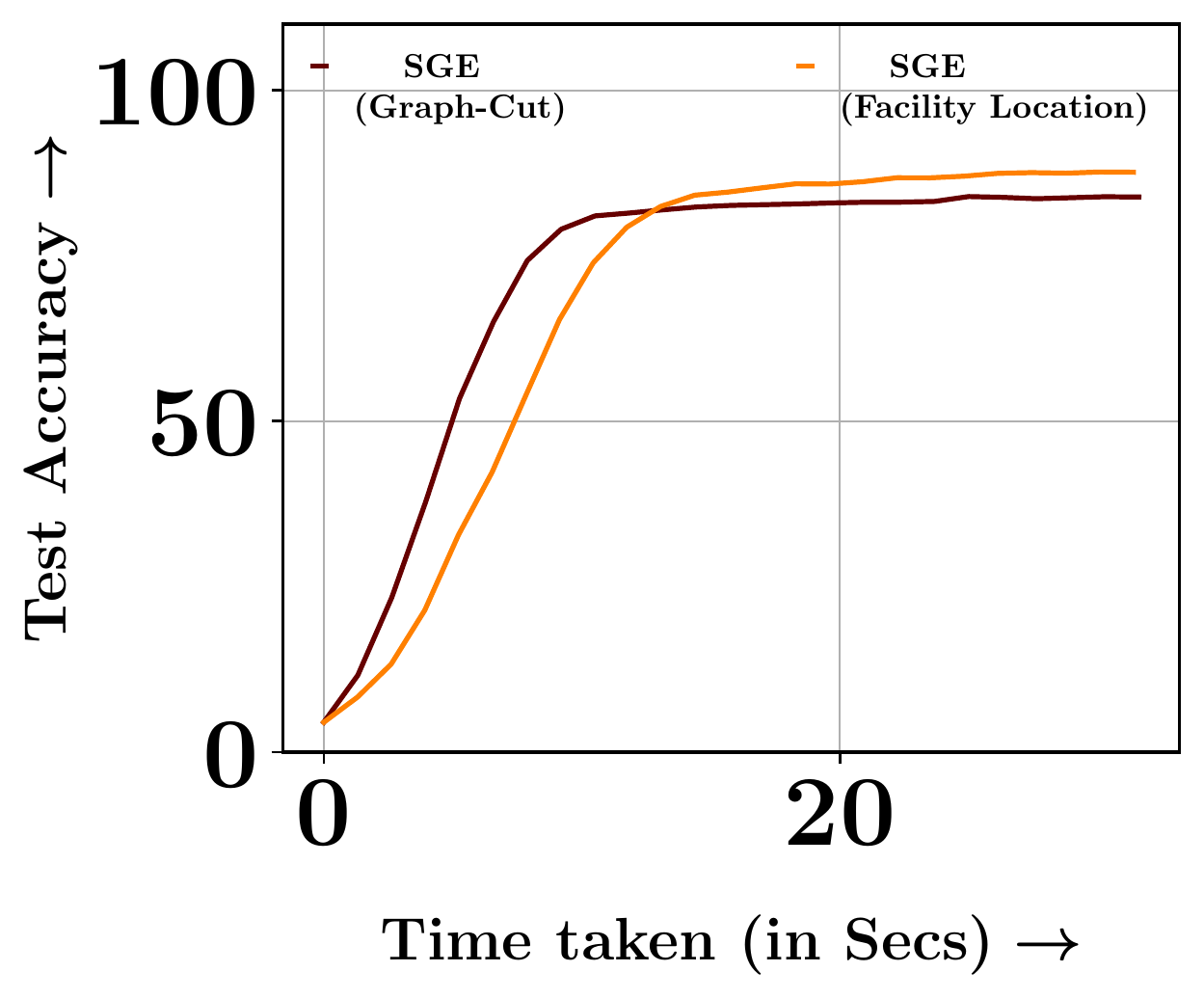}
\caption*{(d) \scriptsize TREC6(30\%) \\ LSTM}
\phantomcaption
\label{fig:trec6_sge}
\end{subfigure}
\begin{subfigure}[b]{0.32\textwidth}
\centering
\includegraphics[width=4.5cm]{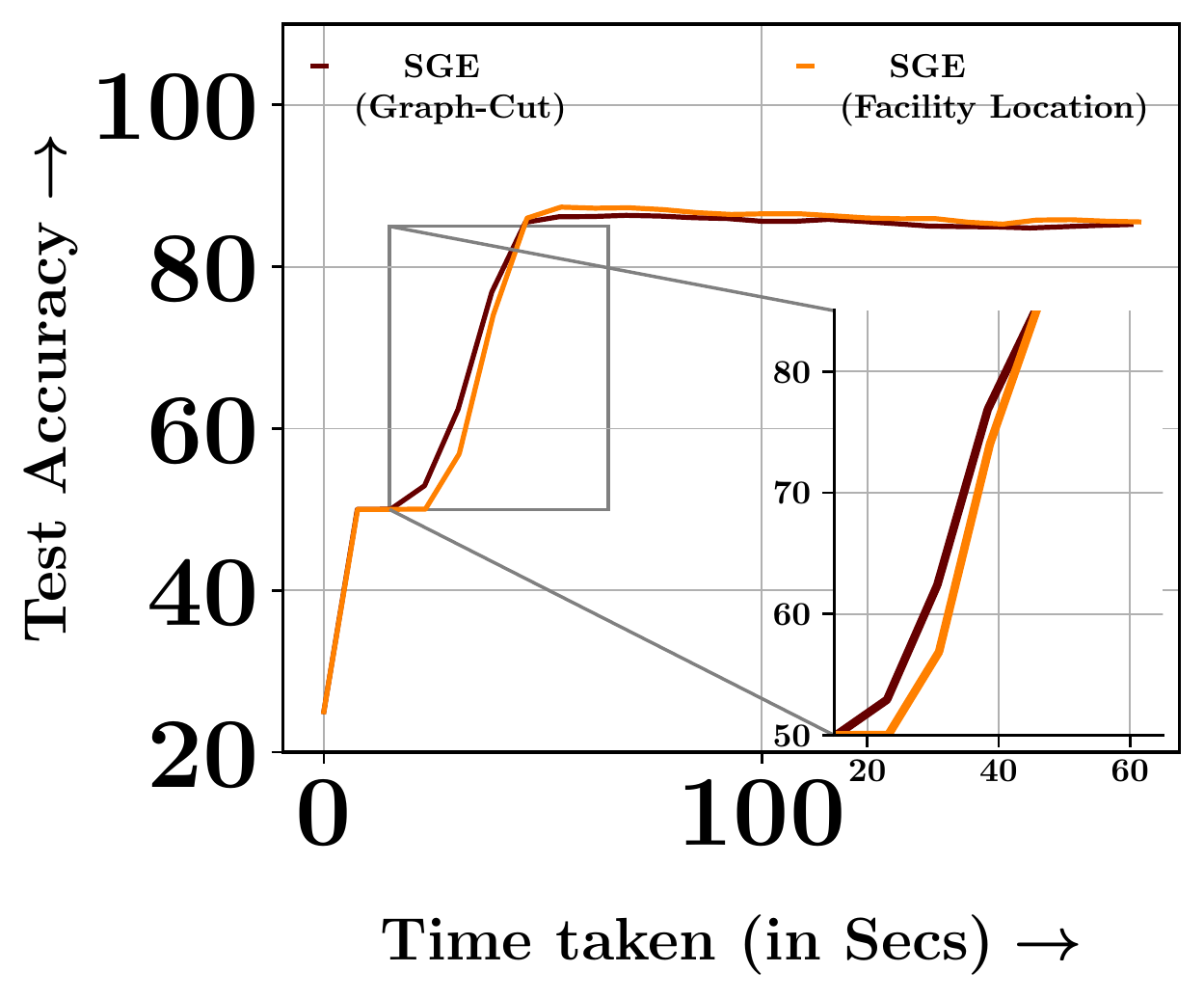}
\caption*{(e) \scriptsize IMDB (30\%) \\ LSTM}
\phantomcaption
\label{fig:imdb_sge}
\end{subfigure}
\begin{subfigure}[b]{0.32\textwidth}
\centering
\includegraphics[width=4.5cm]{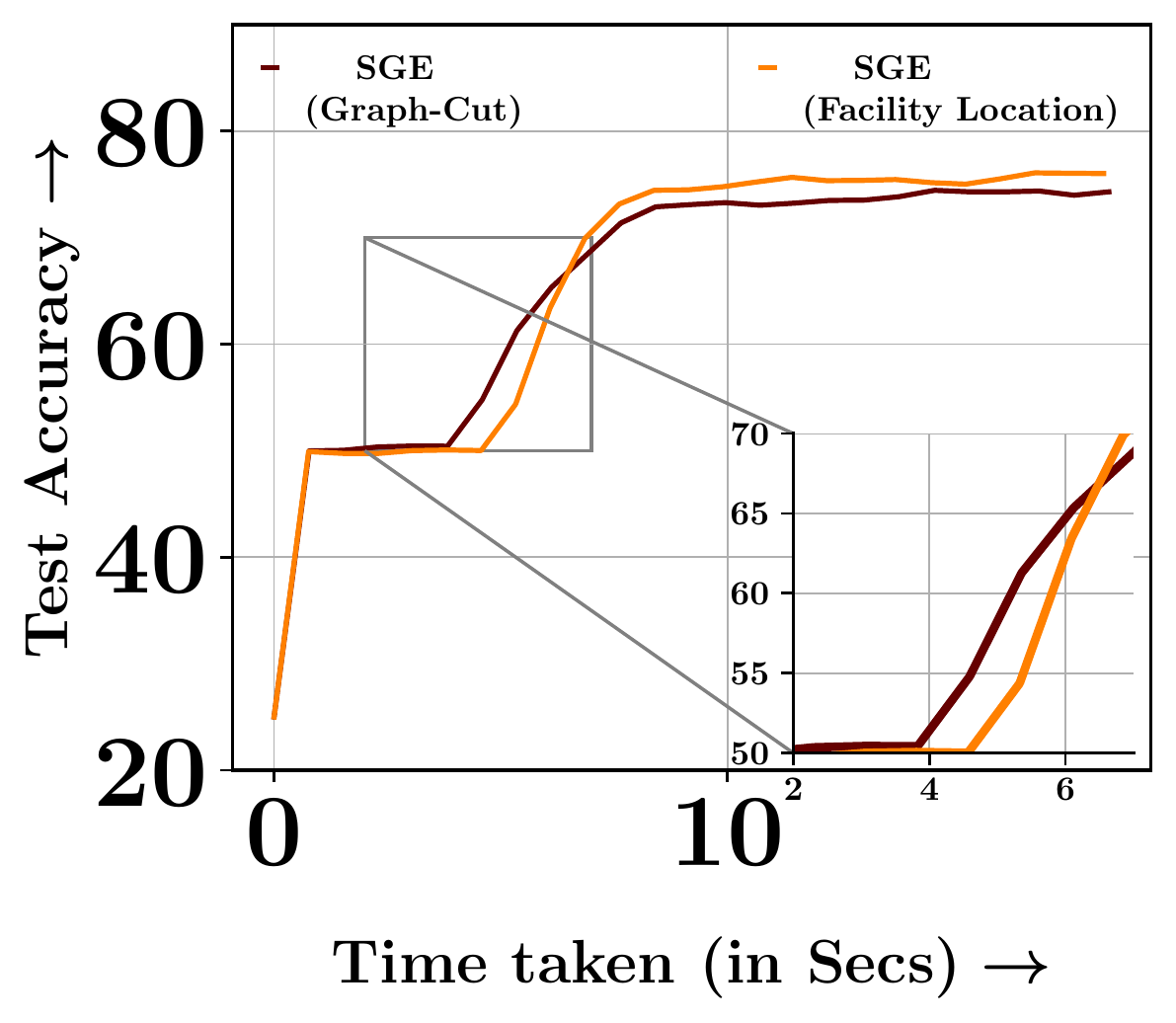}
\caption*{(f) \scriptsize Rotten Tomatoes (10\%) \\ LSTM}
\phantomcaption
\label{fig:rotten_tomatoes_sge}
\end{subfigure}
\caption{\footnotesize{Comparison of initial convergence of SGE with Graph-Cut and SGE with Facility Location on a variety of datasets using different subset sizes. Results show that SGE with Graph-Cut achieves faster initial convergence compared to SGE with Facility Location across all the datasets considered and for different subset sizes.}}
\label{fig:sge_gc_vs_sge_fl}
\end{figure}

Figure~\ref{fig:sge_gc_vs_sge_fl} shows the model convergence curves trained using SGE with Graph-Cut and SGE with Facility Location on various datasets using different subset sizes. Results show that SGE with Graph-Cut achieves faster initial convergence than SGE with Facility Location across all the datasets considered and for different subset sizes. The reason why SGE with Graph-Cut gives superior initial model convergence compared to SGE with Facility Location is that the sum-sum formulation in graph-cut(~\eqref{eq:gc}), as opposed to the sum-max formulation in facility location(~\eqref{eq:fl}), results in the selection of more number of easy samples. More specifically, with the sum-max formulation for facility location, having a single representative sample of data from each cluster in the dataset is sufficient to achieve a high facility location value for that cluster and having a representative sample from all the clusters in the dataset leads to further maximization of facility location values. This prevents the selection of samples only from densely populated regions while achieving broader coverage of the selected dataset in the facility location. Whereas, with the sum-sum formulation of the graph-cut function, selecting more samples from dense regions, leads to the high graph-cut function values. Thus, graph-cut can result in the selection of subsets consisting of higher number of easy samples from very dense regions in the dataset than facility location.

\subsection{SGE (Graph-Cut) vs. WRE (Graph-Cut)} \label{app:sge_gc_vs_wre_gc}
\begin{figure}[!ht]
\centering
\hspace{-0.6cm}
\begin{subfigure}[b]{0.32\textwidth}
\centering
\includegraphics[width=4.5cm]{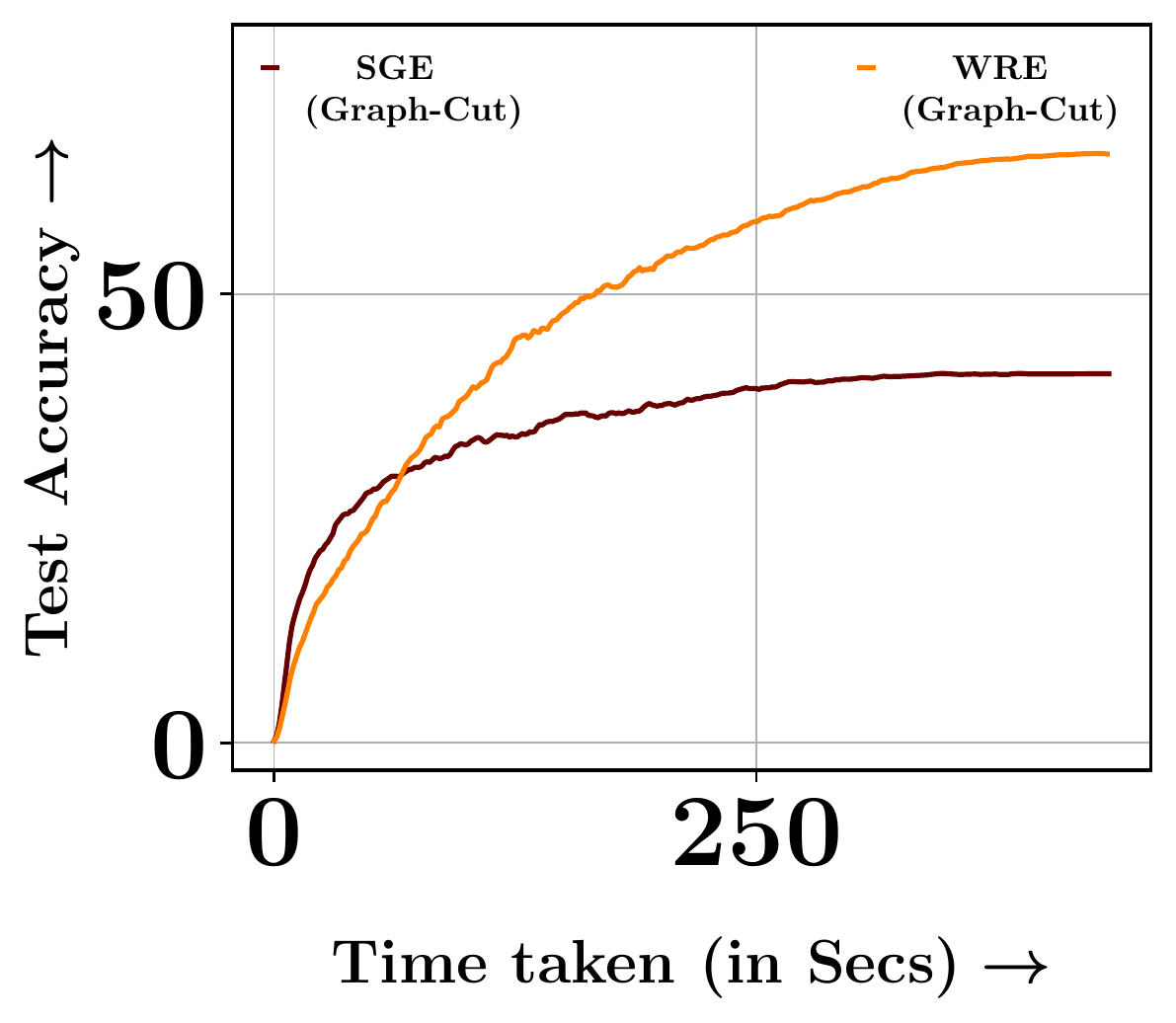}
\caption*{(a) \scriptsize \textsc{Cifar100}(5\%) \\ ResNet18}
\phantomcaption
\label{fig:cifar100_wre}
\end{subfigure}
\begin{subfigure}[b]{0.32\textwidth}
\centering
\includegraphics[width=4.5cm]{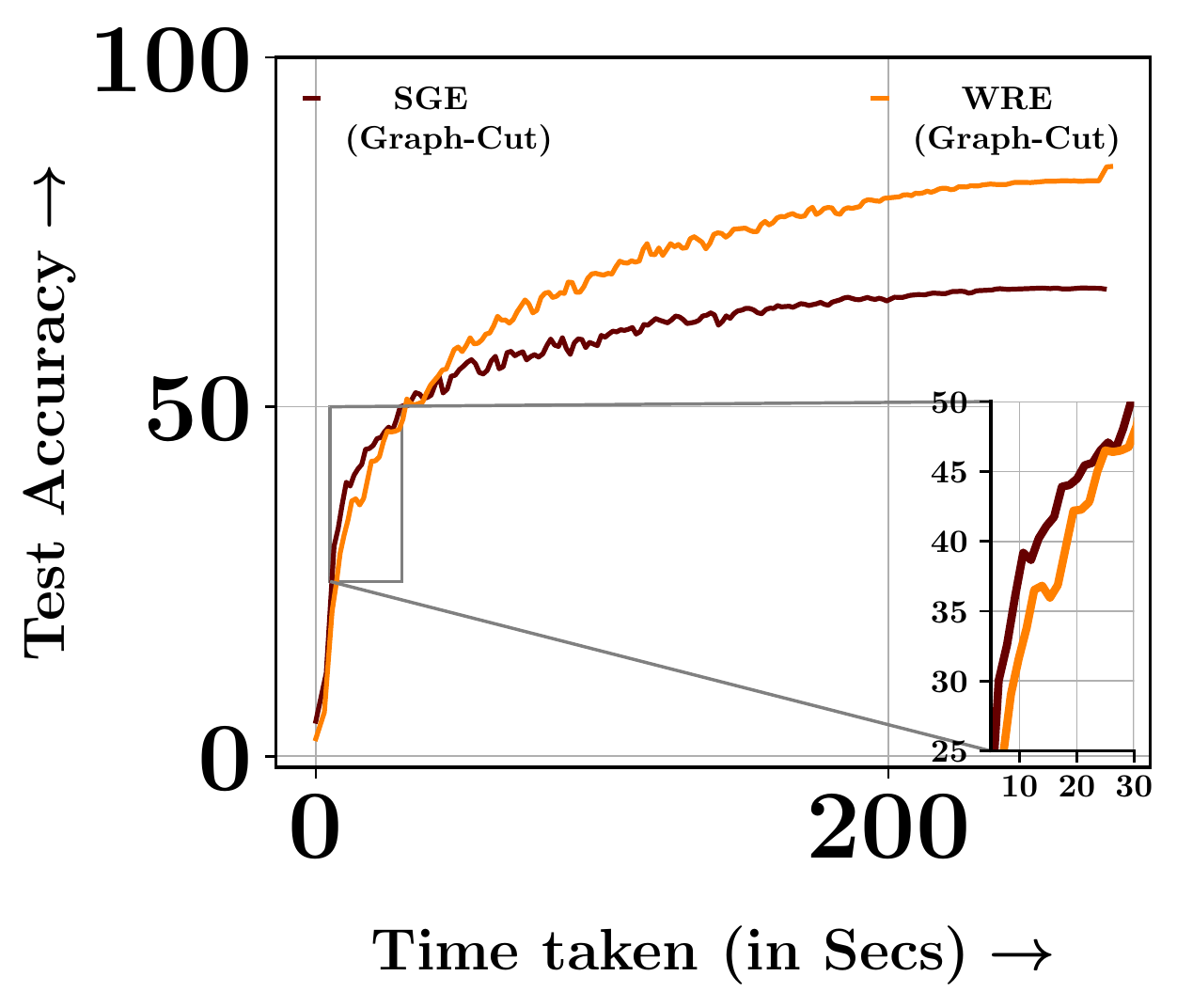}
\caption*{(b) \scriptsize \textsc{Cifar10}(5\%) \\ ResNet18}
\phantomcaption
\label{fig:cifar10_wre}
\end{subfigure}
\begin{subfigure}[b]{0.32\textwidth}
\centering
\includegraphics[width=4.5cm]{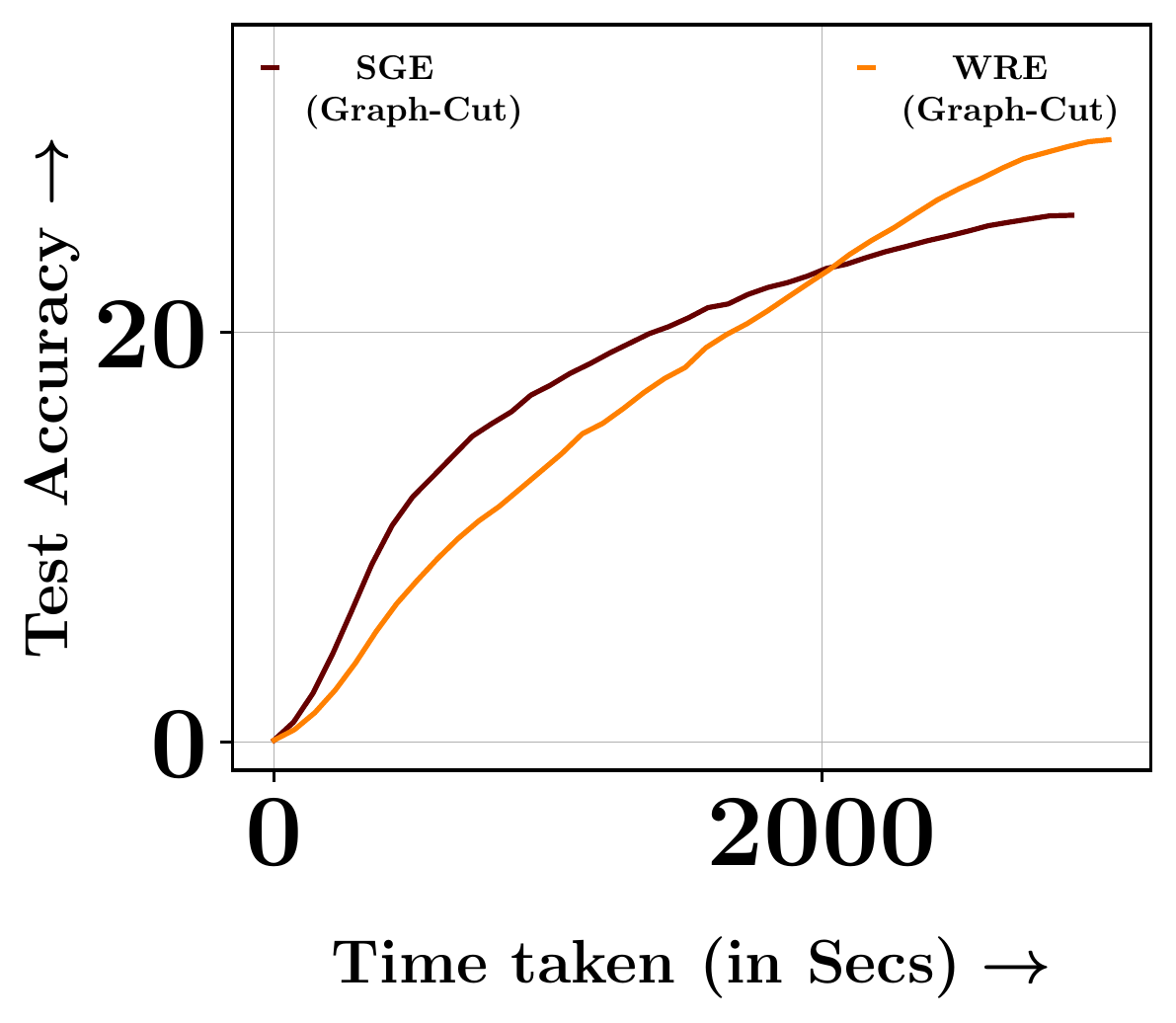}
\caption*{(c) \scriptsize \textsc{TinyImageNet}(5\%) \\ ResNet18}
\phantomcaption
\label{fig:tinyimagenet_wre}
\end{subfigure}
\begin{subfigure}[b]{0.32\textwidth}
\centering
\includegraphics[width=4.5cm]{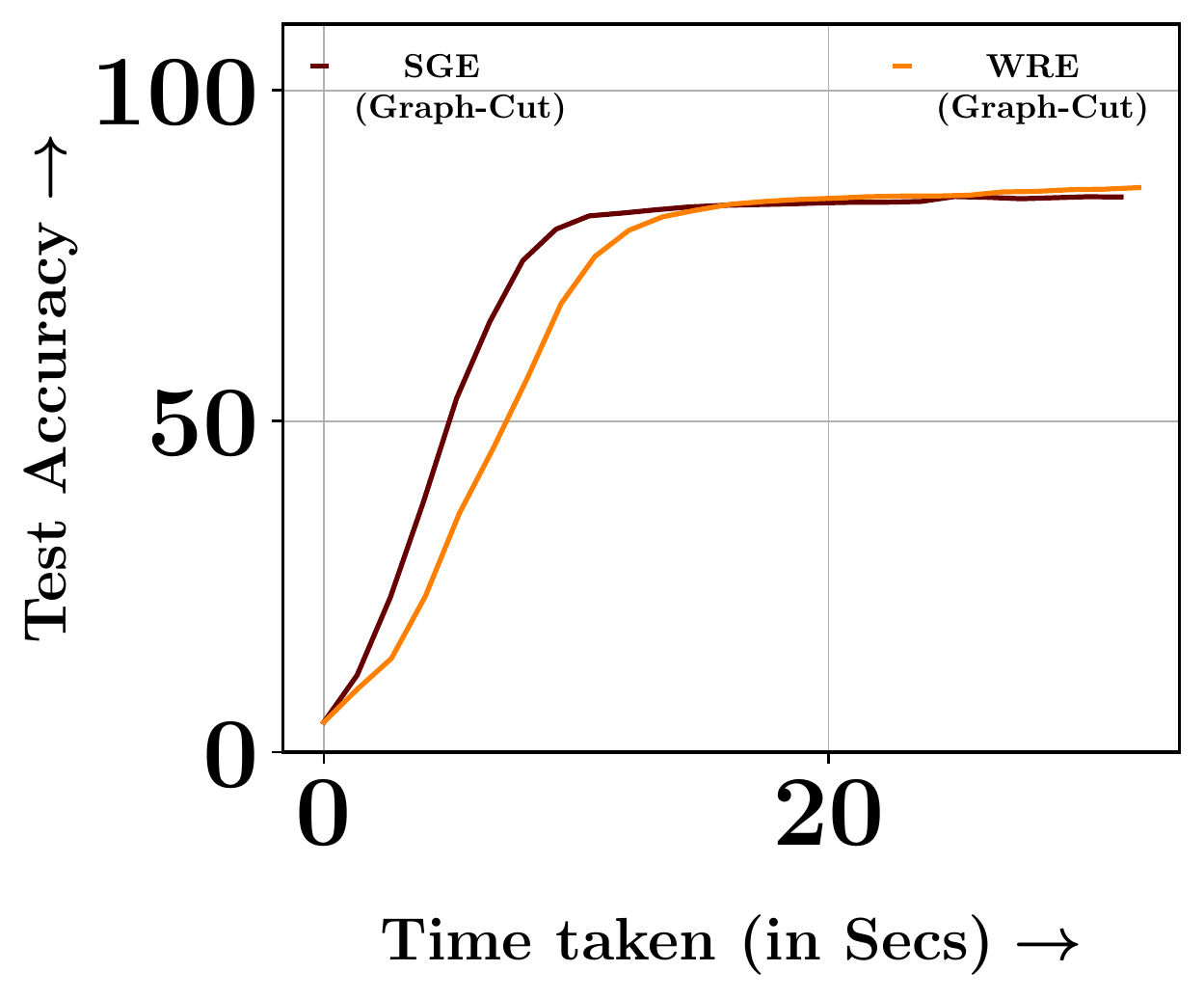}
\caption*{(d) \scriptsize TREC6(30\%) \\ LSTM}
\phantomcaption
\label{fig:trec6_wre}
\end{subfigure}
\begin{subfigure}[b]{0.32\textwidth}
\centering
\includegraphics[width=4.5cm]{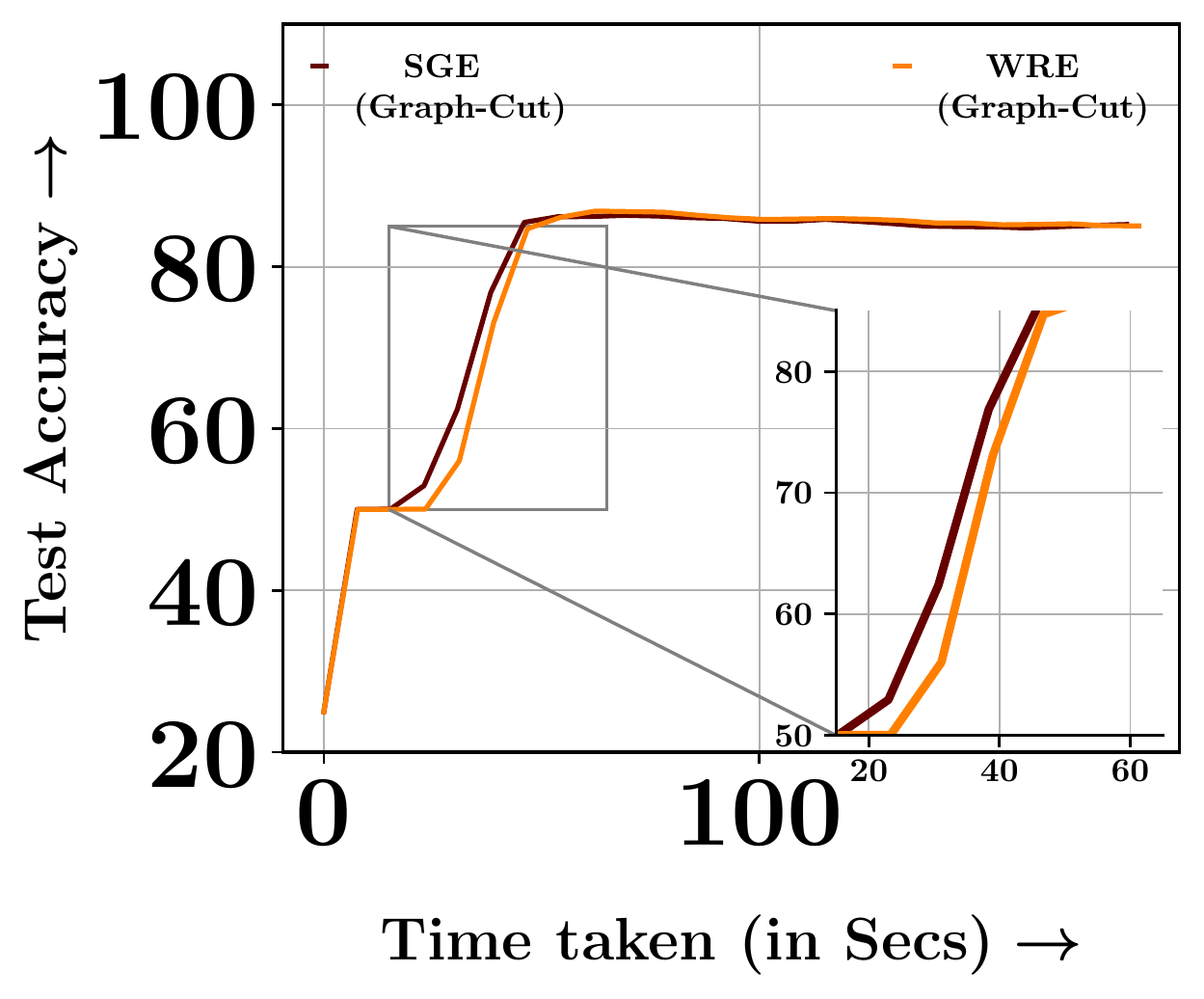}
\caption*{(e) \scriptsize IMDB (30\%) \\ LSTM}
\phantomcaption
\label{fig:imdb_wre}
\end{subfigure}
\begin{subfigure}[b]{0.32\textwidth}
\centering
\includegraphics[width=4.5cm]{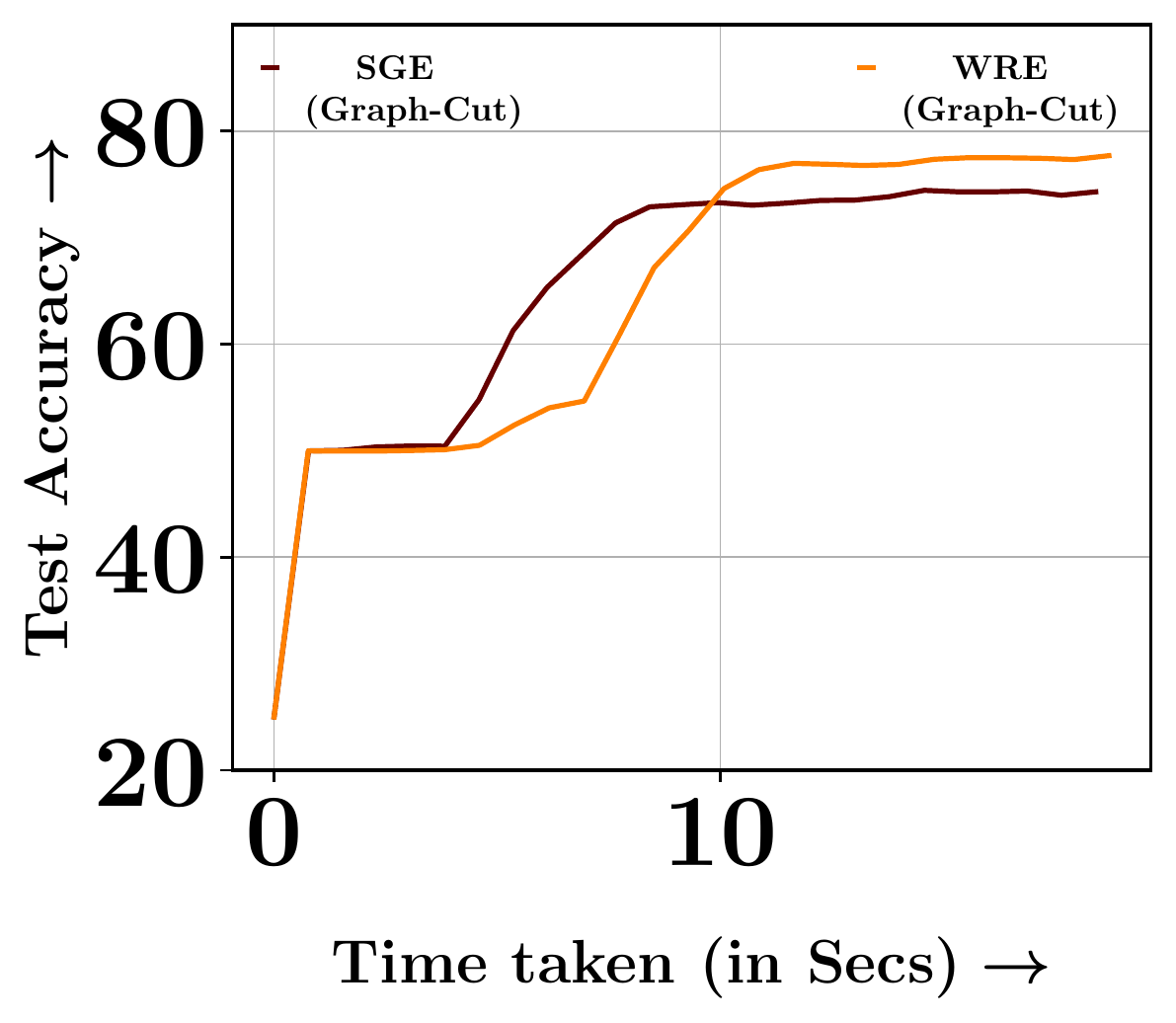}
\caption*{(f) \scriptsize Rotten Tomatoes (10\%) \\ LSTM}
\phantomcaption
\label{fig:rotten_tomatoes_wre}
\end{subfigure}
\caption{\footnotesize{Comparison of initial convergence of SGE with Graph-Cut and WRE with Graph-Cut on a variety of datasets using different subset sizes. Results show that SGE with Graph-Cut achieves faster initial convergence compared to WRE with Graph-Cut across all the datasets considered and for different subset sizes.}}
\label{fig:sge_gc_vs_wre_gc}
\end{figure}

Figure~\ref{fig:sge_gc_vs_wre_gc} shows the model convergence curves trained using SGE with Graph-Cut and WRE with Graph-Cut on various datasets using different subset sizes. Results show that SGE with Graph-Cut achieves faster initial convergence than WRE with Graph-Cut across all the datasets considered and for different subset sizes. The reason why SGE with Graph-Cut gives superior initial model convergence compared to WRE with Graph-Cut is that SGE emphasizes more exploitation than WRE. More specifically, SGE results in selecting subsets with high set function values because of the approximation guarantees of the stochastic-greedy algorithm~\citep{mirzasoleiman2015lazier}. Whereas with WRE, there is no guarantee that the sampled subsets using the constructed probability distribution $\vp$ have set function values. Since subsets with higher Graph-Cut function values correspond to subsets with more easy samples, SGE with Graph-Cut achieves superior model convergence than SGE with Facility Location.

\begin{table}[!ht]
\scalebox{0.7}{\begin{tabular}{c c c | c  c  c  c  c  c  c c} \hline \hline
\multicolumn{11}{c}{Tuning of the hyper-parameter: $\kappa$}\\ \hline
\hline
\multicolumn{3}{c|}{} & \multicolumn{8}{c}{Mean Test Accuracy of the Model(for 5 runs)} \\ 
\multicolumn{1}{c}{} & \multicolumn{1}{c}{} & \multicolumn{1}{c|}{Graphcut Interval} & \multicolumn{1}{c}{$0$} & \multicolumn{1}{c}{$\frac{1}{12}$} & \multicolumn{1}{c}{$\frac{1}{10}$} & \multicolumn{1}{c}{$\frac{1}{8}$} & \multicolumn{1}{c}{$\frac{1}{6}$} & \multicolumn{1}{c}{$\frac{1}{4}$} & \multicolumn{1}{c}{$\frac{1}{2}$} & \multicolumn{1}{c}{$1$}\\ \hline
\multicolumn{1}{c}{Dataset} & \multicolumn{1}{c}{Model}  &\multicolumn{1}{c|}{Budget} & \multicolumn{8}{c}{} \\ \hline
\textsc{Cifar100} & ResNet101 &   1\%   & 4.116\% & 9.27 \% & 11.795 \% & 12.135\% & \textbf{12.985\%} & 10.68\% & 11.11\% & 3.48\% \\
 & & 5\%  & 45.19\% & 45.7\%  & 45.43\%   & 47.15\%  & \textbf{49.885\%} & 49.01\% & 48.77\% & 35.76\%\\
& & 10\% & 60.4\%  & 64.08\% & 64.07\%   & 64.84\%  & \textbf{66.11\%}  & 64.97\% & 62.43\% &  53.41\%\\
& & 30\% & 74.28\% & 74.39\% & 75.24\%   & 74.27\%  & \textbf{75.28\%}   & 74.68\% & 73.18\% & 71.48\%\\ \cline{3-11}
\textsc{Cifar10} & ResNet101 & 1\%  & 38.18\% & 39.23\% & 34.61\% & 38.25\% & 43.37\% & \textbf{43.92\%} & 38.885\% & 38.76\% \\
& & 5\% & 71.77\% & 74.13\% & 77.98\% & 75.14\% & \textbf{81.63\%} & 77.87\% & 75.18\% & 63.9\% \\
& & 10\% & 87.21\% & 85.74\% & 87.32\% & 87.96\% & \textbf{88.72\%} & 88.17\% & 86.61\% & 83.25\% \\
& & 30\% & 93.79\% & 93.82\% & 93.52\% & 93.75\% & \textbf{94.24\%} & 93.73\% & 93.5\% & 91.23\%\\ \hline
\end{tabular}}
\caption{Mean test set accuracy of ResNet101 trained on \textsc{Cifar10} and \textsc{Cifar100} dataset with different subset sizes (of 1\%, 5\%, 10\%, and 30\%) and 200 epochs using curriculum-based data exploration with various curriculum intervals.}
\label{tab:graphcut_interval_results}
\end{table}

\begin{figure}[!ht]
  \centering
  \begin{subfigure}[b]{0.48\columnwidth}
    \includegraphics[width=\linewidth]{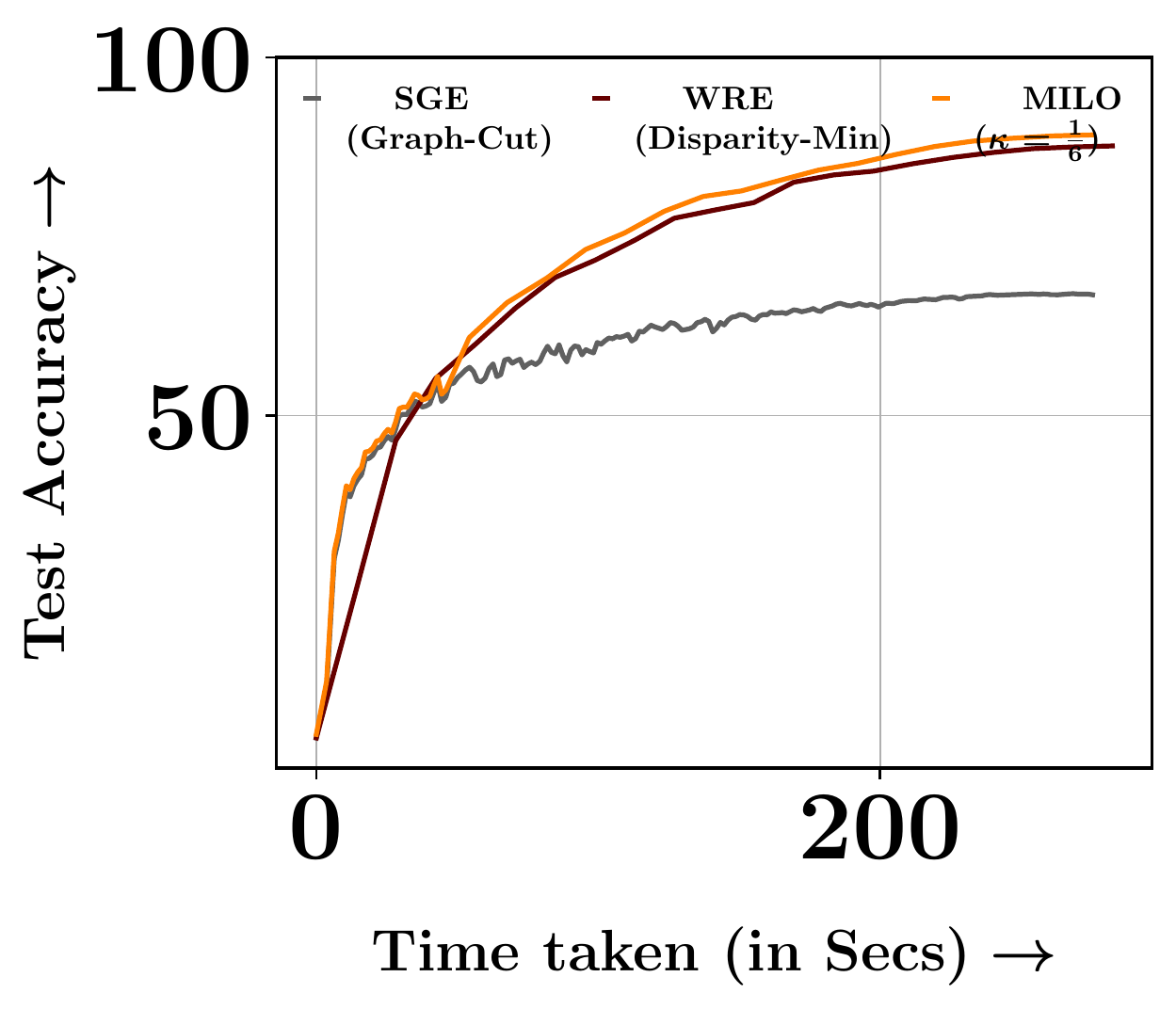}
    \caption{ResNet18 Convergence on \textsc{Cifar10} using 5\% subsets}
    \label{subfig:cifar10_conv_curriculum_comp}
  \end{subfigure}
  \hfill 
  \begin{subfigure}[b]{0.48\columnwidth}
    \includegraphics[width=\linewidth]{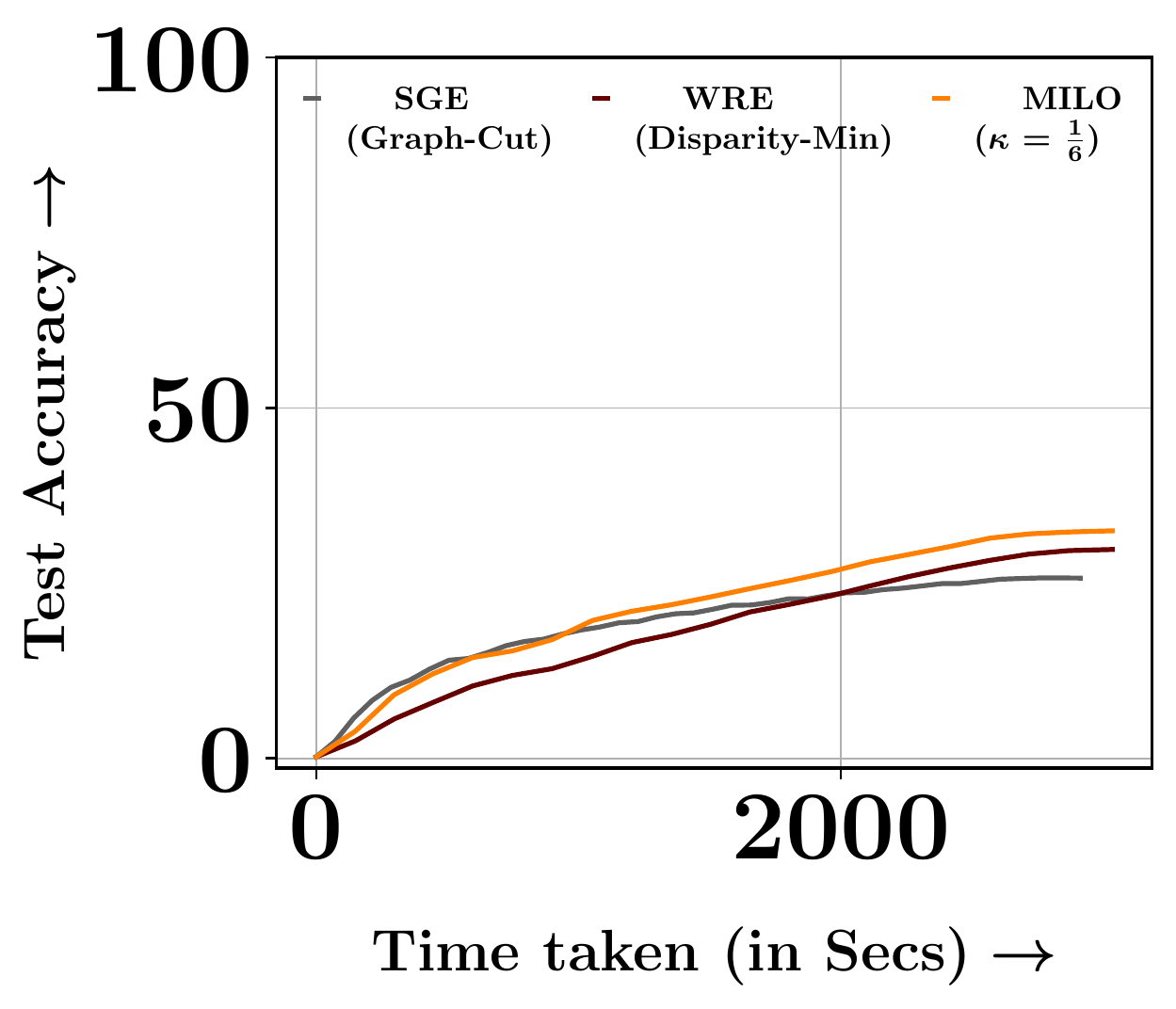}
    \caption{ResNet18 Convergence on \textsc{TinyImageNet} using 5\% subsets}
    \label{subfig:tinyimagener_conv_curriculum_comp}
  \end{subfigure}
  \caption{\footnotesize{Sub-figure (a) shows the convergence rate of the ResNet18 model on \textsc{Cifar10} dataset using 5\% subsets selected by SGE with Graph-Cut, WRE with Disparity-Min, and \model\ with curriculum exploration. Sub-figure (b) shows the convergence rate of the ResNet18 model on \textsc{TinyImageNet} dataset using 5\% subsets selected by SGE with Graph-Cut, WRE with Disparity-Min, and \model\ with curriculum exploration.}}
    \label{fig:curriculum_conv_comp}
\end{figure}

\subsection{Advantages of using a Curriculum for Data Exploration} \label{app:curriculum_ablation}
Results given in Table~\ref{tab:graphcut_interval_results} show that $\kappa=0$ (WRE with disparity-min) and $\kappa=1$ (SGE with graph-cut) perform poorly compared to using a curriculum $\kappa>0$, thereby proving the efficiency of curriculum-based data exploration over WRE in achieving better-performing models. Further, we also showcase the efficiency of curriculum-based data exploration in achieving faster convergence by showing the convergence curves of the ResNet18 model trained on the \textsc{TinyImageNet}, \textsc{Cifar10} datasets using 5\% subsets in Figure~\ref{fig:curriculum_conv_comp}. The convergence curves show that curriculum-based data exploration converges faster than WRE with disparity-min while achieving better performance than SGE with the graph-cut function.

\subsubsection{Finding Optimal Curriculum for Data Exploration} \label{app:kappa_tuning}
In order to find the optimal value of $\kappa$, we test for a number of different values that represent the fraction of the total number of epochs for which the model is initially trained using SGE with the graph-cut function. We present the mean test accuracies of the ResNet101 model trained for 200 epochs on different subset sizes of the \textsc{Cifar10} and \textsc{Cifar100} datasets in Table~\ref{tab:graphcut_interval_results}. Based on our observations, $\kappa=\frac{1}{6}$ is the optimal value and results in higher model accuracy. In Table~\ref{tab:graphcut_interval_results}, the results given in column $\kappa=0$ correspond to the ResNet101 model trained by simply using WRE with the disparity-min set function without any curriculum. At the same time, using very large values of $\kappa$ prioritizes more SGE and results in poor performance as SGE is shown to be less effective than WRE based on the results given in Sub-figure~\ref{fig:data_exploration_submod_comp}. 

\subsection{Optimal \texorpdfstring{$R$}{R} analysis} \label{app:R_analysis}
In order to find the optimal $R$ value($R$ signifies how frequently we select a new subset using \model\ ), we experiment with $R$ values of [1, 2, 5, 10] on the \textsc{Cifar100} dataset using ResNet18 model for subsets sizes of 10\%, and 30\%. Mean-test accuracy of the ResNet18 model obtained using different $R$ values is presented in Table~\ref{tab:R_analysis}. Results show that using $R=1$,i.e., selecting a new subset every epoch, results in a better-performing model than using higher values of $R$. Further, with small subset sizes, the gap between the model's performance obtained using $R=1$ and other values of $R$ is even more significant. This shows that it is paramount to select new subsets as frequently as possible when using small subset sizes to achieve the best possible performance.

\begin{table}[!ht]
\centering
\scalebox{0.8}{
\begin{tabular}{c c c | c  c  c  c} \hline \hline
\multicolumn{7}{c}{Tuning of the hyper-parameter: $R$}\\ \hline
\hline
\multicolumn{3}{c|}{} & \multicolumn{4}{c}{Mean Test Accuracy of the Model(for 5 runs)} \\ 
\multicolumn{1}{c}{} & \multicolumn{1}{c}{} & \multicolumn{1}{c|}{R} & \multicolumn{1}{c}{$1$} & \multicolumn{1}{c}{$2$} & \multicolumn{1}{c}{$5$} & \multicolumn{1}{c}{$10$}\\ \hline
\multicolumn{1}{c}{Dataset} & \multicolumn{1}{c}{Model}  &\multicolumn{1}{c|}{Budget} & \multicolumn{4}{c}{} \\ \hline
\textsc{Cifar100} & ResNet18 &   10\%   & \textbf{69.28\% $\pm$ 0.1041} & 69.21\% $\pm$ 3.82 & 66.63\% $\pm$ 3.38  & 64.07\% $\pm$ 3.32 \\
 & & 30\%  & \textbf{74.95\% $\pm$ 0.39}  & 74.13\% $\pm$ 0.36& 73.55\% $\pm$ 0.16 & 72.72\% $\pm$ 0.01   \\\hline
\end{tabular}}
\caption{Mean test set accuracy of ResNet18 trained on \textsc{Cifar100} dataset for subset sizes of 10\%, and 30\% selected using \model{} for 200 epochs for different values of $R$.}
\label{tab:R_analysis}
\end{table}

\subsection{Effectiveness of WRE}
In our previous discussions, we delved into how Stochastic Greedy Exploration (SGE) tends to favor subsets with near-optimal values, providing room for a modest degree of exploration. In stark contrast, Weighted Random Exploration (WRE) advocates for a more thorough exploration of data while simultaneously prioritizing the inclusion of highly informative samples.

In a bid to reinforce the effectiveness of WRE, we carry out an empirical comparison with a variant of SGE, particularly one that promotes increased exploration. The operational procedure of this SGE variant is as follows: We earmark a subset size $k^{'}$ to be selected by SGE, where $k^{'} \leq k$ and $k$ symbolize the overall budget. The balance of the samples are selected at random. As the training sequence progresses, the size of the subset chosen by SGE ($k^{'}$) is steadily reduced, thereby facilitating a broader scope for exploration. 

\begin{table}[h]
\centering
\begin{tabular}{|c|c|c|c|c|}
\hline
Dataset & Model & Fraction & Strategy & Test Accuracy \\ 
\hline
CIFAR100 & ResNet18 & 0.05 & MILO & \textbf{64.4} \\ 
 & & & SGE Variant (more exploration) & 59.02 \\ 
\hline
 & & 0.1 & MILO & \textbf{69.28} \\ 
 & & & SGE Variant (more exploration) & 66.24 \\ 
\hline
\end{tabular}
\caption{Comparison of Test Accuracy Between MILO and SGE Variant (that favors more exploration) on CIFAR100 with ResNet18 Model to validate the effectiveness of WRE in MILO}
\label{tab:sgev_cifar100}
\end{table}

\begin{table}[h]
\centering
\begin{tabular}{|c|c|c|c|c|}
\hline
Dataset & Model & Fraction & Strategy & Test Accuracy \\ 
\hline
CIFAR10 & ResNet18 & 0.05 & MILO & \textbf{88.37} \\ 
 & & & SGE Variant (more exploration) & 84.38 \\ 
\hline
 & & 0.1 & MILO & \textbf{91.11} \\ 
 & & & SGE Variant(more exploration) & 88.88 \\ 
\hline
\end{tabular}
\caption{Comparison of Test Accuracy Between MILO and SGE Variant (that favors more exploration) on CIFAR10 with ResNet18 Model to validate the effectiveness of WRE in MILO}
\label{tab:sgev_cifar10}
\end{table}

We contrast the performance of WRE with the aforementioned SGE variant, which integrates samples chosen by SGE with randomly selected samples. In the case of the SGE variant, we utilize a cosine decay rate, allowing the ratio of SGE-chosen samples to total samples to decay from 1 to 0 over the course of the training period. 

The results obtained from the CIFAR100 and CIFAR10 for both WRE and the SGE variant are presented in Tables~\ref{tab:sgev_cifar100} and ~\ref{tab:sgev_cifar10}. The outcomes suggest that WRE proves to be more effective in comparison to SGE, even when the latter allows for increased exploration. Consequently, we provide empirical evidence to suggest that WRE serves as a critical component of the \textsc{Milo}, contributing significantly to enhanced convergence.

\subsection{Comparison with Self-supervised Data Pruning Metric}
The self-supervised data pruning metric~\citep{sorscher2022beyond} is applied to prune a static subset from the entire dataset. As stated previously in Section~\ref{sec:methodology}, the utilization of such fixed data subsets necessitates selecting larger proportions in order to approach optimal performance levels, akin to full data training. This requirement, unfortunately, compromises efficiency.

\begin{table}[!ht]
    \centering
    \begin{tabular}{|l|l|l|l|l|l|}
    \hline
        Dataset & Model & Fraction & Strategy & Test Accuracy & SpeedUp \\ \hline
        CIFAR100 & ResNet18 & 0.3 & MILO & \textbf{74.95} & \textbf{3.4} \\ \hline
        ~ & ~ & 0.3 & Self-Supervised Metric & 60.41 & 3.4 \\ \hline
        ~ & ~ & 0.7 & Self-Supervised Metric & 74.31 & 1.4 \\ \hline
    \end{tabular}
\caption{Comparison of MILO with self-supervised pruning metric~\citep{sorscher2022beyond} on the CIFAR100 dataset using ResNet18 model.}
\label{tab:selfsuppruning}
\end{table}

We proceed to detail the results obtained from the CIFAR100 dataset using the ResNet18 model, in which we compare MILO and the self-supervised data pruning metric~\citep{sorscher2022beyond}. The findings, presented in Table~\ref{tab:selfsuppruning}, indicate a more substantial loss in performance when utilizing a 30\% fixed subset as determined by the self-supervised data pruning metric, as compared to using MILO.

Moreover, to attain performance levels similar to those achieved by \textsc{Milo} using a 30\% subset, it becomes necessary to select a 70\% subset via the self-supervised data pruning metric. This, in turn, results in forfeiting the speedups achieved.

\end{document}